\pdfoutput=1
\RequirePackage{fix-cm}
\documentclass[twocolumn]{svjour3}          % twocolumn
\smartqed  % flush right qed marks, e.g. at end of proof
\usepackage{graphicx}
\usepackage{times}
\usepackage{epsfig}
\usepackage{epstopdf}
\usepackage{amsmath}
\usepackage{amssymb}

\usepackage{caption}
\usepackage{tabularx}
\usepackage[normalem]{ulem}
\usepackage{algorithm}
\usepackage[noend]{algpseudocode}
\usepackage[section]{placeins}

\usepackage{cite}
\usepackage{color}
\usepackage{url}
\usepackage{booktabs}
\usepackage{bm}

\usepackage{multirow}
\usepackage[table,xcdraw]{xcolor}
\usepackage{array}
\usepackage{enumerate}

\usepackage{tikz}
\usetikzlibrary{shapes.geometric,calc}
\newcommand\score[2]{
\pgfmathsetmacro\pgfxa{#1+1}
\tikzstyle{scorestars}=[star, star points=5, star point ratio=2.25, draw,inner sep=0.15em,anchor=outer point 3]
\begin{tikzpicture}[baseline]
  \foreach \i in {1,...,#2} {
    \pgfmathparse{(\i<=#1?"black":"white")}  %%background color is white and foreground color is black
    \edef\starcolor{\pgfmathresult}
    \draw (\i*1em,0) node[name=star\i,scorestars,fill=\starcolor]  {};
   }
   \pgfmathparse{(#1>int(#1)?int(#1+1):0}
   \let\partstar=\pgfmathresult
   \ifnum\partstar>0
     \pgfmathsetmacro\starpart{#1-(int(#1))}
     \path [clip] ($(star\partstar.outer point 3)!(star\partstar.outer point 2)!(star\partstar.outer point 4)$) rectangle
    ($(star\partstar.outer point 2 |- star\partstar.outer point 1)!\starpart!(star\partstar.outer point 1 -| star\partstar.outer point 5)$);
     \fill (\partstar*1em,0) node[scorestars,fill=black]  {};
   \fi

\end{tikzpicture}
}

\newcommand{\PreserveBackslash}[1]{\let\temp=\\#1\let\\=\temp}
\newcolumntype{C}[1]{>{\PreserveBackslash\centering}p{#1}}
\journalname{IJCV}
\begin{document}

\title{Weighted Low-rank Tensor Recovery for Hyperspectral Image Restoration\thanks{This work was supported in part by the projects of the National Natural Science Foundation of China under Grants No. 61571207, 61433007 and 41501371.}
}

\author{Yi Chang         \and
        Luxin Yan          \and
        Houzhang Fang   \and
        Sheng Zhong       \and
        Zhijun Zhang.
}

\institute{Yi Chang, Luxin Yan,  Sheng Zhong and Zhijun Zhang\at
              Science and Technology on Multispectral Information Processing Laboratory, School of Automation, Huazhong University of Science and Technology, Wuhan, 430074, China. \\
              \email{yichang, yanluxin, zhongsheng, zhijunzhang@hust.edu.cn}           %  \\
           \and
           Houzhang Fang \at
           National Laboratory of Radar Signal Processing, Xidian University, Xi'an, 710071, China
           \email{houzhangfang@gmail.com}
}

\vspace{-5mm}
\date{Received: date / Accepted: date}

\maketitle
\begin{abstract}
  Hyperspectral imaging, providing abundant spatial and spectral information simultaneously, has attracted a lot of interest in recent years. Unfortunately, due to the hardware limitations, the hyperspectral image (HSI) is vulnerable to various degradations, such noises (random noise, HSI denoising), blurs (Gaussian and uniform blur, HSI deblurring), and down-sampled (both spectral and spatial downsample, HSI super-resolution). Previous HSI restoration methods are designed for one specific task only. Besides, most of them start from the 1-D vector or 2-D matrix models and cannot fully exploit the structurally spectral-spatial correlation in 3-D HSI. To overcome these limitations, in this work, we propose a unified low-rank tensor recovery model for comprehensive HSI restoration tasks, in which non-local similarity between spectral-spatial cubic and spectral correlation are simultaneously captured by 3-order tensors. Further, to improve the capability and flexibility, we formulate it as a weighted low-rank tensor recovery (WLRTR) model by treating the singular values differently, and study its analytical solution. We also consider the exclusive stripe noise in HSI as the gross error by extending WLRTR to robust principal component analysis (WLRTR-RPCA). Extensive experiments demonstrate the proposed WLRTR models consistently outperform state-of-the-arts in typical low level vision HSI tasks, including denoising, destriping, deblurring and super-resolution.
\end{abstract}

%%%%%%%%%%%%%%%%%%%%%%%%
\section{Introduction}
\label{intro}
%%%%%%%%%%%%%%%%%%%%%%%%
  Hyperspectral image consists of multiple discrete bands at specific frequencies. The HSI can deliver additional information the human eye fails to capture for real scenes, and has been attracting a lot of interests for the research from wide range of application fields, such as anomaly detection \cite{du2014discriminative}, classification \cite{ji2014spectral}, and environmental monitoring \cite{moroni2013hyperspectral}. However, HSI always suffers from various degradations, such as the random noise (caused by photon effects), stripe noise (due to calibration error between adjacent detectors), blurring (on account of atmospheric turbulence or system motion), and low spatial resolution (because of the exposure time). It is economically unsustainable and impractical to improve its quality in HSI merely by hardware scheme. Therefore, it is natural to introduce the image processing based approaches for obtaining a high quality HSI before the subsequent applications. Mathematically, the problem of HSI restoration can be generally formulated by a linear model as follow:
  \begin{equation}\label{eq:Spatial_Degradation}
  \setlength{\abovedisplayskip}{2pt}
  \setlength{\belowdisplayskip}{2pt}
  {\boldsymbol{\mathcal{Y}}} = {\mathcal{T}_{sa}}({\boldsymbol{\mathcal{X}}}) + {\boldsymbol{\mathcal{E}}} + {\boldsymbol{\mathcal{N}}},
  \end{equation}
  where ${\boldsymbol{\mathcal{Y}}} \in {\mathbb{R}^{r \times c \times B}}$ is an observed low spatial-resolution image, ${\boldsymbol{\mathcal{X}}} \in {\mathbb{R}^{R \times C \times B}}$ ($r \ll R,c \ll C$) represents the original high spatial-resolution image, ${\boldsymbol{\mathcal{E}}} \in {\mathbb{R}^{r \times c \times B}}$ denotes the sparse error (mainly the stripe noise), ${\boldsymbol{\mathcal{N}}} \in {\mathbb{R}^{r \times c \times B}}$ means the additive random noise, and ${\mathcal{T}_{sa}}( \bullet )$ stands for the spatially linear degradation operator on the clean HSI. With different settings, Eq. (\ref{eq:Spatial_Degradation}) can represent different HSI restoration problems; When ${\mathcal{T}_{sa}}( \bullet )$ is an identity tensor, the problem (\ref{eq:Spatial_Degradation}) becomes HSI denoising (only consider ${\boldsymbol{\mathcal{N}}}$) or HSI destriping (only consider ${\boldsymbol{\mathcal{E}}}$), or HSI mixed noise removal (both ${\boldsymbol{\mathcal{N}}}$ and ${\boldsymbol{\mathcal{E}}}$); when ${\mathcal{T}_{sa}}( \bullet )$ is a blur operator, the problem (\ref{eq:Spatial_Degradation}) turns into the HSI deblurring; For HSI super-resolution, ${\mathcal{T}_{sa}}( \bullet )$ is a composite operator of blurring and spatial down-sampling. Moreover, in HSI super-resolution, there is another guided low spectral resolution multispectral image ${\boldsymbol{\mathcal{Z}}} \in {\mathbb{R}^{R \times C \times b}}$ ($b \ll B$) (usually RGB image), which can be formulated as follow:
  \begin{equation}\label{eq:Spectral_Degradation}
  \setlength{\abovedisplayskip}{2pt}
  \setlength{\belowdisplayskip}{2pt}
  {\boldsymbol{\mathcal{Z}}} = {\mathcal{T}_{se}}({\boldsymbol{\mathcal{X}}}) + {\boldsymbol{\mathcal{N}}},
  %{\boldsymbol{\mathcal{Z}}} = {\boldsymbol{\mathcal{X}}}{ \times _3}\textbf{\emph{P}} + {\boldsymbol{\mathcal{N}}},
  \end{equation}
  where ${\mathcal{T}_{se}}( \bullet )$ denotes a spectral downsampling procedure, which can be expressed as ${\boldsymbol{\mathcal{X}}}{ \times _3}\textbf{\emph{P}}$, and $\textbf{\emph{P}} \in {\mathbb{R}^{b \times B}}$ is a transformation matrix mapping the HSI ${\boldsymbol{\mathcal{X}}} \in {\mathbb{R}^{R \times C \times B}}$ to its RGB representation ${\boldsymbol{\mathcal{Z}}} \in {\mathbb{R}^{R \times C \times b}}$ ($b \ll B$). The tensor product is defined in Section II.

  To cope with the ill-posed nature of the HSI restoration task, various prior knowledge of the HSI is proposed to regularize the solution space:
  \begin{equation}\label{eq:Regularization Framework}
  \setlength{\abovedisplayskip}{2pt}
  \setlength{\belowdisplayskip}{2pt}
  \mathop {\min }\limits_{\boldsymbol{\mathcal{X}}} \frac{1}{2}||{\boldsymbol{\mathcal{Y}}} - {\mathcal{T}_{sa}}({\boldsymbol{\mathcal{X}}}) - {\boldsymbol{\mathcal{E}}}||_F^2 + \frac{1}{2}||{\boldsymbol{\mathcal{Z}}} - {\mathcal{T}_{se}}({\boldsymbol{\mathcal{X}}})||_F^2 + \lambda \Phi ({\boldsymbol{\mathcal{X}}}),
  \end{equation}
  where $|| \bullet ||_F^2$ stands for the Euclidean norm, the first two data fidelity terms represent the spatial and spectral degradation process respectively, $\Phi ({\boldsymbol{\mathcal{X}}})$ is a regularization term to enforce the solution with desired property, and $\lambda$ is a trade-off regularization parameter. The success of the HSI restoration heavily depends on how we choose proper prior knowledge. From the perspective of the data format in the prior, we classify the existing HSI restoration methods into three categories: one-dimensional vector based sparse representation methods, two-dimensional matrix based low-rank matrix recovery methods, and three-dimensional tensor based approximation methods.

  Sparse representation methods assumes that the clean signals lie in a low-dimensional subspace and can be well approximated by the linear combination of a few atoms in a dictionary \cite{elad2006image}. Othman and Qian \cite{othman2006noise} presented a hybrid spatial-spectral derivative domain wavelet shrinkage model with a fixed wavelet dictionary to reduce the noise in HSI. In \cite{dong2016hyperspectral, wei2015hyperspectral, akhtar2014sparse}, the authors learned the dictionary directly from the observed images for HSI super-resolution. The above mentioned methods are synthesis based, while analysis-based sparse representation methods transform a signal into sparsity domain through various forward measurements on it \cite{cai2012image}, such as well-known total variation \cite{rudin1992nonlinear} with fixed gradient filtering template. Yuan et al. \cite{yuan2012hyperspectral} proposed a HSI denoising algorithm by employing a spectral-spatial adaptive total variation model.

  With the development of the robust principle component analysis (RPCA) \cite{wright2009robust}, the two-dimensional low-rank matrix recovery methods have shown its effectiveness to discover the intrinsic low-dimensional structures in high-dimensional HSI data. In \cite{zhang2014hyperspectral}, by lexicographically ordering the 3-D cube into a 2-D matrix representation along the spectral dimension, Zhang et al. proposed a low-rank matrix restoration model for mixed noise removal in HSI. A lot of works follow this research line \cite{zhao2015hyperspectral, xie2016hyperspectral, he2016total}. On the contrary, Veganzones et al. \cite{veganzones2016hyperspectral} considered that the HSI are spatially local low rank for HSI super-resolution, since spatial neighborhood pixels represented the same materials span a very low-dimensional subspace/manifold. In \cite{chang2016remote}, we proposed a globally low-rank decomposition model for HSI destriping, since only parts of data vectors are corrupted by the stripes but the others are not.

  The tensor based method \cite{Lathauwer2000A} is quite suitable for multidimensional HSI processing. The pioneer works \cite{letexier2008noise, liu2012denoising, guo2013hyperspectral} have achieved promising results in HSI denoising. Recently, when the tensor decomposition meets the sparsity property, its potential has been further tapped and enlightened state-of-the-arts HSI denoising work. In \cite{peng2014decomposable}, the authors proposed a tensor dictionary learning model for the task of HSI denoising with enforcing hard constraints on the rank of the core tensor. Dong et al. \cite{dong2015low} presented a novel low-rank tensor approximation framework with Laplacian Scale Mixture (LSM) modeling in a principle manner, in which the LSM parameters and sparse tensor coefficients can be jointly optimized with closed-form solutions. In \cite{Chang2017Hyper}, we give a detailed analysis about the rank properties both in matrix and tensor cases, a unidirectional low-rank tensor recovery model is proposed for HSI denoising.

  However, there are three main drawbacks in the aforementioned methods. Firstly, transforming the multi-way HSI data into a vector or matrix usually leads to lose the spectral-spatial structural correlation. The latest work \cite{xie2016multispectral, dian2017hyperspectral} consistently indicate that the tensor-based methods substantially preserve the intrinsic structure correlation with better restoration results. Besides, some tensor based methods are quite heuristic without taking the sparsity prior into consideration, which make them hard to be extended to other HSI restoration tasks. Secondly, compared with natural 2-D images, the HSI data could provide us extra multiple spectral information. Unfortunately, many HSI restoration methods fall into the 'trap' of high spectral correlation, while ignoring the non-local similarity \cite{he2016total}, and vice versa \cite{dian2017hyperspectral}. Thus the result of these approaches may be suboptimal. Finally, most of previous methods only employ the conventional $\textbf{\emph{L}}_1$ or nuclear norm as the sparsity constraint for HSI restoration. As a result, each patch is encoded equally. Such a mechanism may not take advantage of the image patch structure difference sufficiently. The reweighting strategy interprets the fine-grained structural discrepancy, and has proven to be effective in vector or matrix cases \cite{candes2008enhancing, liu2016truncated, Gu2014Weighted, Yan2013Nonlocal}, while it has received less attention in tensor based methods for HSI restorations.

  In this work, to overcome the aforementioned drawbacks, our starting point is to present a unified HSI restoration model from the tensor perspective in which the fine-grained intrinsic low-rank of the constructed tensor is took into consideration. The tensor format naturally offers a unified understanding for the vector/matrix-based recovery models. Compared with the state-of-the-art HSI restoration methods, the contributions of the proposed work are as follows:
  \begin{itemize}
    \item We propose an effective and universal tensor-based low-rank prior for HSI image modeling, and validate it on several representative HSI image restoration tasks, such as denoising, destriping, deblurring, and super-resolution. To our knowledge, this is the first work that \emph{comprehensively} considers the low-level HSI restoration tasks in a unified model.
    \item Most of HSI restoration methods encode the spectral or spatial information independently, ignores the spatial-spectral structural correlation. Our method employs the low-rank tensor prior to model the spatial non-local self-similarity and spectral correlation property simultaneously, better preserving the intrinsic spectral-spatial structural correlation.
    \item We extend the idea of the weighted strategy in matrix low-rank minimization problem \cite{Gu2014Weighted} to tensor case, where the singular values in the core tensor are with different importance and assigned different weights. This simple operation has immediate physical interpretation and facilitates better restoration results. Meanwhile, the analytical solution of the WLRTR model has been given.
    \item To handle the real noise case in HSI, including both the random and stripe noise, we propose a tensor based RPCA decomposition model, in which the structural stripe noise is regarded as the gross error. The $\textbf{\emph{L}}_{211}$ norm (defined in section III) is introduced to capture the directional and structural property of the stripe noise.
    \item For various low-level HSI restoration tasks, the proposed WLRTR model consistently outperforms the state-of-the-arts by a large marginal. Further, we demonstrate that the WLRTR can be well applied to multispectral images, such as three channel color image and MRI.
  \end{itemize}

  The remainder of this paper is organized as follows. The related HSI restoration methods are introduced in Section II. Section III presents the weighted low-rank tensor recovery modeling, and analyze its close-formed solution. Section IV proposes the concrete objective functional of each individual HSI restoration task, and gives the corresponding optimization procedures. Extensive experimental results are reported in Section V. Section VI concludes this paper.

%%%%%%%%%%%%%%%%%%%%%%%%%%%%%%%%%
\section{Related work}
\label{sec:related}
%%%%%%%%%%%%%%%%%%%%%%%%%%%%%%%%%
We discuss the related state-of-the-art work for various HSI restoration tasks in detail, including denoising, destriping, deblurring, and super-resolution, and compare our work with them.

   %%%%%%%%%%%%%%%%%%%%%%%%%%%%%%%%%%%%%%%%%%%%%%%%%%%
   \subsection{HSI Denoising}
   %%%%%%%%%%%%%%%%%%%%%%%%%%%%%%%%%%%%%%%%%%%%%%%%%%%
    Image denoising is a test bed for the various technique. Consequently, numerous approaches for HSI denoising have been proposed \cite{zhang2016cluster,fu20163d,fu2017adaptive}. The spectral correlation and nonlocal self-similarity are two kinds of intrinsic characteristic underlying a HSI. Most of previous HSI denoising methods focus on the spectral correlation such as the wavelet methods \cite{othman2006noise}, total variational methods \cite{yuan2012hyperspectral}, the low-rank matrix recovery methods \cite{zhang2014hyperspectral, he2016total, zhao2015hyperspectral,cao2015low}, or the nonlocal self-similarity such as BM4D \cite{Maggioni2012Nonlocal}, HOSVD \cite{rajwade2013image} individually. Recently, Peng et al. \cite{peng2014decomposable} firstly modeled them simultaneously in tensor format. However, the TDL \cite{peng2014decomposable} is quite heuristic and short of a concise formulation, thus lacking of the flexibility to other HSI restoration tasks. Several tensor works \cite{dong2015low, xie2016multispectral, Chang2017Hyper} followed the research line of \cite{peng2014decomposable}, and model the sparsity of the core tensor coefficients in a principled manner. Interested readers could refer to \cite{xie2017kronecker} for detailed background of HSI denoising. In this work, we further take the fine-grained intrinsic sparsity of the core tensor coefficients into consideration with the reweighting strategy, so as to better encode the structural correlation. Moreover, our WLRTR model can be well extended to other HSI issues.

   %%%%%%%%%%%%%%%%%%%%%%%%%%%%%%%%%%%%%%%%%%%%%%%%%%%
   \subsection{HSI Destriping}
   %%%%%%%%%%%%%%%%%%%%%%%%%%%%%%%%%%%%%%%%%%%%%%%%%%%
   Stripe noise is a very common structure noise in HSI. Traditional HSI destriping methods \cite{shen2009map, bouali2011toward, chang2015anisotropic, lu2013graph, zhang2014hyperspectral} treated this problem as a denoising task and estimate the image directly with Gaussian white noise assumption. Further, Meng et al. \cite{meng2013robust, cao2015low} hold the point that the stripe line is a kind of structure noise, and introduced the mixture of Gaussians (MoG) noise assumption to accommodate the stripe noise characteristic. On the contrary, some works started from the opposite direction by estimating the stripe noise only \cite{carfantan2010statistical, fehrenbach2012variational, liu2015stripe}. These methods regard the stripe noise as a kind of specific image structure with less variables and regular patterns, which makes the problem easier to be solved. Our recent work \cite{chang2016remote} proposed to treat the HSI destriping task as an image decomposition task, in which the clear image and stripe component are treated equally and estimated iteratively. Previous methods are all 2-D matrix based and most of them fail to capture the directional characteristic of the stripe noise. In this work, we present a tensor based RPCA decomposition method to simultaneously estimate the image and stripe, in which directional induced $\textbf{\emph{L}}_{211}$ norm (defined in section III) is introduced to appropriately capture the subspace of the stripe noise.

   %%%%%%%%%%%%%%%%%%%%%%%%%%%%%%%%%%%%%%%%%%%%%%%%%%%
   \subsection{HSI Deblurring}
   %%%%%%%%%%%%%%%%%%%%%%%%%%%%%%%%%%%%%%%%%%%%%%%%%%%
    Natural image deblurring aims to recover a sharp latent image from a blurred one \cite{levin2009understanding}, which is a classical and active research field within the last decade. Numerous HSI deblurring methods directly learn from the natural single image priors by assuming the widely used sparsity of image gradients, e.g., Huber-Markov prior \cite{shen2012blind}, the total variational (TV) \cite{zhao2013deblurring}, and Gaussian mixture model (GMM) \cite{zhong2015learning}. Only recently, the spatial-spectral joint total variational has been introduced to HSI deblurring \cite{henrot2013fast, fang2017hyperspectral}. In general, most of previous HSI deblurring methods only exploit the spatial information, while none of them have utilized the nonlocal self-similarity presented in HSI. In our work, we focus on the non-blind HSI deblurring, and show that the additional spectral correlation and nonlocal information would significantly improve the HSI deblurring performance.

   %%%%%%%%%%%%%%%%%%%%%%%%%%%%%%%%%%%%%%%%%%%%%%%%%%%
   \subsection{HSI Super-resolution}
   %%%%%%%%%%%%%%%%%%%%%%%%%%%%%%%%%%%%%%%%%%%%%%%%%%%
    HSI super-resolution refers to the fusion of a hyperspectral image (low spatial but high spectral resolution) with a panchromatic/multispectral image (high spatial but low spectral resolution, usually RGB image). The most popular sparsity promoting methods mainly include the sparse representation \cite{wei2015hyperspectral, akhtar2014sparse, jiang2015pan, li2013remote, chen2014image, akhtar2015bayesian} and the matrix factorization approach \cite{yokoya2012coupled, dong2016hyperspectral, kawakami2011high, wycoff2013non, lanaras2015hyperspectral}. In \cite{wei2015hyperspectral}, the authors applied the dictionary learning embedded in the spectral subspace to exploit the sparsity of hyperspectral images. In \cite{dong2016hyperspectral}, Dong et al. proposed a non-negative structured sparse representation (NSSR) approach with the prior knowledge about spatio-spectral sparsity of the hyperspectral image. Analog to classical super-resolution \cite{yang2010image}, a sparse matrix factorization method \cite{kawakami2011high} borrowed the idea that both the LR hyperspectral image and HR RGB image share the same coding coefficients. The HR hyperspectral image was then reconstructed by multiplying the learned basis from the HR RGB image and sparse coefficients from the LR hyperspectral image. The interested readers can refer to the survey \cite{loncan2015hyperspectral}.

    Tensor-based methods in HSI super-resolution is not entirely new. Very recently, Dian et al. \cite{dian2017hyperspectral} proposed a non-local sparse Tucker tensor factorization (NLSTF) model for HSI super-resolution. While both NLSTF and our technique tackle this problem from the tensor perspective, NLSTF fails to make use of an auxiliary HR RGB image. Similar to \cite{peng2014decomposable}, its realization is relatively heuristic, and hard to be incorporate additional information. Our unified WLRTR method by-passes the tensor dictionary learning process by using high order singular value decomposition (HOSVD) \cite{rajwade2013image}, benefiting us enough flexibility, such as the auxiliary HR RGB image. Besides, NLSTF does not apply any sparsity prior for the underlying HR HSI. In contrast, we introduce the weighted low-rank tensor prior to further refine the solution.

\begin{figure}
    \includegraphics[width=0.48\textwidth]{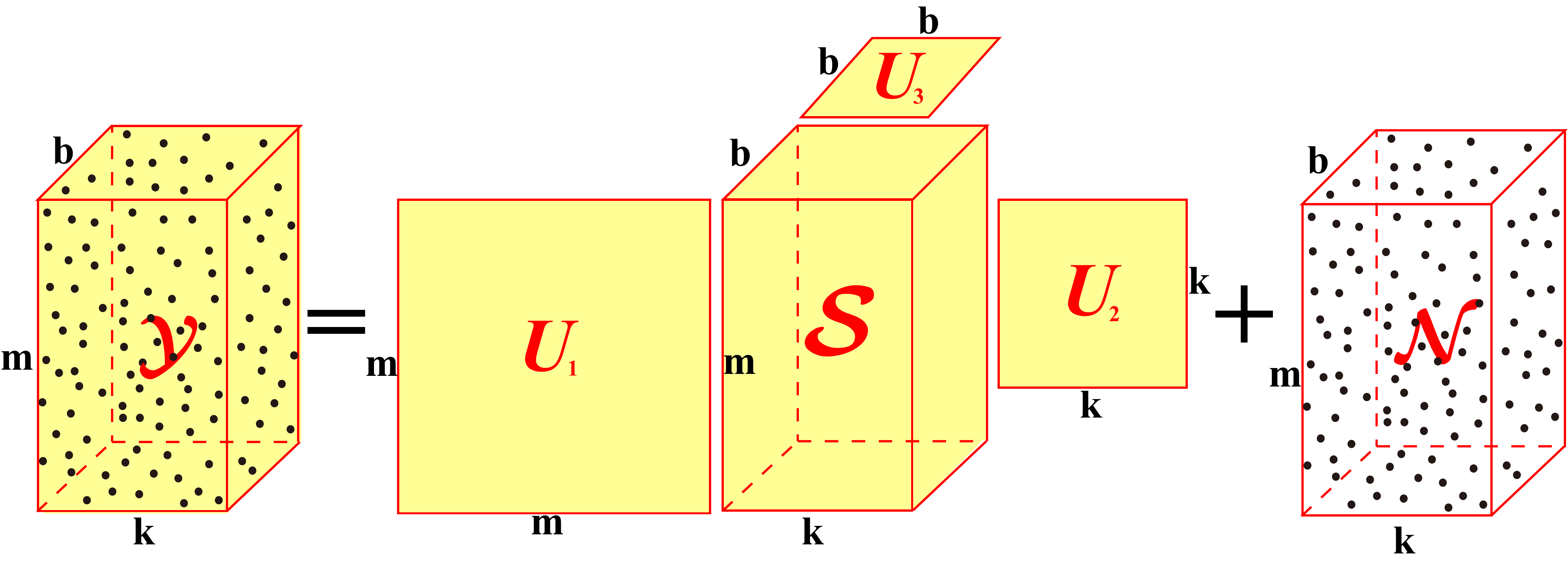}
   \caption{The illustration of the HOSVD.}
\label{Degradation including HOSVD}
\end{figure}

%%%%%%%%%%%%%%%%%%%%%%%%
\section{Weighted Low-rank Tensor Recovery Model}
\label{sec:classifiers}
%%%%%%%%%%%%%%%%%%%%%%%%

    In this section, we first introduce some notations and the low-rank approach for 2-D images. Then, we will present the advantageous of the tensor based method, and the details of our WLRTR model. At last, we offer its analytical solution.

   %%%%%%%%%%%%%%%%%%%%%%%%%%%%%%%%%%%%%%%%%%%%%%%%%%%
   \subsection{Notations and Preliminaries }
   %%%%%%%%%%%%%%%%%%%%%%%%%%%%%%%%%%%%%%%%%%%%%%%%%%%
     In this paper, we denote tensors by boldface Euler script letters, e.g., $\boldsymbol{\mathcal{X}}$. Matrices are represented as boldface capital letters, e.g., \textbf{\emph{X}}; vectors are expressed with boldface lowercase letters, e.g., \emph{\textbf{x}}, and scalars are denoted by lowercase letters, e.g., \emph{x}. The \emph{i}-th entry of a vector \textbf{\emph{x}} is denoted by ${x_i}$, element $(i,j)$ of a matrix \textbf{\emph{X}} is denoted by ${x_{ij}}$, and element $(i,j,k)$ of a third-order tensor $\boldsymbol{\mathcal{X}}$ is denoted by ${x_{ijk}}$.

    \emph{\textbf{Fibers}} are the higher-order analogue of matrix rows and columns. A fiber of an N-dimensional tensor is a 1-D vector defined by fixing all indices but one \cite{Kolda2005Tensor}. A \emph{\textbf{Slice}} of an N-dimensional tensor is a 2-D matrix defined by fixing all but two indices \cite{Kolda2005Tensor}. For a third order tensor, its column, row, and tube fibers, denoted by ${\textbf{\emph{x}}_{:jk}},{\textbf{\emph{x}}_{i:k}}, and {\textbf{\emph{x}}_{ij:}}$, respectively.

    \noindent
    \textbf{Definition 1 (Tensor norms)} The \emph{Frobenius} norm of an \emph{N} order tensor ${\boldsymbol{\mathcal{X}}} \in {\mathbb{R}^{{I_1} \times {I_2} \times  \cdots  \times {I_N}}}$ is the square root of the sum of the squares of all its elements, i.e., $||\boldsymbol{\mathcal{X}}|{|_F} = \sqrt {\sum\nolimits_{{i_1} = 1}^{{I_1}} {\sum\nolimits_{{i_2} = 1}^{{I_2}} { \cdots \sum\nolimits_{{i_N} = 1}^{{I_N}} {x_{{i_1}{i_2} \cdots {i_N}}^2} } } }$. The ${\mathcal{L}_1}$ norm of an \emph{N} order tensor is the sum of the absolute value of all its elements, i.e., $||\boldsymbol{\mathcal{X}}|{|_1} = \sum\nolimits_{{i_1} = 1}^{{I_1}} {\sum\nolimits_{{i_2} = 1}^{{I_2}} { \cdots \sum\nolimits_{{i_N} = 1}^{{I_N}} {|{x_{{i_1}{i_2} \cdots {i_N}}}|} } }$. These norms for tensor are analogous to the matrix norm.

    \noindent
    \textbf{Definition 2 (Tensor matricization)} Matricization, also named as \emph{unfolding or flattening}, is the process of reordering the elements of an \emph{N}-order tensor into a matrix. The mode-\emph{n} matricization ${\textbf{\emph{X}}_{(n)}} \in {\mathbb{R}^{{I_n} \times ({I_1} \cdots {I_{n - 1}}{I_n} \cdots {I_N})}}$ of a tensor ${\boldsymbol{\mathcal{X}}} \in {\mathbb{R}^{{I_1} \times {I_2} \times  \cdots  \times {I_N}}}$ is obtained by taking all the mode-\emph{n} fibers to be the columns of the resulting matrix.

    \noindent
    \textbf{Definition 3 (Tensor product)} Here we just consider the tensor \emph{n}-mode product, i.e., multiplying a tensor by a matrix in mode \emph{n}, which will be used in HOSVD latter. Interested readers can refer to Bader and Kolda \cite{Kolda2005Tensor} for a full treatment of tensor multiplication. For a tensor ${\boldsymbol{\mathcal{X}}} \in {\mathbb{R}^{{I_1} \times {I_2} \times  \cdots  \times {I_N}}}$, its \emph{n}-mode product with a matrix $\textbf{\emph{U}} \in {\mathbb{R}^{J \times {I_n}}}$ is denoted by ${\boldsymbol{\mathcal{Z}}} = {\boldsymbol{\mathcal{X}}}{ \times _n}\textbf{\emph{U}}$, and ${\boldsymbol{\mathcal{Z}}} \in {\mathbb{R}^{{I_1} \times  \cdots  \times {I_{n - 1}} \times J \times {I_{n + 1}} \times  \cdots  \times {I_N}}}$. Each element in $\boldsymbol{\mathcal{Z}}$ can be represented as
    \begin{equation}\label{eq:Tensor Product}
    {z_{{i_1} \cdots {i_{n - 1}}j{i_{n + 1}} \cdots {i_N}}} = \sum\limits_{{i_n} = 1}^{{I_n}} {{x_{{i_1}{i_2} \cdots {i_N}}}{u_{j{i_n}}}.}
    \end{equation}

    \noindent
    \textbf{Definition 4 (Tensor SVD)} The Tucker decomposition is a kind of higher-order SVD, which decomposes a tensor into a core tensor multiplied by a matrix along each mode as follow \cite{Lathauwer2000A}:
    \begin{equation}\label{eq:Tensor SVD}
    {\boldsymbol{\mathcal{X}}} = {\boldsymbol{\mathcal{S}}}{ \times _1}{\textbf{\emph{U}}_1}{ \times _2}{\textbf{\emph{U}}_2}{ \times _3} \cdots { \times _N}{\textbf{\emph{U}}_N},
    \end{equation}
    where ${\boldsymbol{\mathcal{X}}} \in {\mathbb{R}^{{I_1} \times {I_2} \times  \cdots  \times {I_N}}}$, ${\boldsymbol{\mathcal{S}}} \in {\mathbb{R}^{{I_1} \times {I_2} \times  \cdots  \times {I_N}}}$ is the \emph{core tensor} similar to the singular values in matrix, and its intensity shows the level of interaction between the different components, ${\textbf{\emph{U}}_i} \in {\mathbb{R}^{{I_i} \times {I_i}}}$ is the orthogonal factor matrix and can be regarded as the principal components in each mode. The main purpose of our work is to estimate the core tensors $\boldsymbol{\mathcal{S}}$ and the clean image tensor $\boldsymbol{\mathcal{X}}$ in presence of the degraded tensor $\boldsymbol{\mathcal{Y}}$, as shown in Fig. \ref{Degradation including HOSVD}.

\begin{figure}
    \includegraphics[width=0.5\textwidth]{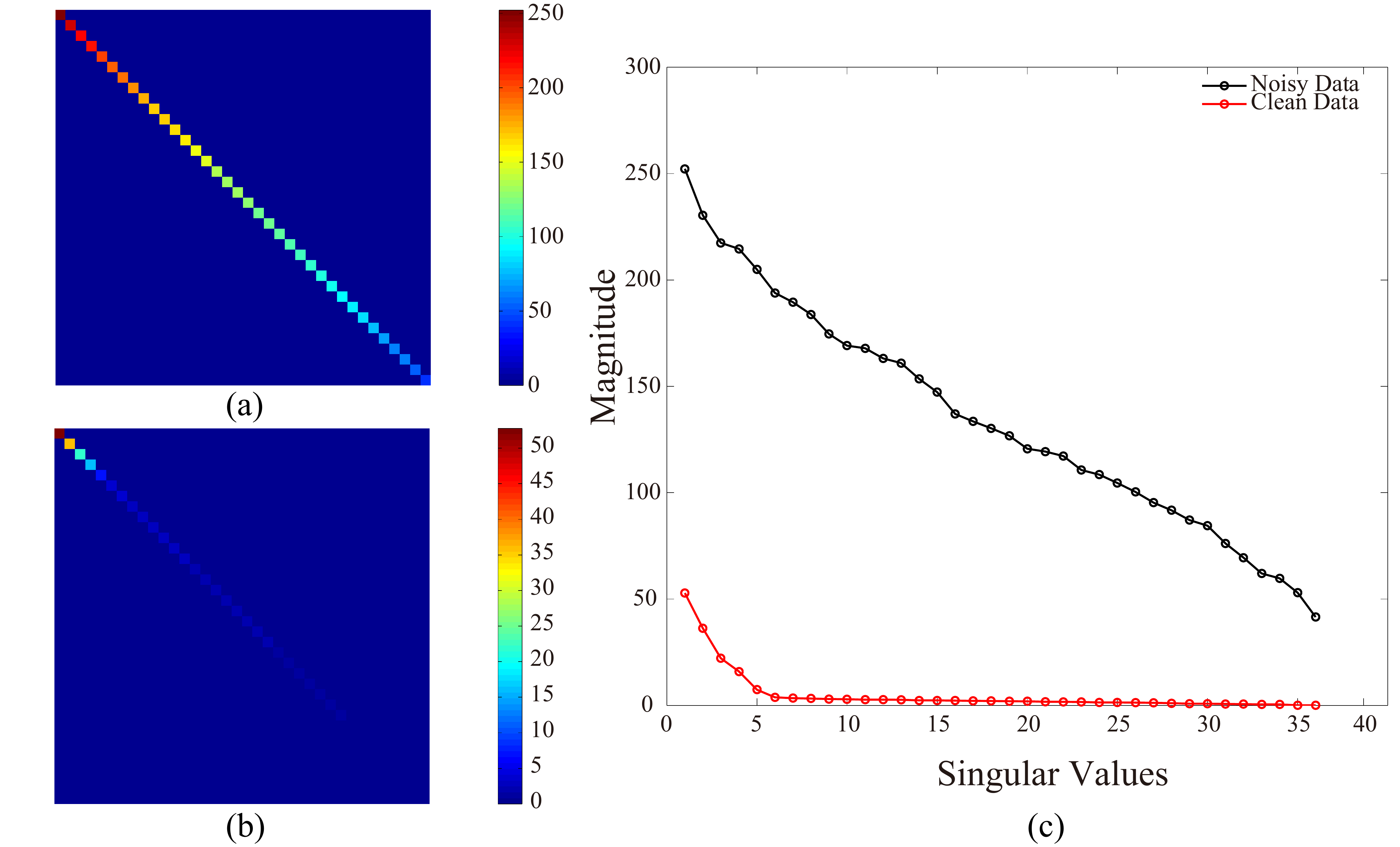}
   \caption{SVD analysis on 2-D matrix on both clean patch and noisy patch. (a) and (b) show the singular values of the noisy patch and clean patch, respectively. (c) Plot of the diagonal element of the singular value matrix.}
\label{Matrix SVD}
\end{figure}

  %%%%%%%%%%%%%%%%%%%%%%%%%%%%%%%%%%%%%%%%%%%%%%%%%%%
  \subsection{Low-rank matrix recovery for 2-D image denoising}
  %%%%%%%%%%%%%%%%%%%%%%%%%%%%%%%%%%%%%%%%%%%%%%%%%%%
    Given a degraded matrix \textbf{\emph{Y}}, low rank matrix recovery aims to recover its underlying low-dimensional structure \textbf{\emph{X}}, which has a wide range of applications in computer vision and machine learning \cite{candes2011robust}. In 2-D image denoising, the degraded matrix $\emph{\textbf{Y}} \in {\mathbb{R}^{p \times (k + 1)}}$ is usually formed by a group of non-local similar vectorization patches, where \emph{p} is the length of the vectorization patches and \emph{k} is the number of the non-local similar patches. Mathematically, the low-rank matrix recovery problem is:
    \begin{equation}\label{eq:LRMR_Original}
    \setlength{\abovedisplayskip}{2pt}
    \setlength{\belowdisplayskip}{2pt}
    \hat{\textbf{\emph{X}}} = \arg \mathop {\min }\limits_\textbf{\emph{X}} \frac{1}{2}||\textbf{\emph{X}} - \textbf{\emph{Y}}||_F^2 + \lambda rank(\textbf{\emph{X}}),
    \end{equation}
    where $\lambda$ is the regularization parameter to balance the terms. Because of the nonconvexity of the rank constraint, the nuclear norm is usually introduced to replace it as its convex surrogate functional \cite{fazel2002matrix}:
    \begin{equation}\label{eq:LRMR_Transformed}
    \setlength{\abovedisplayskip}{2pt}
    \setlength{\belowdisplayskip}{2pt}
    \hat{\textbf{\emph{X}}} = \arg \mathop {\min }\limits_\textbf{\emph{X}} \frac{1}{2}||\textbf{\emph{X}} - \textbf{\emph{Y}}||_F^2 + \lambda ||\textbf{\emph{X}}|{|_ * },
    \end{equation}
    where $||\textbf{\emph{X}}|{|_*}$ is defined as the sum of its singular values, i.e. $||\textbf{\emph{X}}|{|_ * } = {\sum\nolimits_j {|{\sigma _j}(\textbf{\emph{X}})|}_1}$, and ${\sigma _j}(\textbf{\emph{X}})$ means the \emph{j}-th singular value of \textbf{\emph{X}}. In Fig. \ref{Matrix SVD}(a) and  \ref{Matrix SVD}(b), we show the singular values of the noisy patch and clean patch, respectively. We can find that the singular values only exist at the diagonal in 2-D matrix SVD, and the noisy degradation causes the singular values difference. In Fig. \ref{Matrix SVD}(c), we can observe that singular values of the clean patch exist much more sparsity than those of the noisy patch, and follow an exponential decay rule. That is why the low-rank prior, namely the singular value sparsity, is effective in 2-D matrix recovery.

    According to \cite{candes2009exact}, the nuclear norm is the tightest convex relaxation to the non-convex low-rank minimization problem. Under suitable conditions, Eq. (\ref{eq:LRMR_Original}) and (\ref{eq:LRMR_Transformed}) are formally equivalent in the sense that they have exactly the same unique solution. Equation (\ref{eq:LRMR_Transformed}) can be easily solved by a soft singular values thresholding algorithm \cite{cai2010singular}:
    \begin{equation}\label{eq:2D Solution}
    \setlength{\abovedisplayskip}{2pt}
    \setlength{\belowdisplayskip}{2pt}
    \left\{ {\begin{array}{*{20}{c}}
    {\hat{\textbf{\emph{X}}} = \textbf{\emph{U}}\left( {shrink\_{L_ * }(\mathbf{\Sigma} ,\lambda )} \right){\textbf{\emph{V}}^T}}\\
    {shrink\_{L_ * }(\mathbf{\Sigma} ,\lambda ) = diag{{\{\max ({\mathbf{\Sigma} _{jj}} - \lambda ,0)\} }_j},}
    \end{array}} \right.
    \end{equation}
    where $(\textbf{\emph{U}},\mathbf{\Sigma },\textbf{\emph{V}}) = svd(\textbf{\emph{Y}})$ is the singular value decomposition and ${\mathbf{\Sigma} _{jj}}$ is the diagonal element of the singular value matrix ${\mathbf{\Sigma}}$.

  %%%%%%%%%%%%%%%%%%%%%%%%%%%%%%%%%%%%%%%%%%%%%%%%%%%
  \subsection{WLRTR for 3-D HSI denoising}
  %%%%%%%%%%%%%%%%%%%%%%%%%%%%%%%%%%%%%%%%%%%%%%%%%%%
    In this sub-section, we start the presentation of the intuition of the proposed denoising algorithm by first introducing why sparsity of high order singular values is brought to work. Then, we show the detail about how we construct the low-rank 3-order tensor. Once this is set, we present the low-rank tensor recovery model with weighted sparsity.

\begin{figure}
    \includegraphics[width=0.48\textwidth]{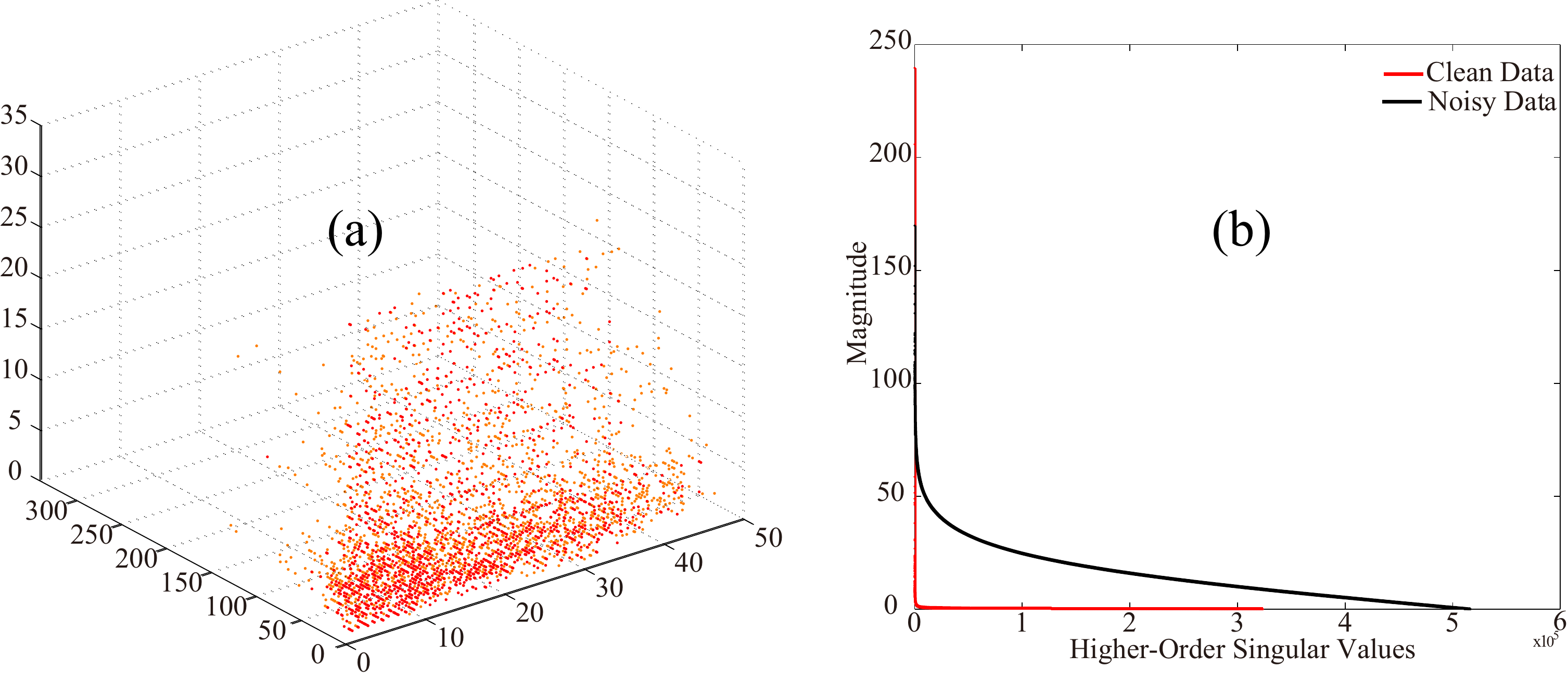}
   \caption{HOSVD analysis on 3-order tensor. (a) Singular values of core tensor bigger than 1. (b) Plot of the singular values of both clean and noisy tensors.}
\label{HOSVD Illustration}
\end{figure}

  %%%%%%%%%%%%%%%%%%%%%%%%%%%%%%%%%%%%%%%%%%%%%%%%%%%
  \subsubsection{Why Low-rank Tensor Recovery}
  %%%%%%%%%%%%%%%%%%%%%%%%%%%%%%%%%%%%%%%%%%%%%%%%%%%

    One major shortcoming of 2-D low-rank is that it can only work in presence of 2-way (matrix) data. However, the real data, such as HSIs and color images, are ubiquitously in three-dimensional way, also referred to as 3-order tensor. To preserve the structural information, we introduce the low-rank tensor recovery model to handle the tensor data by taking the advantage of its multi-dimensional structure.

    In Fig. \ref{Matrix SVD}, we have analyzed the sparsity of the singular values of both the noisy patch and clean patch, namely their 2-D low-rank properties. Analog to this, given a clean 3-order low-rank tensor (we will show how to construct this low-rank tensor in the next section), we apply the HOSVD on it to see how the sparsity of its high order singular value distributes in 3-order tensor, namely its higher order low-rank property. In Fig. \ref{HOSVD Illustration}, we give a visualization for facilitating the understanding of the sparsity in core tensor. There are two observations we make here. First, Figure \ref{HOSVD Illustration}(a) shows the location of singular values in the core tensor, according to their magnitudes. In Fig. \ref{HOSVD Illustration}(a), we can observe that singular value of the core tensor exhibit significant sparsity with different degree along each mode. More specifically, the magnitude of the singular value shows a general descending tendency along the 2-mode and 3-mode. Along the 2-mode, due to the strong redundancy of the non-local cubics, the coefficients in the core tensor along this mode tends to be decreasing very fast to zero. And along the 3-mode, due to the high spectral correlation, the coefficients in the core tensor along this mode tends to decrease to zero with a relative slow speed. Second, from Fig. \ref{HOSVD Illustration}(b), we can observe that singular values of the clean cubic exist much more sparsity than that of those noisy cubic, and follow an extremely sharp exponential decay rule. Moreover, the intrinsic sparsity of the high order singular values of the the 3-D cubic is much more apparent than that of the singular values of the 2-D patch. Therefore, it is natural to use the tensor low-rank model for MSIs recovery problem.

\begin{figure}
\begin{center}
    \includegraphics[width=0.5\textwidth]{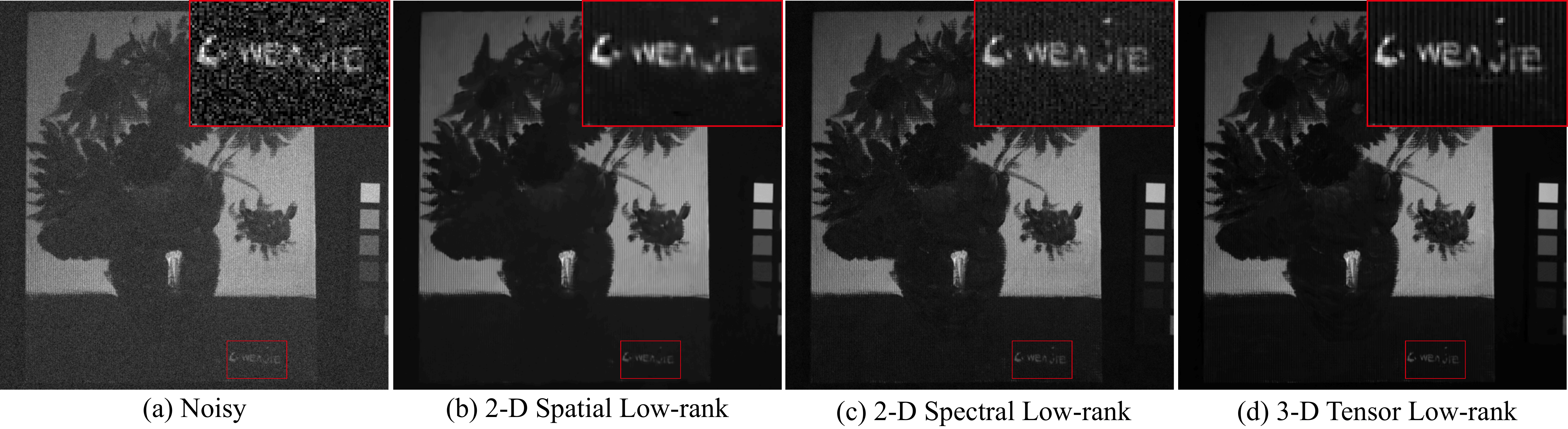}
\end{center}
   \caption{The advantage of the low-rank tensor recovery over low-rank matrix recovery. (a) Simulated noisy image (The original image cubic is with size 512*512*31. Here, we choose band 15 as an example.) under Gaussian noise (sigma=10, PSNR = 28.13dB). (b) 2-D low-rank matrix recovery result via spatial non-local similarity (PSNR = 37.56dB). (c) 2-D low-rank matrix recovery result via spectral correlation (PSNR = 39.18dB). (d) 3-D low-rank tensor recovery result (PSNR = 42.95dB).}
\label{tensor advantage}
\end{figure}

    In Fig. \ref{tensor advantage}, we give a visual comparison between the 3-order cubic recovery and 2-order matrix recovery. Figure \ref{tensor advantage}(b) shows the result of 2-D spatial low-rank recovery result, where the low-rank matrix is formed via spatial non-local similar patches. Figure \ref{tensor advantage}(c) shows the result of 2-D spectral low-rank recovery result, where the low-rank matrix is formed via spectral similar bands. Figure \ref{tensor advantage}(d) shows the result of the proposed low-rank tensor recovery. It can be inferred from the visual appearance and PSNR values that the proposed low-rank tensor recovery method has obvious advantage over low-rank matrix recovery methods in terms of both noise reduction and texture preserving.

  %%%%%%%%%%%%%%%%%%%%%%%%%%%%%%%%%%%%%%%%%%%%%%%%%%%
  \subsubsection{Low-rank Tensor Construction}
  %%%%%%%%%%%%%%%%%%%%%%%%%%%%%%%%%%%%%%%%%%%%%%%%%%%

\begin{figure}
    \includegraphics[width=0.48\textwidth]{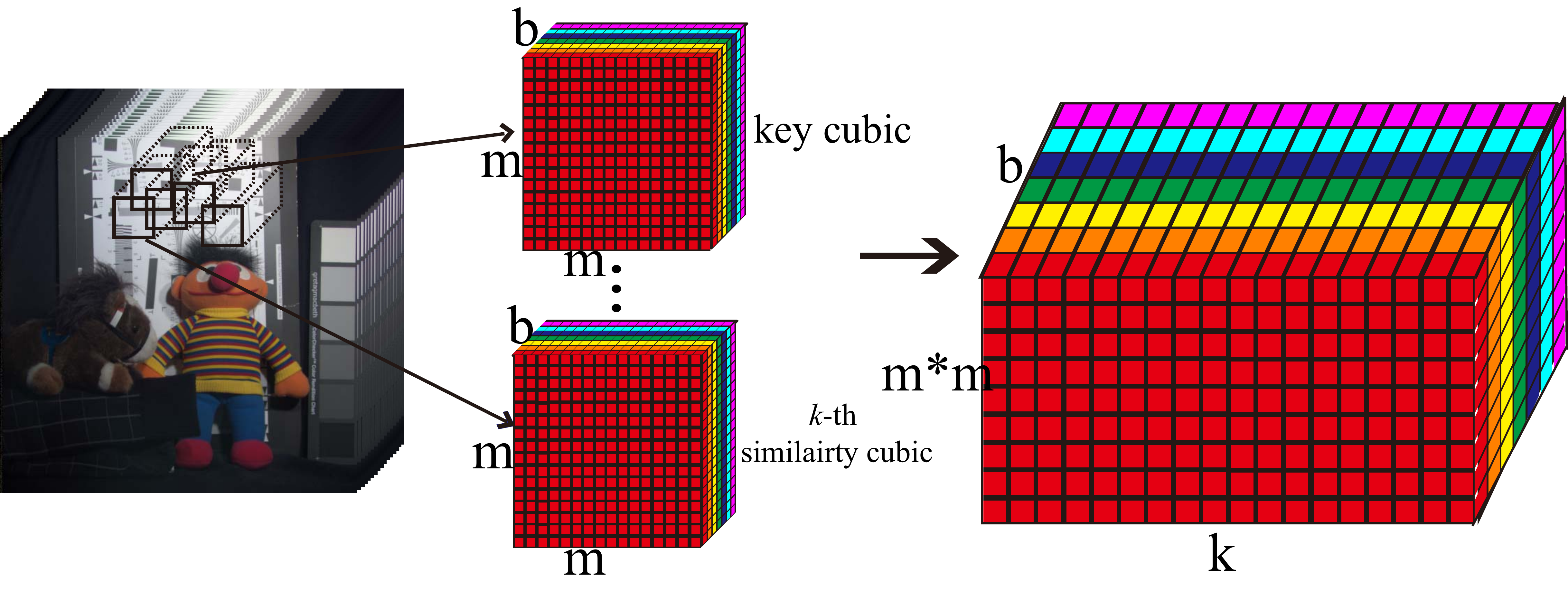}
   \caption{Illustration of the 3-order tensor construction in our work.}
\label{tensor construction}
\end{figure}

    In Fig. \ref{tensor construction}, we give an illustration about how we construct the 3-order tensor as the input for LRTR model. For one key cubic with size $m\times m \times b$ in the whole 3-D images, we search for its \emph{k} nearest neighbors non-local cubics in a local window. Then, the constructed 3-order tensor with the size $m^2 \times (k+1) \times b$ is formed by a group of non-local similar matricization cubics, where $m^2 \times b$ is the size of the matricization cubic and \emph{k} is the number of the non-local similar patches. The constructed 3-order tensor simultaneously utilizes the spatial local sparsity (mode-1), the non-local similarity between spectral-spatial cubic (mode-2) and spectral high correlation (mode-3), which would benefit us from this unified framework. The tensor format offers a unified understanding for the matrix-based recovery model. When $b = 0$ or $k = 0$, the constructed tensor degenerates into a matrix by taking only non-local self-similarity [Fig. \ref{tensor advantage}(b)] or spectral correlation [Fig. \ref{tensor advantage}(c)].

    Although it sounds better to construct a 4-order tensor, expecting better preservation of the spatially horizontal and vertical structural relationship, we found the final denoising result is even slightly degenerating. We infer that the length of the patches in spatial domain is too small (usually significantly small than the number of bands and also the non-local similar cubics) for the HOSVD to exactly extract the intrinsic subspace bases of the spatial information. While for comparative large length of the patches in spatial horizontal and vertical model, we can not afford such huge computational and memory load. Therefore, in this work, we choose to construct the 3-order tensor for recovery.

  %%%%%%%%%%%%%%%%%%%%%%%%%%%%%%%%%%%%%%%%%%%%%%%%%%%
  \subsubsection{Weighted Low-rank Tensor Recovery Model}
  %%%%%%%%%%%%%%%%%%%%%%%%%%%%%%%%%%%%%%%%%%%%%%%%%%%

    For the problem (\ref{eq:Regularization Framework}), the variable splitting technique \cite{lin2011linearized, goldstein2009split} is usually introduced to decouple the data fidelity term and regularization term  by introducing an auxiliary variable ${\boldsymbol{\mathcal{L}}}$. Consequently, for the regularization-related subproblem, it can be regarded as a image denoising problem as follow:
    \begin{equation}\label{eq:Concrete Format}
    \setlength{\abovedisplayskip}{2pt}
    \setlength{\belowdisplayskip}{2pt}
    \Phi ({\boldsymbol{\mathcal{L}}_i}) = {\kern 1pt} \sum\nolimits_i {\frac{1}{{\sigma _i^2}}||{{\mathcal{R}}_i}{\boldsymbol{\mathcal{X}}} - {{\boldsymbol{\mathcal{L}}_i}}||_F^2 + rank({{\boldsymbol{\mathcal{L}}_i}})},
    \end{equation}
    where ${{\mathcal{R}}_i}{\boldsymbol{\mathcal{X}}}$ is the constructed 3-order tensor for each exemplar cubic at location \emph{i}, our goal is to estimate the corresponding low-rank approximation ${\boldsymbol{\mathcal{L}}_i}$ under noise variance $\sigma _i^2$.

    The low-rank regularization has been widely used in matrix recovery, and nuclear norm is usually introduced as the surrogate functional of low-rank constraint. In this work, we borrow this notion in 2-D matrix to define the tensor nuclear norm of ${\boldsymbol{\mathcal{L}}_i}$ as $||{\boldsymbol{\mathcal{L}}_i}|{|_*} = {\sum\nolimits_j{\left| {{\sigma_j}({\boldsymbol{\mathcal{L}}_i})}\right|}_1}$, namely the sum of its high order singular values. Then, the low-rank tensor ${\boldsymbol{\mathcal{L}}_i}$ can be recovered by solving the following optimization problem\footnote{Here, we omit the $\sum\nolimits_i {}$ for simplicity.}:
    \begin{equation}\label{eq:Tensor Nuclear Norm}
    \setlength{\abovedisplayskip}{2pt}
    \setlength{\belowdisplayskip}{2pt}
    {\hat{\boldsymbol{\mathcal{L}}}_i} = \arg \mathop {\min }\limits_{{\boldsymbol{\mathcal{X}}_i}} {\kern 1pt} \frac{1}{{\sigma _i^2}}||{{\mathcal{R}}_i}{\boldsymbol{\mathcal{X}}} - {\boldsymbol{\mathcal{L}}_i}||_F^2 + ||{{\boldsymbol{\mathcal{L}}}_i}|{|_ * }.
    \end{equation}

\begin{figure}
\begin{center}
    \includegraphics[width=0.48\textwidth]{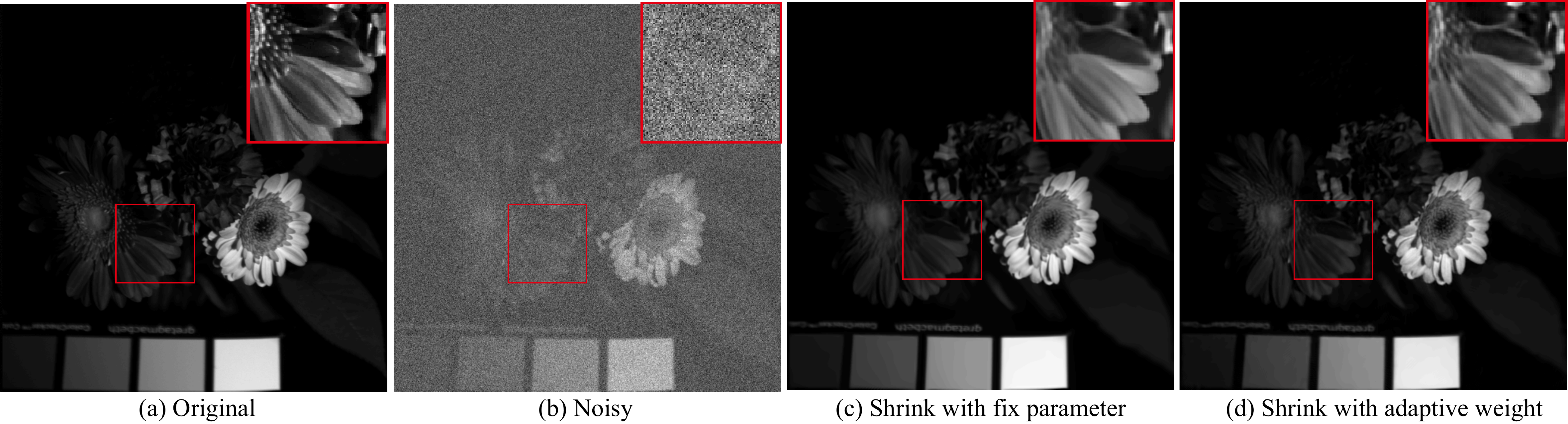}
\end{center}
   \caption{Effectiveness of the adaptive weighted strategy. (a) Original clean image. (b) Simulated noisy image band 15 under Gaussian noise (sigma=30, PSNR = 18.59dB). (c) Denoising result of the proposed method without adaptive weighted strategy (PSNR = 41.27dB). (d) Denoising result of the proposed method (PSNR = 41.85dB).}
\label{weighted efectiveness}
\end{figure}

    However, this tensor nuclear norm has not considered the fine-grained sparsity configurations inside the coefficient tensor. As seen in Fig. \ref{HOSVD Illustration}(b), the singular values of clean tensor ${{\mathcal{R}}_i}{\boldsymbol{\mathcal{X}}}$ exhibit strongly sparsity, in which most of the singular values are close to zero. For the singular values of its corresponding noisy tensor, the larger singular values are close to the singular values of clean tensor, while the small trivial singular values are obviously larger than the singular values of clean tensor. This phenomenon motivates us to penalize the larger singular values less and small singular values more. Thus, we replace the conventional tensor nuclear norm with the weighted one on ${\boldsymbol{\mathcal{L}}_i}$:
    \begin{equation}\label{eq:Weighted Tensor Nuclear Norm}
    \setlength{\abovedisplayskip}{2pt}
    \setlength{\belowdisplayskip}{2pt}
    {\hat{\boldsymbol{\mathcal{L}}}_i} = \arg \mathop {\min }\limits_{{\boldsymbol{\mathcal{X}}_i}} {\kern 1pt} \frac{1}{{\sigma _i^2}}||{{\mathcal{R}}_i}{\boldsymbol{\mathcal{X}}} - {\boldsymbol{\mathcal{L}}_i}||_F^2 + ||{\boldsymbol{\mathcal{L}}_i}|{|_{\emph{\textbf{w}}, * }}.
    \end{equation}
    where $||{\boldsymbol{\mathcal{L}}_i}|{|_{\emph{\textbf{w}}, * }} = {\sum\nolimits_j {\left| {{w_j}{\sigma _j}({\boldsymbol{\mathcal{L}}_i})} \right|} _1}$, $\textbf{\emph{w}} = [{w_1}, \ldots ,{w_n}]$ and ${w_j}$ is a non-negative weight assigned to ${\sigma _j}({\boldsymbol{\mathcal{L}}_i})$. The intuition behind this weighted process are two-folds. For one hand, larger singular values corresponding to the major projection orientations should be penalized less to preserve the major data components. For the other hand, the success of the reweighting strategy, where the regularization parameter is adaptive and inversely proportional to the underlying signal magnitude, has been verified in the various computer vision task \cite{candes2008enhancing, Yan2013Nonlocal}. For a small value after \emph{t} iterations, due to the reweighted process in sparsity constraint, it will enforce a larger reweighting factor in the next \emph{t}+1 iteration, which would naturally result in a sparser result. In this work, we set
    \begin{equation}\label{eq:Weighted Definition}
    \setlength{\abovedisplayskip}{2pt}
    \setlength{\belowdisplayskip}{2pt}
    w_j^{t + 1}{\rm{ = }}{c \mathord{\left/
    {\vphantom {c {\left( {\left| {\sigma _j^t({\cal X})} \right| + \varepsilon } \right)}}} \right.
    \kern-\nulldelimiterspace} {\left( {\left| {\sigma _j^t({\boldsymbol{\mathcal{L}}_i})} \right| + \varepsilon } \right)}},
    \end{equation}
    where $c = 0.04$ is a constant, and $\varepsilon$ is a small constant to avoid dividing by zero. In Fig. \ref{weighted efectiveness}, we give a visual comparison between LRTR model with/without the weighted strategy. The result of Fig. \ref{weighted efectiveness}(d) shows more clear edge than that of without reweighted strategy, which demonstrate the effectiveness of the sparsity reweighted strategy in high order singular values.

  %%%%%%%%%%%%%%%%%%%%%%%%%%%%%%%%%%%%%%%%%%%%%%%%%%%
  \subsubsection{Analytical solution of (\ref{eq:Weighted Tensor Nuclear Norm})}
  %%%%%%%%%%%%%%%%%%%%%%%%%%%%%%%%%%%%%%%%%%%%%%%%%%%

    By replacing ${\boldsymbol{\mathcal{L}}_i}$ in (\ref{eq:Weighted Tensor Nuclear Norm}) with the corresponding HOSVD, we obtain the following problem:
    \begin{equation}\label{eq:Core Tensor Subproblem}
    \begin{aligned}
    \resizebox{0.97\hsize}{!}{${{{\hat{\boldsymbol{\mathcal{S}}}}_i}} = \arg \mathop {\min }\limits_{{{\boldsymbol{\mathcal{S}}}_i}} {\kern 1pt} ||{{\cal R}_i}{\boldsymbol{\mathcal{X}}} - {{\boldsymbol{\mathcal{S}}}_i}{ \times _1}{\textbf{\emph{U}}_1}{ \times _2}{\textbf{\emph{U}}_2}{ \times _3}{\textbf{\emph{U}}_3}||_F^2 + \sigma _i^2||{\textbf{\emph{w}}_i} \circ {{\boldsymbol{\mathcal{S}}}_i}|{|_1}$}.
    \end{aligned}
    \end{equation}
    where $\circ $ denotes the element-wise multiplication. Analog to the matrix case \cite{cai2010singular, Gu2014Weighted}, we can also get its global solution for the non-convex problem (\ref{eq:Core Tensor Subproblem}).

    \noindent
    \textbf{Theorem 1} Given ${{\boldsymbol{\mathcal{X}}}_i},{{\boldsymbol{\mathcal{Y}}}_i} \in {\mathbb{R}^{{I_1} \times {I_2} \times  \cdots  \times {I_N}}}$, let ${{\boldsymbol{\mathcal{Y}}}_i} = {\tilde {\boldsymbol{\mathcal{S}}}_i}{ \times _1}\newline
    {\tilde{\textbf{\emph{U}}}_1}{ \times _2}{\tilde{\textbf{\emph{U}}}_2}{ \times _3}{\tilde{\textbf{\emph{U}}}_3}$ be the HOSVD of ${{\boldsymbol{\mathcal{Y}}}_i} = {{\cal R}_i}{\boldsymbol{\mathcal{X}}}$ and ${{\boldsymbol{\mathcal{X}}}_i} = {\bar {\boldsymbol{\mathcal{S}}}_i}{ \times _1}{\bar{\textbf{\emph{U}}}_1}{ \times _2}{\bar{\textbf{\emph{U}}}_2}{ \times _3}{\bar{\textbf{\emph{U}}}_3}$ be the HOSVD of ${{\boldsymbol{\mathcal{X}}}_i} = {{\boldsymbol{\mathcal{L}}}_i}$ in (\ref{eq:Weighted Tensor Nuclear Norm}). The global optimum of the weighted tensor low-rank minimization problem in (\ref{eq:Core Tensor Subproblem}) can be expressed as ${{\boldsymbol{\mathcal{X}}}_i} = {{\boldsymbol{\mathcal{S}}}_{{\textbf{\emph{w}}_i}\sigma _i^2}} \newline \left( {{{\tilde{\boldsymbol{\mathcal{S}}}}_i}} \right){ \times _1}{\tilde{\textbf{\emph{U}}}_1}{ \times _2}{\tilde{\textbf{\emph{U}}}_2}{ \times _3}{\tilde{\textbf{\emph{U}}}_3}$, where ${\hat{\boldsymbol{\mathcal{S}}}_i} = {{\boldsymbol{\mathcal{S}}}_{{{{\textbf{\emph{w}}_i}\sigma _i^2} \mathord{\left/
    {\vphantom {{{\textbf{\emph{w}}_i}\sigma _i^2} 2}} \right.
    \kern-\nulldelimiterspace} 2}}}\left( {{{\tilde{\boldsymbol{\mathcal{S}}}}_i}} \right) = \max \newline ({\tilde{\boldsymbol{\mathcal{S}}}_i} - {{{\textbf{\emph{w}}_i}\sigma _i^2} \mathord{\left/
    {\vphantom {{{\textbf{\emph{w}}_i}\sigma _i^2} 2}} \right.
    \kern-\nulldelimiterspace} 2},0)$.

    Before proving Theorem 1, we first give two lemmas.

    \noindent
    \textbf{Lemma 1} (\emph{Von Neumann's Trace Inequality for Tensors}) \cite{Chr2015Von} For ${\boldsymbol{\mathcal{X}}},{\boldsymbol{\mathcal{Y}}} \in {\mathbb{R}^{{I_1} \times {I_2} \times  \cdots  \times {I_N}}}$ be tensors. Then for all $\emph{n} = 1, \ldots ,\emph{N}$, we have
    \begin{equation}\label{eq:Tensors Von Neumann's Trace Inequality}
    \begin{aligned}
    \setlength{\abovedisplayskip}{2pt}
    \setlength{\belowdisplayskip}{2pt}
    \left\langle {{\boldsymbol{\mathcal{X}}},{\boldsymbol{\mathcal{Y}}}} \right\rangle  \le \left\langle {{\sigma ^{(n)}}({\boldsymbol{\mathcal{X}}}),{\sigma ^{(n)}}({\boldsymbol{\mathcal{Y}}})} \right\rangle
    \end{aligned}
    \end{equation}
    where ${\sigma ^{(n)}}({\boldsymbol{\mathcal{X}}})$ and ${\boldsymbol{\mathcal{Y}}}$ denote the vectorization of singular values of ${{\boldsymbol{\mathcal{X}}}_{(n)}}$ and ${{\boldsymbol{\mathcal{Y}}}_{(n)}}$, respectively. The equality in (\ref{eq:Tensors Von Neumann's Trace Inequality}) holds simultaneously for all $\emph{n} = 1, \ldots ,\emph{N}$ if and only there exist orthogonal matrices ${{\textbf{\emph{U}}}_n} \in {\mathbb{R}^{{I_n} \times {I_n}}}$ for $\emph{n} = 1, \ldots ,\emph{N}$ and tensors ${\boldsymbol{\mathcal{S}}},\hat{\boldsymbol{\mathcal{S}}} \in {\mathbb{R}^{{I_1} \times {I_2} \times  \cdots  \times {I_N}}}$ such that $$ \setlength{\abovedisplayskip}{2pt}
    \setlength{\belowdisplayskip}{2pt}
    {\boldsymbol{\mathcal{X}}} = {\boldsymbol{\mathcal{S}}}{ \times _1}{\textbf{\emph{U}}_1}{ \times _2}{\textbf{\emph{U}}_2}{ \times _3} \cdots { \times _N}{\textbf{\emph{U}}_N},$$
    $$\setlength{\abovedisplayskip}{2pt}
    \setlength{\belowdisplayskip}{2pt}
    {\boldsymbol{\mathcal{Y}}} = \hat{\boldsymbol{\mathcal{S}}}{ \times _1}{\textbf{\emph{U}}_1}{ \times _2}{\textbf{\emph{U}}_2}{ \times _3} \cdots { \times _N}{\textbf{\emph{U}}_N},$$
    where ${\boldsymbol{\mathcal{S}}}$ and $\hat{\boldsymbol{\mathcal{S}}}$ satisfy the following properties:
    \begin{itemize}
      \item ${\boldsymbol{\mathcal{S}}}$ and $\hat{\boldsymbol{\mathcal{S}}}$ are block-wise diagonal with the same number and size of blocks. The tensor called block-wise diagonal if in the block representation above all off-diagonal blocks are zero blocks.
      \item Let \emph{L} be the number of blocks and ${\left\{{{{\boldsymbol{\mathcal{S}}}_{(l)}}} \right\}_{l = 1, \ldots ,\emph{L}}}$ be the blocks on the diagonal of ${\boldsymbol{\mathcal{S}}}$. Then for each $\emph{l} = 1, \ldots ,\emph{L}$, the two blocks ${{\boldsymbol{\mathcal{S}}}_{(l)}}$ and ${\hat{\boldsymbol{\mathcal{S}}}_{(l)}}$ are proportional.
    \end{itemize}

    \noindent
    \textbf{Lemma 2} (\emph{Frobenius norm unitarily invariant}) \cite{Lathauwer2000A} let the HOSVD of ${\boldsymbol{\mathcal{X}}}$ be given as in Definition 4; then the following holds:
    \begin{equation}\label{eq:Frobenius norm unitarily invariant}
    \begin{aligned}
    \setlength{\abovedisplayskip}{2pt}
    \setlength{\belowdisplayskip}{2pt}
    \resizebox{0.87\hsize}{!}{$||{\boldsymbol{\mathcal{X}}}||_F^2 = \sum\limits_{j = 1}^{{R_1}} {{{\left( {\sigma _j^{(1)}({\boldsymbol{\mathcal{X}}})} \right)}^2}}  =  \cdots  = \sum\limits_{j = 1}^{{R_N}} {{{\left( {\sigma _j^{(N)}({\boldsymbol{\mathcal{X}}})} \right)}^2}}  = ||{\boldsymbol{\mathcal{S}}}||_F^2$},
    \end{aligned}
    \end{equation}

    Proof of Theorem 1. Based on lemma 2, the following derivations hold:
    $$\begin{array}{l}
    {\kern 1pt} {\kern 1pt} {\kern 1pt} {\kern 1pt} {\kern 1pt} {\kern 1pt} {\kern 1pt} {\kern 1pt} {\kern 1pt} {\kern 1pt} {\kern 1pt} |{\kern 1pt} |{{\boldsymbol{\mathcal{Y}}}_i} - {{\boldsymbol{\mathcal{X}}}_i}||_F^2 + \sigma _i^2||{\textbf{\emph{w}}_i} \circ {{\bar{\boldsymbol{\mathcal{S}}}}_i}|{|_1}\\
    = \left\langle {{{\boldsymbol{\mathcal{Y}}}_i},{{\boldsymbol{\mathcal{Y}}}_i}} \right\rangle  - 2\left\langle {{{\boldsymbol{\mathcal{X}}}_i},{{\boldsymbol{\mathcal{Y}}}_i}} \right\rangle  + \left\langle {{{\boldsymbol{\mathcal{X}}}_i},{{\boldsymbol{\mathcal{X}}}_i}} \right\rangle  + \sigma _i^2||{\textbf{\emph{w}}_i} \circ {{\bar{\boldsymbol{\mathcal{S}}}}_i}|{|_1}\\
    = \left\langle {{{\tilde{\boldsymbol{\mathcal{S}}}}_i},{{\tilde{\boldsymbol{\mathcal{S}}}}_i}} \right\rangle - 2\left\langle {{{\boldsymbol{\mathcal{X}}}_i},{{\boldsymbol{\mathcal{Y}}}_i}} \right\rangle  + \left\langle {{{\bar{\boldsymbol{\mathcal{S}}}}_i},{{\bar{\boldsymbol{\mathcal{S}}}}_i}} \right\rangle  + \sigma _i^2||{\textbf{\emph{w}}_i} \circ {{\bar{\boldsymbol{\mathcal{S}}}}_i}|{|_1}.
    \end{array}$$
    Based on the von Neumann trace inequality for tensors in Lemma 1, we know that $\left\langle {{{\boldsymbol{\mathcal{X}}}_i},{{\boldsymbol{\mathcal{Y}}}_i}} \right\rangle$ achieves its upper bound if  ${\bar{\textbf{\emph{U}}}_1} = {\tilde{\textbf{\emph{U}}}_1}$, ${\bar{\textbf{\emph{U}}}_2} = {\tilde{\textbf{\emph{U}}}_2}$ and ${\bar{\textbf{\emph{U}}}_3} = {\tilde{\textbf{\emph{U}}}_3}$. Thus (\ref{eq:Core Tensor Subproblem}) is equivalent to
    \begin{equation}\label{eq:Convex Problem}
    \begin{aligned}
    {{{\hat{\boldsymbol{\mathcal{S}}}}_i}}= \arg \mathop {\min }\limits_{{{\bar{\boldsymbol{\mathcal{S}}}}_i}} {\kern 1pt} |{\kern 1pt} |{\tilde{\boldsymbol{\mathcal{S}}}_i} - {\bar{\boldsymbol{\mathcal{S}}}_i}||_F^2 + \sigma _i^2||{\textbf{\emph{w}}_i} \circ {\bar{\boldsymbol{\mathcal{S}}}_i}|{|_1}.
    \end{aligned}
    \end{equation}

    Assume each coefficient in the core tensor as ${\tilde{s}_{ijk}}$ and ${\bar{s}_{ijk}}$, respectively. Thus, the problem (\ref{eq:Convex Problem}) can convert into the scalar format:
    $$\begin{array}{l}
    {\kern 1pt} {\kern 1pt} {\kern 1pt} {\kern 1pt} {\kern 1pt} {\kern 1pt} {\kern 1pt} {\kern 1pt} {\kern 1pt} {\kern 1pt} {\kern 1pt} {\kern 1pt} {\kern 1pt} {\kern 1pt} \mathop {\min }\limits_{{{\bar s}_{ijk}}} {\kern 1pt} {\left( {{{\tilde s}_{ijk}} - {{\bar s}_{ijk}}} \right)^2} + \sigma _i^2{w_i}{{\bar s}_{ijk}}\\
    \Leftrightarrow {\kern 1pt} {\kern 1pt} {\kern 1pt} \mathop {\min }\limits_{{{\bar s}_{ijk}}} {\kern 1pt} {\left( {{{\bar s}_{ijk}} - \left( {{{\tilde s}_{ijk}} - \frac{{\sigma _i^2{w_i}}}{2}} \right)} \right)^2}.
    \end{array}$$
    It is easy to derive its global optimum as:
    $${\bar s_{ijk}} = \max \left( {{{\tilde s}_{ijk}} - \frac{{\sigma _i^2{w_i}}}{2},0} \right).$$
    The proof is completed. Theorem 1 shows that the problem (\ref{eq:Core Tensor Subproblem}) can be solved via the singular value thresholding method.

%%%%%%%%%%%%%%%%%%%%%
\section{HSI Restoration With WLRTR Model}
%%%%%%%%%%%%%%%%%%%%%
   In this section, we show the concrete objective functional and its optimization procedure of each HSI restoration task.

  %%%%%%%%%%%%%%%%%%%%%%%%%%%%%%%%%%%%%%%%%%%%%%%%%%%
  \subsection{WLRTR for HSI denoising}
  %%%%%%%%%%%%%%%%%%%%%%%%%%%%%%%%%%%%%%%%%%%%%%%%%%%
    For HSI denoising, we only consider the random noise ${\boldsymbol{\mathcal{N}}}$ with an identity tensor operator. Thus, by combining the data fidelity term $\frac{1}{2}||{\boldsymbol{\mathcal{Y}}} - {\boldsymbol{\mathcal{X}}}||_F^2$ with the WLRTR prior, the Eq. (\ref{eq:Regularization Framework}) can be formulated as the following minimization problem:
    \begin{equation}\label{eq:WLRTR}
    \begin{aligned}
    \setlength{\abovedisplayskip}{2pt}
    \setlength{\belowdisplayskip}{2pt}
    &\resizebox{0.55\hsize}{!}{$\left\{ {\hat{\boldsymbol{\mathcal{X}}},{{\hat{\boldsymbol{\mathcal{S}}}}_i}} \right\} = \arg \mathop {\min }\limits_{{\boldsymbol{\mathcal{X}}},{{\boldsymbol{\mathcal{S}}}_i}} {\kern 1pt} \frac{1}{2}{\rm{||}}{\boldsymbol{\mathcal{Y}}} - {\boldsymbol{\mathcal{X}}}{\rm{||}}_F^2 + $} \\
    &\resizebox{0.87\hsize}{!}{$ \eta \sum\nolimits_i {\left( {||{{\mathcal{R}}_i}{\boldsymbol{\mathcal{X}}} - {{\boldsymbol{\mathcal{S}}}_i}{ \times _1}{\textbf{\emph{U}}_1}{ \times _2}{\textbf{\emph{U}}_2}{ \times _3}{\textbf{\emph{U}}_3}||_F^2 + \sigma _i^2||{\textbf{\emph{w}}_i} \circ {{\boldsymbol{\mathcal{S}}}_i}|{|_1}} \right)}$}.
    \end{aligned}
    \end{equation}

    The alternating minimization strategy is introduced to solve (\ref{eq:WLRTR}). The ${{\boldsymbol{\mathcal{S}}}_i}$-related subproblem (\ref{eq:Weighted Tensor Nuclear Norm}) can be solved via Theorem 1. After solving for each ${{\boldsymbol{\mathcal{S}}}_i}$, we can reconstruct the whole image ${{\boldsymbol{\mathcal{X}}}}$ by solving the following sub-problem:
    \begin{equation}\label{eq:Image Resotration}
    \begin{aligned}
    \resizebox{0.86\hsize}{!}{${\hat{\boldsymbol{\mathcal{X}}}} = \arg \mathop {\min }\limits_{\boldsymbol{\mathcal{X}}} {\kern 1pt} \frac{1}{2}{\rm{||}}{\boldsymbol{\mathcal{Y}}} - {\boldsymbol{\mathcal{X}}}{\rm{||}}_F^2 + \eta \sum\nolimits_i {||{{\cal R}_i}{\boldsymbol{\mathcal{X}}} - {{\hat{\boldsymbol{\mathcal{S}}}}_i}{ \times _1}{{\tilde{\textbf{\emph{U}}}}_1}{ \times _2}{{\tilde{\textbf{\emph{U}}}}_2}{ \times _3}{{\tilde{\textbf{\emph{U}}}}_3}||_F^2}$}.
    \end{aligned}
     \end{equation}
    Eq. (\ref{eq:Image Resotration}) is a quadratic optimization problem admitting a closed-form solution:
    \begin{equation}\label{eq:Solution of Whole Image}
    \begin{aligned}
    \resizebox{0.85\hsize}{!}{$\hat{\boldsymbol{\mathcal{X}}} = {(I + \eta \sum\nolimits_i {{\cal R}_i^T{{\cal R}_i}} )^{ - 1}}({\boldsymbol{\mathcal{Y}}} + \eta \sum\nolimits_i {({\cal R}_i^T{{\hat{\boldsymbol{\mathcal{S}}}}_i}){ \times _1}{{\tilde{\textbf{\emph{U}}}}_1}{ \times _2}{{\tilde{\textbf{\emph{U}}}}_2}{ \times _3}{{\tilde{\textbf{\emph{U}}}}_3}} )$},
    \end{aligned}
    \end{equation}
    where $\eta \sum\nolimits_i {{\cal R}_i^T{{\cal R}_i}}$ denotes the number of overlapping cubics that cover the pixel location, and $\eta \sum\nolimits_i {({\cal R}_i^T{{\hat{\boldsymbol{\mathcal{S}}}}_i}){ \times _1}{{\tilde{\textbf{\emph{U}}}}_1}{ \times _2}{{\tilde{\textbf{\emph{U}}}}_2}{ \times _3}{{\tilde{\textbf{\emph{U}}}}_3}}$ means the sum value of all overlapping reconstruction cubics that cover the pixel location. Eq. (\ref{eq:Solution of Whole Image}) can be computed in tensor format efficiently.

    After obtaining an improved estimate of the unknown image, the low-rank tensors ${\hat{\boldsymbol{\mathcal{S}}}_i}$ can be updated by Eq. (\ref{eq:Core Tensor Subproblem}). The updated ${\hat{\boldsymbol{\mathcal{S}}}_i}$ is fed back to Eq. (\ref{eq:Solution of Whole Image}) improving the estimate of $\hat{\boldsymbol{\mathcal{X}}}$. Such process is iterated until the convergence. The overall procedure is summarized in \textbf{Algorithm 1}.

\begin{algorithm}[tbp]
\caption{WLRTR for HSI denoising}
     \algorithmicrequire{ Noisy image ${\boldsymbol{\mathcal{Y}}}$}
     \begin{algorithmic}[1]\label{alg:algo1}
         \Procedure{Denoising}{}
     \State{\textbf{Initialize:} Set parameters ${\eta}$; ${\boldsymbol{\mathcal{X}}}^{(1)} = {\boldsymbol{\mathcal{Y}}}$;}
         \For{\emph{n}=1:\emph{N}}
          \State{Low-rank tensor construction:similar cubics grouping};
                \For{(Low-rank tensor approximation) \emph{i}=1:\emph{I} }
                   \State{ Update the thresholds using Eq. (\ref{eq:Weighted Definition})};
                   \State{ Solve Eq. (\ref{eq:Core Tensor Subproblem}) for ${\boldsymbol{\mathcal{S}}}_i$};
                \EndFor
           \State{\textbf{end for}}
           \State{Reconstruct the whole image ${\boldsymbol{\mathcal{X}}}$ from {${\boldsymbol{\mathcal{S}}}_i$} via Eq.(\ref{eq:Solution of Whole Image})}.
          \EndFor
          \State{\textbf{end for}}
          \EndProcedure
     \end{algorithmic}
     \algorithmicensure{ Clean image ${\boldsymbol{\mathcal{X}}}$}
\end{algorithm}

\begin{figure}
    \includegraphics[width=0.48\textwidth]{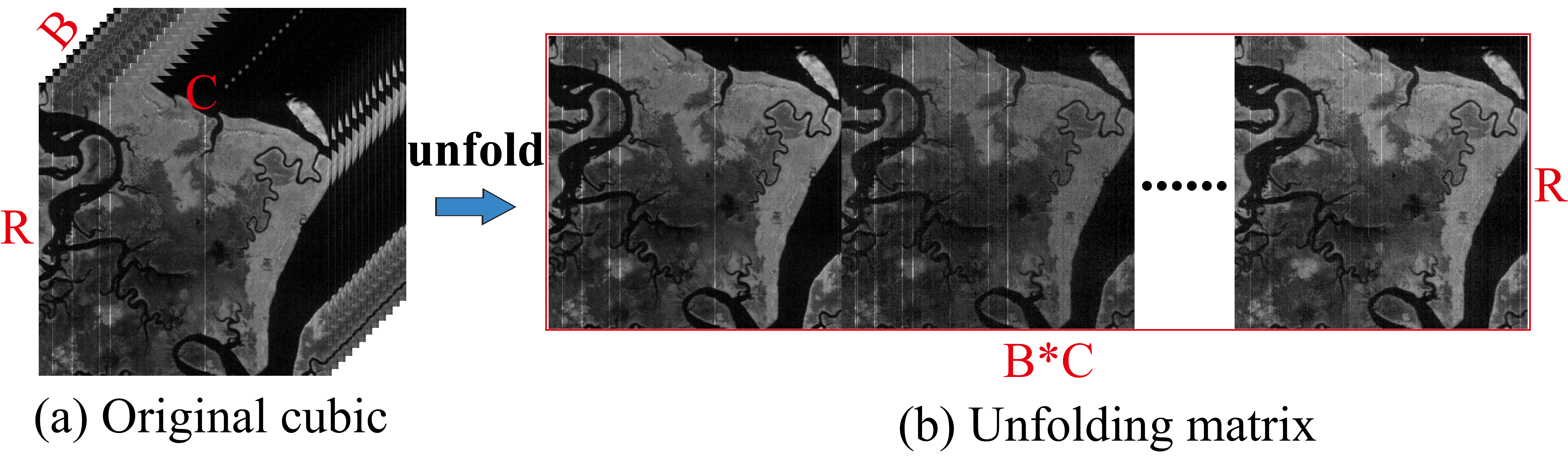}
   \caption{Illustration of HSI stripe mode-1 unfolding. (a) Original HSI cubic with vertical stripe; (b) Mode-1 unfolding matrix.}
\label{Stripe unfolding}
\end{figure}

  %%%%%%%%%%%%%%%%%%%%%%%%%%%%%%%%%%%%%%%%%%%%%%%%%%%
  \subsection{WLRTR-RPCA for HSI destriping}
  %%%%%%%%%%%%%%%%%%%%%%%%%%%%%%%%%%%%%%%%%%%%%%%%%%%

    In real HSI, there always exists system structural noise such as stripe noise. The stripes in HSIs via push-broom imaging spectrometer are always non-periodic, and arise from the unstable detectors during a scanning cycle. Therefore, it is natural for us to borrow the RPCA model \cite{wright2009robust} to accommodate the sparse error component, mainly the stripe noise ${\boldsymbol{\mathcal{E}}}$. The RPCA has shown its robustness in presence of the sparse error, such as background subtraction \cite{gu2016weighted}, structural noise removal \cite{chang2016remote}, face recognition under occlusion \cite{chen2012low}, since it has taken the sparse error into consideration.

    In this section, we extend the WLRTR to the WLRTR-RPCA for stripe noise removal. For the image prior, we will utilize the weighted low-rank tensor prior to model them. While for the stripe noise with obviously directional characteristic, as shown in Fig. \ref{Stripe unfolding}, we argue the ${{{L}}_{2,1}}$-norm with direction discriminative ability is more appropriate than ${{{L}}_{1}}$-norm. Since ${{{L}}_{2,1}}$-norm \resizebox{0.45\hsize}{!}{ $||\textbf{\emph{E}}|{|_{2,1}} = \sum\nolimits_{j = 1}^C {\sqrt {\sum\nolimits_{i = 1}^R {{{({E_{ij}})}^2}} } }$} encourages the intensity of columns to be zero, the underlying assumption here is that the corruptions are sample-specific, i.e., some data vectors are corrupted and the others are clean, just corresponding to the broken and intact detectors, respectively.

    In this work, we extend the matrix ${{{L}}_{2,1}}$-norm to its 3-order tensor case \resizebox{0.55\hsize}{!}{$||{\boldsymbol{\mathcal{E}}}|{|_{2,1,1}} = \sum\nolimits_{k = 1}^B {\sum\nolimits_{j = 1}^C {\sqrt {\sum\nolimits_{i = 1}^R {{{({E_{ijk}})}^2}} } } }$}, and incorporate it into the WLRTR model as follow:
    \begin{equation}\label{eq:LRTR-RPCA}
    \begin{aligned}
    \begin{array}{l}
    \setlength{\abovedisplayskip}{2pt}
    \setlength{\belowdisplayskip}{2pt}
    \resizebox{0.83\hsize}{!}{$\left\{ {\hat{\boldsymbol{\mathcal{X}}},{{\hat{\boldsymbol{\mathcal{S}}}}_i},\hat{\boldsymbol{\mathcal{E}}}} \right\} = \arg \mathop {\min }\limits_{{\boldsymbol{\mathcal{X}}},{{\boldsymbol{\mathcal{S}}}_i},{\boldsymbol{\mathcal{E}}}} {\kern 1pt} \frac{1}{2}{\rm{||}}{\boldsymbol{\mathcal{Y}}} - {\boldsymbol{\mathcal{X}}} - {\boldsymbol{\mathcal{E}}}{\rm{||}}_F^2 + \rho ||{\boldsymbol{\mathcal{E}}}|{|_{2,1,1}}+$}\\
    \resizebox{0.83\hsize}{!}{$\eta \sum\nolimits_i {\left( {||{{\cal R}_i}{\boldsymbol{\mathcal{X}}} - {{\boldsymbol{\mathcal{S}}}_i}{ \times _1}{\textbf{\emph{U}}_1}{ \times _2}{\textbf{\emph{U}}_2}{ \times _3}{\textbf{\emph{U}}_3}||_F^2 + \sigma _i^2||{\textbf{\emph{w}}_i} \circ {{\boldsymbol{\mathcal{S}}}_i}|{|_1}} \right)}$} ,
    \end{array}
    \end{aligned}
    \end{equation}
    where $\rho$ and $\eta$ are the regularization parameters for balancing each term. The WLRTR-RPCA model (\ref{eq:LRTR-RPCA}) is simple and easy to understand, in which the local sparsity, non-local similarity, and spectral consistency of the images are utilized via the tensor low-rank prior, whereas the stripe noise are well depicted by the ${{{L}}_{2,1}}$-norm, so that the mixed random and stripe noise can be separating from the images satisfactorily.

    The procedure of estimation ${\hat{\boldsymbol{\mathcal{S}}}_i}$ and ${\boldsymbol{\mathcal{X}}}$ is the same as the \textbf{Algorithm 1}. Here, we show how we estimate ${\boldsymbol{\mathcal{E}}}$. Once ${\hat{\boldsymbol{\mathcal{S}}}_i}$ and ${\boldsymbol{\mathcal{X}}}$ has been estimated, we can estimate ${\boldsymbol{\mathcal{E}}}$ by solving the following sub-problem:
    \begin{equation}\label{eq:Sparse error subproblem}
    \begin{aligned}
    \setlength{\abovedisplayskip}{2pt}
    \setlength{\belowdisplayskip}{2pt}
    \hat{\boldsymbol{\mathcal{S}}} = \arg \mathop {\min }\limits_{\boldsymbol{\mathcal{E}}} {\kern 1pt} \frac{1}{2}{\rm{||}}{\boldsymbol{\mathcal{Y}}} - {\boldsymbol{\mathcal{X}}} - {\boldsymbol{\mathcal{E}}}{\rm{||}}_F^2 + \rho ||{\boldsymbol{\mathcal{E}}}|{|_{2,1,1}},
    \end{aligned}
    \end{equation}
    It is hard to directly obtain the final result. However, we have the following lemma:
    \noindent
    \textbf{Lemma 3} \cite{liu2013robust}: Let $\textbf{\emph{Q}} = [{\textbf{\emph{q}}_1},{\textbf{\emph{q}}_2}, \cdots ,{\textbf{\emph{q}}_i}, \cdots ]$ be a given matrix and ${\left\| \bullet \right\|_F}$ be the Frobenius norm. If $\hat{\textbf{\emph{W}}}$ is the optimal solution of
    $$\hat{\textbf{\emph{W}}} = \arg \mathop {\min }\limits_\textbf{\emph{W}} {\kern 1pt} \frac{1}{2}\left\| {\textbf{\emph{W}} - \textbf{\emph{Q}}} \right\|_F^2 + \mu {\left\| \textbf{\emph{W}} \right\|_{2,1}}$$
    then the \emph{i}-th column of $\hat{\textbf{\emph{W}}}$ is
    $$\hat{\textbf{\emph{W}}}(:,i) = \left\{ {\begin{array}{*{20}{c}}
    {\frac{{\left\| {{\textbf{\emph{q}}_i}} \right\| - \mu }}{{\left\| {{\textbf{\emph{q}}_i}} \right\|}}{\textbf{\emph{q}}_i},}&{if\mu  \le \left\| {{\textbf{\emph{q}}_i}} \right\|,}\\
    {0,}&{otherwise.}
    \end{array}} \right.$$

    Thus, it is natural for us to unfold the tensors into the matrix (tensor matricization) so that we can apply Lemma 3 directly. By unfolding of the tensors into their mode-1, (\ref{eq:Sparse error subproblem}) is converted into the equivalent problem
    \begin{equation}\label{eq:Sparse error subproblem2}
    \begin{aligned}
    {\hat{\textbf{\emph{E}}}_{(1)}} = \arg \mathop {\min }\limits_{{\textbf{\emph{E}}_{(1)}}} {\kern 1pt} {\rm{||}}{\textbf{\emph{Y}}_{(1)}} - {\textbf{\emph{X}}_{(1)}} - {\textbf{\emph{E}}_{(1)}}{\rm{||}}_F^2 + \rho ||{\textbf{\emph{E}}_{(1)}}|{|_{2,1}},
    \end{aligned}
    \end{equation}

    The Eq. (\ref{eq:Sparse error subproblem2}) can be solved efficiently via Lemma 3. It is worth noting that we chose the mode-1 unfolding since only in this way the resulting matrix still preserve the directional characteristic [Fig. \ref{Stripe unfolding}(b)], while mode-2 and mode-3 unfolding may lose this property. After we obtain the sparse error matrix ${\hat{\textbf{\emph{E}}}_{(1)}}$, we fold it into tensor format. The overall procedure is summarized in \textbf{Algorithm 2}.

\begin{algorithm}[tbp]
\caption{WLRTR-RPCA for HSI destriping}
\algorithmicrequire { Noisy image ${\boldsymbol{\mathcal{Y}}}$}
     \begin{algorithmic}[1]\label{alg:algo2}
     \Procedure{Destriping}{}
        \State{\textbf{Initialize:} Set parameters ${\eta}$ and ${\rho}$; ${\boldsymbol{\mathcal{X}}}^{(1)} = {\boldsymbol{\mathcal{Y}}}$;}
         \For{\emph{n}=1:\emph{N}}
           \State{Step 1: Update sparse error ${\boldsymbol{\mathcal{E}}}$ by solving (\ref{eq:Sparse error subproblem})};
           \State{Step 2: Update clean image ${\boldsymbol{\mathcal{X}}}$ by \textbf{Algorithm 1}.}
          \EndFor
          \State{\textbf{end for}}
          \EndProcedure
     \end{algorithmic}
\algorithmicensure{ Clean image ${\boldsymbol{\mathcal{X}}}$ and stripe component ${\boldsymbol{\mathcal{E}}}$}.
\end{algorithm}

  %%%%%%%%%%%%%%%%%%%%%%%%%%%%%%%%%%%%%%%%%%%%%%%%%%%
  \subsection{WLRTR-RPCA for HSI deblurring}
  %%%%%%%%%%%%%%%%%%%%%%%%%%%%%%%%%%%%%%%%%%%%%%%%%%%

    For HSI deblurring, we only consider the random noise ${\boldsymbol{\mathcal{N}}}$ with the blurring operator. Thus, by combining the data fidelity term $\frac{1}{2}||{\boldsymbol{\mathcal{Y}}} - {\mathcal{T}}({\boldsymbol{\mathcal{X}}})||_F^2$ with the WLRTR prior, the Eq. (\ref{eq:Regularization Framework}) can be formulated as the following minimization problem:
    \begin{equation}\label{eq:deblurring-WLRTR}
    \begin{aligned}
    \setlength{\abovedisplayskip}{2pt}
    \setlength{\belowdisplayskip}{2pt}
    &\resizebox{0.65\hsize}{!}{$\left\{ {\hat{\boldsymbol{\mathcal{X}}},{{\hat{\boldsymbol{\mathcal{S}}}}_i}} \right\} = \arg \mathop {\min }\limits_{{\boldsymbol{\mathcal{X}}},{{\boldsymbol{\mathcal{S}}}_i}} {\kern 1pt} \frac{1}{2}{\rm{||}}{\boldsymbol{\mathcal{Y}}} - {\boldsymbol{\mathcal{X}}} \otimes {\boldsymbol{\mathcal{H}}} {\rm{||}}_F^2 + $} \\
    &\resizebox{0.87\hsize}{!}{$ \eta \sum\nolimits_i {\left( {||{{\mathcal{R}}_i}{\boldsymbol{\mathcal{X}}} - {{\boldsymbol{\mathcal{S}}}_i}{ \times _1}{\textbf{\emph{U}}_1}{ \times _2}{\textbf{\emph{U}}_2}{ \times _3}{\textbf{\emph{U}}_3}||_F^2 + \sigma _i^2||{\textbf{\emph{w}}_i} \circ {{\boldsymbol{\mathcal{S}}}_i}|{|_1}} \right)}$},
    \end{aligned}
    \end{equation}
    where $\otimes$ denotes the convolution operator, and ${\boldsymbol{\mathcal{H}}}$ is a linear shift-invariant point spread function (PSF). Here, we do not take the stripe noise component ${\boldsymbol{\mathcal{E}}}$ into consideration. Joint destriping and deblurring for HSI is another much harder problem, which is out of the scope of this work. For the problem (\ref{eq:deblurring-WLRTR}), we employ the alternative direction multiplier method (ADMM) \cite{lin2011linearized} by introducing auxiliary variable ${\boldsymbol{\mathcal{A}}} = {\boldsymbol{\mathcal{X}}}$ to decouple the fidelity from regularization term:
    \begin{subequations}\label{eq:deblurring-WLRTR-ADMM}
    \begin{align}
    \setlength{\abovedisplayskip}{2pt}
    \setlength{\belowdisplayskip}{2pt}
    \label{eq:a}
    &\resizebox{0.68\hsize}{!}{${\hat {\boldsymbol{\mathcal{A}}} = \arg \mathop {\min }\limits_{\boldsymbol{\mathcal{A}}} \frac{1}{2}{\kern 1pt} {\rm{||}}{\boldsymbol{\mathcal{Y}}} - {\boldsymbol{\mathcal{A}}} \otimes {\boldsymbol{\mathcal{H}}}{\rm{||}}_F^2 + \frac{\alpha }{2}||{\boldsymbol{\mathcal{A}}} - {\boldsymbol{\mathcal{X}}} - \frac{\boldsymbol{\mathcal{J}}}{\alpha }||_F^2}$}\\
    \label{eq:b}
    &\resizebox{0.87\hsize}{!}{${\hat {\boldsymbol{\mathcal{X}}} = \arg \mathop {\min }\limits_{\boldsymbol{\mathcal{X}}} \frac{\alpha }{2}||{\boldsymbol{\mathcal{A}}} - {\boldsymbol{\mathcal{X}}} - \frac{\boldsymbol{\mathcal{J}}}{\alpha }||_F^2 + \eta \sum\nolimits_i {||{{\mathcal{R}}_i}{\boldsymbol{\mathcal{X}}} - {{\boldsymbol{\mathcal{S}}}_i}{ \times _1}{\textbf{\emph{U}}_1}{ \times _2}{\textbf{\emph{U}}_2}{ \times _3}{\textbf{\emph{U}}_3}||_F^2} }$}\\
    \label{eq:c}
    &\resizebox{0.8\hsize}{!}{${{{\hat {\boldsymbol{\mathcal{S}}}}_i} = \arg \mathop {\min }\limits_{{{\boldsymbol{\mathcal{S}}}_i}} ||{{\mathcal{R}}_i}{\boldsymbol{\mathcal{X}}} - {{\boldsymbol{\mathcal{S}}}_i}{ \times _1}{\textbf{\emph{U}}_1}{ \times _2}{\textbf{\emph{U}}_2}{ \times _3}{\textbf{\emph{U}}_3}||_F^2 + \sigma _i^2||{\textbf{\emph{w}}_i} \circ {{\boldsymbol{\mathcal{S}}}_i}|{|_1}}$},
    \end{align}
    \end{subequations}
    where ${\boldsymbol{\mathcal{J}}}$ is the corresponding Lagrangian multiplier, and $\alpha$ is a positive scalar. The Eq. (\ref{eq:a}) performs the image deconvolution, and the remain two terms Eq. (\ref{eq:b}) and (\ref{eq:c}) denotes the image denoising process.

    According to Plancherel's theorem \cite{bracewell1966fourier}, which states that the sum of the square of a function equals the sum of the square of its Fourier transform. In view of the convolution operator in Eq. (\ref{eq:a}), we operate in the frequency domain using 3-D fast Fourier transforms (3-D FFT) to make the computation efficient\footnote{The PSF kernel and convolutional operator can be computed by the Matlab function `psf2otf' and `fftn', respectively.}. Thus, we can transform the  Eq. (\ref{eq:a}) into the following:
    \begin{equation}\label{eq:deblurring}
    \begin{aligned}
    &\resizebox{0.99\hsize}{!}{$\hat{{\mathcal{F}}({\boldsymbol{\mathcal{A}}})} = \arg \mathop {\min }\limits_{{\mathcal{F}}({\boldsymbol{\mathcal{A}}})} \frac{1}{2}{\kern 1pt} {\rm{||}}{\mathcal{F}}({\boldsymbol{\mathcal{Y}}}) - {\mathcal{F}}({\boldsymbol{\mathcal{A}}}) \circ {\mathcal{F}}({\boldsymbol{\mathcal{H}}}){\rm{||}}_F^2 + \frac{\alpha }{2}||{\mathcal{F}}({\boldsymbol{\mathcal{A}}}) - {\mathcal{F}}({\boldsymbol{\mathcal{X}}}) - {\mathcal{F}}(\frac{{\boldsymbol{\mathcal{J}}}}{\alpha })||_F^2$}.
    \end{aligned}
    \end{equation}
    The close-formed solution of Eq. (\ref{eq:deblurring}) can be expressed as:
    \begin{equation}\label{eq:deblurring-solution}
    \begin{aligned}
    &\resizebox{0.75\hsize}{!}{$\hat {\boldsymbol{\mathcal{A}}} = {{\mathcal{F}}^{{\rm{ - }}1}}\left( {\frac{{{{\mathcal{F}}^ * }({\boldsymbol{\mathcal{H}}}) \circ {\mathcal{F}}({\boldsymbol{\mathcal{Y}}}) + \alpha {\mathcal{F}}({\boldsymbol{\mathcal{X}}}) + {\mathcal{F}}({\boldsymbol{\mathcal{J}}})}}{{{{\mathcal{F}}^ * }({\boldsymbol{\mathcal{H}}}) \circ {\mathcal{F}}({\boldsymbol{\mathcal{H}}}) + \alpha {\boldsymbol{\mathcal{I}}}}}} \right)$},
    \end{aligned}
    \end{equation}
    where ${\mathcal{F}}$, ${{\mathcal{F}}^ * }$ and ${{\mathcal{F}}^{{\rm{ - }}1}}$ denotes the FFT operator, its conjugate and its inverse, respectively. The solution of  ${\boldsymbol{\mathcal{X}}}$ in Eq. (\ref{eq:b}) can be calculated similar to that of Eq. (\ref{eq:Image Resotration}), and the low-rank tensors ${\hat{\boldsymbol{\mathcal{S}}}_i}$ can be updated by Eq. (\ref{eq:Core Tensor Subproblem}). Finally, the Lagrangian multipliers and penalization parameter are updated as follows:
    \begin{equation}\label{eq:Lagrangian-update}
    \begin{aligned}
    \left\{ {\begin{array}{*{20}{l}}
    {{{\boldsymbol{\mathcal{J}}}^{k + 1}} = {{\boldsymbol{\mathcal{J}}}^k} + \alpha \left( {{\boldsymbol{\mathcal{X}}} - {\boldsymbol{\mathcal{A}}}^{k + 1}} \right)}\\
    {{\alpha ^{k + 1}} {\kern 1pt} {\kern 1pt} {\kern 1pt} {\kern 0.5pt} = \delta  \cdot {\alpha ^k}}.
    \end{array}} \right.
    \end{aligned}
    \end{equation}
    The overall procedure is summarized in \textbf{Algorithm 3}.

\begin{algorithm}[tbp]
\caption{WLRTR for HSI deblurring}
\algorithmicrequire{ Blurring image ${\boldsymbol{\mathcal{Y}}}$ and PSF ${\boldsymbol{\mathcal{H}}}$}
     \begin{algorithmic}[1]\label{alg:algo3}
      \Procedure{deblurring}{}
        \State{\textbf{Initialize:} Set parameters ${\eta, \alpha, \delta}$; ${\boldsymbol{\mathcal{X}}}^{(1)} = {\boldsymbol{\mathcal{Y}}}$;}
         \For{\emph{n}=1:\emph{N}}
           \State{Step 1: Image deconvolution ${\boldsymbol{\mathcal{A}}}$ by solving (\ref{eq:a})};
           \State{Step 2: Image reconstruction ${\boldsymbol{\mathcal{X}}}$ by solving (\ref{eq:b})};
                \For{(Low-rank tensor approximation) \emph{i}=1:\emph{I} }
                   \State{ Step 3: Core tensor estimation ${\boldsymbol{\mathcal{S}}_i}$ by solving (\ref{eq:c})};
                \EndFor
                \State{\textbf{end for}}
           \State{Step 4: Lagrangian multipliers update via (\ref{eq:Lagrangian-update})};
          \EndFor
          \State{\textbf{end for}}
          \EndProcedure
     \end{algorithmic}
\algorithmicensure{ Clean image ${\boldsymbol{\mathcal{X}}}$}.
\end{algorithm}

  %%%%%%%%%%%%%%%%%%%%%%%%%%%%%%%%%%%%%%%%%%%%%%%%%%%
  \subsection{WLRTR-RPCA for HSI super-resolution}
  %%%%%%%%%%%%%%%%%%%%%%%%%%%%%%%%%%%%%%%%%%%%%%%%%%%

    For HSI super-resolution, we consider the random noise ${\boldsymbol{\mathcal{N}}}$ with both the blurring and downsampling in spatial domain ${\boldsymbol{\mathcal{Y}}}$ and downsampling in spectral domain ${\boldsymbol{\mathcal{Z}}}$. Thus, the Eq. (\ref{eq:Regularization Framework}) can be formulated as the following minimization problem:
    \begin{equation}\label{eq:SR-WLRTR}
    \begin{aligned}
    \setlength{\abovedisplayskip}{2pt}
    \setlength{\belowdisplayskip}{2pt}
    &\resizebox{0.87\hsize}{!}{$ \left\{ {\hat{\boldsymbol{\mathcal{X}}},{{\hat{\boldsymbol{\mathcal{S}}}}_i}} \right\} = \arg \mathop {\min }\limits_{{\boldsymbol{\mathcal{X}}},{{\boldsymbol{\mathcal{S}}}_i}} \frac{1}{2}||{\boldsymbol{\mathcal{Y}}} - {\mathcal{T}_{sa}}({\boldsymbol{\mathcal{X}}})||_F^2 + \frac{1}{2}||{\boldsymbol{\mathcal{Z}}} - {\mathcal{T}_{se}}({\boldsymbol{\mathcal{X}}})||_F^2$} \\
    & \resizebox{0.87\hsize}{!}{$ +  \eta \sum\nolimits_i {\left( {||{{\mathcal{R}}_i}{\boldsymbol{\mathcal{X}}} - {{\boldsymbol{\mathcal{S}}}_i}{ \times _1}{\textbf{\emph{U}}_1}{ \times _2}{\textbf{\emph{U}}_2}{ \times _3}{\textbf{\emph{U}}_3}||_F^2 + \sigma _i^2||{\textbf{\emph{w}}_i} \circ {{\boldsymbol{\mathcal{S}}}_i}|{|_1}} \right)}$}.
    \end{aligned}
    \end{equation}

    since the variable splitting methods could seperate each term with physical meanings, for the problem HSI super-resolution with both spatial and spectral degradations, we also employ the ADMM \cite{lin2011linearized} by introducing two  auxiliary variables ${\boldsymbol{\mathcal{Q}}} = {\boldsymbol{\mathcal{X}}}$ and ${\boldsymbol{\mathcal{G}}} = {\boldsymbol{\mathcal{X}}}$ to decouple the two data fidelity term from the regularization term as follow:
    \begin{subequations}
    \label{eq:omegai}
    \begin{align}
    \label{eq:SR-a}
    &\resizebox{0.8\hsize}{!}{${\hat {\boldsymbol{\mathcal{Q}}} = \arg \mathop {\min }\limits_{\boldsymbol{\mathcal{Q}}} \frac{1}{2}{\kern 1pt} {\rm{||}}{\boldsymbol{\mathcal{Y}}} - {\mathcal{T}_{sa}}({\boldsymbol{\mathcal{Q}}}){\rm{||}}_F^2 + \frac{\beta }{2}||{\boldsymbol{\mathcal{Q}}} - {\boldsymbol{\mathcal{X}}} - \frac{\boldsymbol{\mathcal{J}}_1}{\beta }||_F^2}$}\\
    \label{eq:SR-b}
    &\resizebox{0.8\hsize}{!}{${\hat {\boldsymbol{\mathcal{G}}} = \arg \mathop {\min }\limits_{\boldsymbol{\mathcal{G}}} \frac{1}{2}{\kern 1pt} {\rm{||}}{\boldsymbol{\mathcal{Z}}} - {\mathcal{T}_{se}}({\boldsymbol{\mathcal{G}}}){\rm{||}}_F^2 + \frac{\gamma }{2}||{\boldsymbol{\mathcal{G}}} - {\boldsymbol{\mathcal{X}}} - \frac{\boldsymbol{\mathcal{J}}_2}{\gamma }||_F^2}$}\\
    \begin{split}\label{eq:SR-c}
    &\resizebox{0.75\hsize}{!}{${\hat {\boldsymbol{\mathcal{X}}} =  \arg \mathop {\min }\limits_{\boldsymbol{\mathcal{X}}} \eta \sum\nolimits_i {||{{\mathcal{R}}_i}{\boldsymbol{\mathcal{X}}} - {{\boldsymbol{\mathcal{S}}}_i}{ \times _1}{\textbf{\emph{U}}_1}{ \times _2}{\textbf{\emph{U}}_2}{ \times _3}{\textbf{\emph{U}}_3}||_F^2} }$} \\
    &{\kern 1pt}{\kern 1pt}{\kern 1pt}{\kern 1pt}{\kern 1pt}{\kern 1pt}{\kern 1pt}{\kern 1pt}{\kern 1pt}{\kern 1pt}{\kern 1pt}{\kern 1pt}{\kern 1pt}\resizebox{0.7\hsize}{!}{$+ \frac{\beta }{2}||{\boldsymbol{\mathcal{Q}}} - {\boldsymbol{\mathcal{X}}} - \frac{\boldsymbol{\mathcal{J}}_1}{\beta }||_F^2 + \frac{\gamma }{2}||{\boldsymbol{\mathcal{G}}} - {\boldsymbol{\mathcal{X}}} - \frac{\boldsymbol{\mathcal{J}}_2}{\gamma }||_F^2$}
    \end{split}
    \\
    \label{eq:SR-d}
    &\resizebox{0.85\hsize}{!}{${{{\hat {\boldsymbol{\mathcal{S}}}}_i} = \arg \mathop {\min }\limits_{{{\boldsymbol{\mathcal{S}}}_i}} ||{{\mathcal{R}}_i}{\boldsymbol{\mathcal{X}}} - {{\boldsymbol{\mathcal{S}}}_i}{ \times _1}{\textbf{\emph{U}}_1}{ \times _2}{\textbf{\emph{U}}_2}{ \times _3}{\textbf{\emph{U}}_3}||_F^2 + \sigma _i^2||{\textbf{\emph{w}}_i} \circ {{\boldsymbol{\mathcal{S}}}_i}|{|_1}}$}.
    \end{align}
    \end{subequations}
    where ${\boldsymbol{\mathcal{J}}_1}$ and ${\boldsymbol{\mathcal{J}}_2}$ are the corresponding Lagrangian multipliers, and $\alpha$ and $\beta$ are positive scalars. The Eq. (\ref{eq:SR-a}) and  Eq. (\ref{eq:SR-b}) performs the HSI super-resolution, respectively, and the remain two terms Eq. (\ref{eq:SR-c}) and (\ref{eq:SR-d}) denotes the image denoising process. Each subproblem has the close-formed solution. For Eq. (\ref{eq:SR-a}) and Eq. (\ref{eq:SR-b}), the subproblems can be solved by computing:
    \begin{equation}\label{eq:Spatial-SR}
    \begin{aligned}
    {\mathcal{T}}_{comp}({\boldsymbol{\mathcal{Q}}}) = {\mathcal{T}}_{sa}^T({\boldsymbol{\mathcal{Y}}}) + \beta {\boldsymbol{\mathcal{X}}} + {{\boldsymbol{\mathcal{J}}}_1},
    \end{aligned}
    \end{equation}
    \begin{equation}\label{eq:Spectral-SR}
    \begin{aligned}
    {\boldsymbol{\mathcal{G}}}{ \times _3}({\textbf{\emph{P}}^T}\textbf{\emph{P}} + \gamma \textbf{\emph{I}}) = {\boldsymbol{\mathcal{Z}}}{ \times _3}{\textbf{\emph{P}}^T} + \gamma {\boldsymbol{\mathcal{X}}} + {\boldsymbol{\mathcal{J}}_2},
    \end{aligned}
    \end{equation}
    where ${\mathcal{T}}_{comp} = {\mathcal{T}}_{sa}^T{{\mathcal{T}}_{sa}} + \beta {\boldsymbol{\mathcal{I}}}$ is the composite operator on ${\boldsymbol{\mathcal{Q}}}$, and ${\mathcal{T}}_{sa}^T({\boldsymbol{\mathcal{Y}}})$ means the transposed blurring and downsampling on ${\boldsymbol{\mathcal{Y}}}$. Since it is hard for us to calculate the Eq. (\ref{eq:Spatial-SR}) and Eq. (\ref{eq:Spectral-SR}) directly, in our implementation, we unfold the 3-D tensors along the mode-3 to the 2-D matrixes as \cite{dong2016hyperspectral}.  The solution of  ${\boldsymbol{\mathcal{X}}}$ in Eq. (\ref{eq:SR-c}) can be calculated similar to that of Eq. (\ref{eq:Image Resotration}), and the low-rank tensors ${\hat{\boldsymbol{\mathcal{S}}}_i}$ can be updated by Eq. (\ref{eq:Core Tensor Subproblem}). Finally, the Lagrangian multipliers and penalization parameter are updated as follows:
    \begin{equation}\label{eq:Lagrangian-update-SR}
    \begin{aligned}
    \left\{ {\begin{array}{*{20}{l}}
    {{{\boldsymbol{\mathcal{J}}_1}^{k + 1}} = {{\boldsymbol{\mathcal{J}}_1}^k} + \beta \left( {{\boldsymbol{\mathcal{X}}} - {\boldsymbol{\mathcal{Q}}}^{k + 1}} \right)}\\
    {{{\boldsymbol{\mathcal{J}}_2}^{k + 1}} = {{\boldsymbol{\mathcal{J}}_2}^k} + \gamma \left( {{\boldsymbol{\mathcal{X}}} - {\boldsymbol{\mathcal{G}}}^{k + 1}} \right)}\\
    {{\beta ^{k + 1}} {\kern 1pt} {\kern 1pt} {\kern 1pt} {\kern 0.5pt} = \delta  \cdot {\beta ^k}}, {{\gamma ^{k + 1}} {\kern 1pt} {\kern 1pt} {\kern 1pt} {\kern 0.5pt} = \delta  \cdot {\gamma ^k}}.
    \end{array}} \right.
    \end{aligned}
    \end{equation}
    The overall procedure is summarized in \textbf{Algorithm 4}.

\begin{algorithm}[tbp]
\caption{WLRTR for HSI super-resolution}
\algorithmicrequire {Spatial and spectral LR image $\{{\boldsymbol{\mathcal{Y}}}, {\boldsymbol{\mathcal{Z}}}\}$, PSF ${\boldsymbol{\mathcal{H}}}$}
     \begin{algorithmic}[1]\label{alg:algo4}
      \Procedure{Super-resolution}{}
\State{\textbf{Initialize:} Set parameters ${\eta, \beta, \gamma, \delta}$; ${\boldsymbol{\mathcal{X}}}^{(1)} = \uparrow_{upsampling}({\boldsymbol{\mathcal{Y}}}$});
         \For{\emph{n}=1:\emph{N}}
           \State{Step 1: Spatial super-resolution ${\boldsymbol{\mathcal{Q}}}$ by solving (\ref{eq:SR-a})};
           \State{Step 2: Spectral super-resolution ${\boldsymbol{\mathcal{G}}}$ by solving (\ref{eq:SR-b})};
           \State{Step 3: Image reconstruction ${\boldsymbol{\mathcal{X}}}$ by solving (\ref{eq:SR-c})};
                 \For{(Low-rank tensor approximation) \emph{i}=1:\emph{I} }
                   \State{Step 4: Core tensor estimation ${\boldsymbol{\mathcal{S}}_i}$ by solving (\ref{eq:SR-d})};
                \EndFor
                \State{\textbf{end for}}
           \State{Step 5: Lagrangian multipliers update via (\ref{eq:Lagrangian-update-SR})};
          \EndFor
           \State{\textbf{end for}}
          \EndProcedure
     \end{algorithmic}
\algorithmicensure{ Clean image ${\boldsymbol{\mathcal{X}}}$}.
\end{algorithm}

%%%%%%%%%%%%%%%%%%%%%
\section{Experimental results}
%%%%%%%%%%%%%%%%%%%%%
    In this section, extensive experiments are presented to evaluate the performance of the proposed methods. We will first introduce the experimental setting about the competing state-of-the-art HSIs restoration methods and also the evaluation indexes. Then, the results of both simulated and real benchmark datasets are presented. At last, we give a discussion about the details of our methods.

\begin{table*}[]
\scriptsize
\centering
\renewcommand{\arraystretch}{1.3}
\caption{Quantitative results of differnent methods under several noise levels on CAVE dataset.}
\label{CAVE_Quantitative}
\begin{tabular}{|C{0.65cm}|C{0.95cm}|C{0.95cm}|C{0.95cm}|C{0.95cm}|C{1.10cm}|C{0.95cm}|C{0.95cm}|C{0.95cm}|C{0.95cm}|C{0.95cm}|C{0.95cm}|C{0.95cm}|}
%{|c|c|c|c|c|c|c|c|c|c|c|c|c|}
\hline
\multirow{2}{*}{Sigma} & \multirow{2}{*}{Index} & \multicolumn{11}{c|}{Methods}                                                                               \\ \cline{3-13}
                       &                        & Noisy   & BM3D   & PARAFAC & LRTA   & LRMR   & ANLM   & NMF    & BM4D   & TDL    & ITSReg   & WLRTR           \\ \hline
\multirow{4}{*}{10}    & PSNR                   & 28.13   & 42.09  & 35.43   & 41.36  & 39.27  & 41.52  & 43.15  & 44.59  & 44.30  & 45.77  & \textbf{46.85}  \\ \cline{2-13}
                       & SSIM                   & 0.4371  & 0.9665 & 0.8767  & 0.9499 & 0.9094 & 0.9576 & 0.9702 & 0.9784 & 0.9797 & 0.9802 & \textbf{0.9873} \\ \cline{2-13}
                       & ERGAS                  & 236.40  & 45.06  & 108.37  & 49.53  & 64.81  & 47.78  & 39.65  & 33.33  & 34.86  & 30.53  & \textbf{25.91}  \\ \cline{2-13}
                       & SAM                    & 0.7199  & 0.1395 & 0.2360  & 0.1719 & 0.3343 & 0.2184 & 0.1358 & 0.1295 & 0.1025 & 0.1086 & \textbf{0.0863} \\ \hline
\multirow{4}{*}{20}    & PSNR                   & 22.11   & 38.46  & 34.53   & 38.04  & 34.38  & 37.42  & 39.02  & 41.02  & 41.06  & 42.54  & \textbf{43.67}  \\ \cline{2-13}
                       & SSIM                   & 0.1816  & 0.9339 & 0.8574  & 0.9119 & 0.7807 & 0.8936 & 0.9169 & 0.9550 & 0.9638 & 0.9650 & \textbf{0.9769} \\ \cline{2-13}
                       & ERGAS                  & 472.88  & 68.38  & 115.81  & 72.16  & 113.47 & 76.15  & 63.61  & 50.38  & 50.47  & 44.12  & \textbf{37.64}  \\ \cline{2-13}
                       & SAM                    & 0.9278  & 0.1984 & 0.2838  & 0.2139 & 0.5009 & 0.3358 & 0.1946 & 0.1981 & 0.1284 & 0.1171 & \textbf{0.1067} \\ \hline
\multirow{4}{*}{30}    & PSNR                   & 18.59   & 36.40  & 33.59   & 36.15  & 31.36  & 34.77  & 36.53  & 38.90  & 39.03  & 40.51  & \textbf{41.68}  \\ \cline{2-13}
                       & SSIM                   & 0.0988  & 0.9034 & 0.8261  & 0.8787 & 0.6451 & 0.8060 & 0.8565 & 0.9277 & 0.9486 & 0.9488 & \textbf{0.9666} \\ \cline{2-13}
                       & ERGAS                  & 709.29  & 88.29  & 128.07  & 91.40  & 157.65 & 104.95 & 86.25  & 65.38  & 63.54  & 53.05  & \textbf{47.36}  \\ \cline{2-13}
                       & SAM                    & 1.0414  & 0.2489 & 0.3455  & 0.2479 & 0.6021 & 0.4376 & 0.2465 & 0.2598 & 0.1520 & 0.1374 & \textbf{0.1248} \\ \hline
\multirow{4}{*}{50}    & PSNR                   & 14.15   & 32.66  & 30.22   & 32.44  & 26.67  & 30.74  & 31.98  & 35.96  & 36.42  & 37.75  & \textbf{39.06}  \\ \cline{2-13}
                       & SSIM                   & 0.0432  & 0.8320 & 0.7051  & 0.7932 & 0.4000 & 0.6057 & 0.7113 & 0.8666 & 0.9175 & 0.9271 & \textbf{0.9457} \\ \cline{2-13}
                       & ERGAS                  & 1181.95 & 115.06 & 155.84  & 118.64 & 264.28 & 164.55 & 123.23 & 91.51  & 85.58  & 70.16  & \textbf{63.83}  \\ \cline{2-13}
                       & SAM                    & 1.1741  & 0.2877 & 0.4460  & 0.2843 & 0.7534 & 0.5806 & 0.3148 & 0.3575 & 0.2000 & 0.1619 & \textbf{0.1580} \\ \hline
\multirow{4}{*}{100}   & PSNR                   & 8.13    & 29.27  & 26.01   & 29.20  & 20.84  & 24.90  & 26.95  & 30.82  & 32.91  & 33.01  & \textbf{35.15}  \\ \cline{2-13}
                       & SSIM                   & 0.0122  & 0.7460 & 0.4346  & 0.6945 & 0.1850 & 0.2826 & 0.4643 & 0.6956 & 0.8344 & 0.8648 & \textbf{0.8876} \\ \cline{2-13}
                       & ERGAS                  & 2364.05 & 171.94 & 253.70  & 175.91 & 469.26 & 324.48 & 225.55 & 141.18 & 128.22 & 120.77 & \textbf{100.44} \\ \cline{2-13}
                       & SAM                    & 1.3271  & 0.3938 & 0.6843  & 0.3381 & 0.9306 & 0.7972 & 0.4321 & 0.5014 & 0.3079 & 0.2376 & \textbf{0.2300} \\ \hline
\end{tabular}
\end{table*}

  %%%%%%%%%%%%%%%%%%%%%%%%%%%%%%%%%%%%%%%%%%%%%%%%%%%
  \subsection{Experimental Setting}
  %%%%%%%%%%%%%%%%%%%%%%%%%%%%%%%%%%%%%%%%%%%%%%%%%%%

    \noindent
    \textbf{Benchmark Datasets}. In our work, we test various datasets, including HSIs, color images (3 channel), and MRIs:

    \begin{itemize}
    \item Columbia Multispectral database (CAVE) \cite{yasuma2010generalized}. The whole dataset consisting of 32 noiseless hyperspectral images of size 512*512*31 are captured with the wavelengths in the range of 400-700 nm at an interval of 10 nm.
    \item Berkeley Segmentation Dataset (BSD) \cite{martin2001a}. The whole clean color image dataset consists of 200 training image and 100 test images of the size 481*321*3. A subset (68 images) of the test section of the BSD is used to evaluate the denoising performance \cite{roth2005fields}.
    \item Harvard real-world hyperspectral datasets (HHD) \cite{chakrabarti2011statistics}. The whole dataset consisting of 50 noisy hyperspectral images of size 1040*1392*31 are captured with the wavelengths in the range of 420-720 nm at an interval of 10.
    \item Prostate MRI Dataset (PMRI)\footnote{\url{http://prostatemrimagedatabase.com}}. The whole dataset consists of prostate samples of 230 patients with varying image bands, due to the different observation days on the patients. One thing is worth to note that the content in this dataset gradually varies from each frame to the other.
    \item Remotely Sensed HSIs\footnote{\url{http://www.ehu.eus/ccwintco/index.php?title=Hyperspectral_Remote_Sensing_Scenes}}. Two remotely sensed hyperspectral datasets are used in this paper, i.e. Salinas, and Gulf Wetlands.
    \end{itemize}

    \noindent
    \textbf{Pre-processing}. First, before the restoration process, all the original images were coded to an 8-bit scale for display convenience and uniform parameter setting. Second, for the non-local similarity cubic matching, we do not directly searching from the 3-D cubics in the noisy data. Instead, for reducing computational load and matching accuracy, we proposed to average each bands of the cubic, which can be regarded as an uniform filtering procedure, so that we can obtain a quite clean 2-D matrix. Note that, the non-local similarity matching processing is on this 2-D matrix, while our restoration process is still on the whole 3-D cubics.

    \noindent
    \textbf{Baselines}. For the HSI denoising methods, we compare with block-matching and 3D filtering (BM3D) \cite{dabov2007image}, parallel factor analysis (PARAFAC) \cite{liu2012denoising}, low-rank tensor approximation (LRTA) \cite{renard2008denoising}, low-rank matrix recovery (LRMR) \cite{zhang2014hyperspectral}, adaptive non-local means denoising (ANLM) \cite{manjon2010adaptive}, nonnegative matrix factorization (NMF) \cite{ye2015multitask}, block-matching and 4D filtering (BM4D) \cite{Maggioni2012Nonlocal}, tensor dictionary learning (TDL) \cite{peng2014decomposable}, intrinsic tensor sparsity regularization  (ITSReg) \cite{xie2016multispectral}; for HSI deblurring, the competing methods include single image based deblurring method hyper-Laplacian (HL) \cite{krishnan2009fast}, and two HSI deblurring methods fast positive deconvolution (FPD) \cite{henrot2013fast} and spectral-spatial total variation (SSTV) \cite{fang2017hyperspectral}; for HSI super-resolution, we compare with coupled nonnegative matrix factorization (CNMF) \cite{yokoya2012coupled}, non-negative structured sparse representation (NSSR) \cite{dong2016hyperspectral} and non-local sparse tensor factorization (NLSTF) \cite{dian2017hyperspectral}.

    We use the codes provided by the authors downloaded from their homepages, and fine tune the parameters by default or following the rules in their papers to achieve the best performance. And the Matlab code of our methods can be downloaded from the homepage of the author\footnote{\url{http://www.escience.cn/people/changyi/index.html}}. For parameter setting of our method, the most important parameter is the number of non-local cubic, which is set between [100, 200] in correspondence with the noise level, respectively. The patch size is between [6, 8]. And another important factor is the regularization parameter $\eta$ for the HSI deblurring and super-resolution, which is set as ${10^{ - 8}}$ and ${10^{ - 5}}$, respectively.

    \noindent
    \textbf{Evaluation Indexes}. In order to give an overall evaluation of the denoising performance, four quantitative quality indices are employed, including peak signal-to-noise ratio (PSNR), structure similarity (SSIM \cite{wang2004image}), erreur relative globale adimensionnelle de synthese (ERGAS \cite{wald2002data}) and spectral angle map (SAM \cite{yuhas1993determination}). PSNR and SSIM are two conventional indexes, which is used to evaluate the similarity between the restored image and the reference image based on MSE and structural consistency, respectively. ERGAS measures fidelity of the restored image based on the weighted sum of MSE in each band. SAM is introduced to measure the spectral fidelity between the restored image and the reference image across all spatial positions. The PSNR and SSIM evaluate the spatial quality, and the ERGAS and SAM assess the spectral qualtiy.  The larger PSNR and SSIM values are, the smaller ERGAS and SAM values are, the better the restored images are.

\begin{figure*}
\begin{center}
    \includegraphics[width=0.905\textwidth]{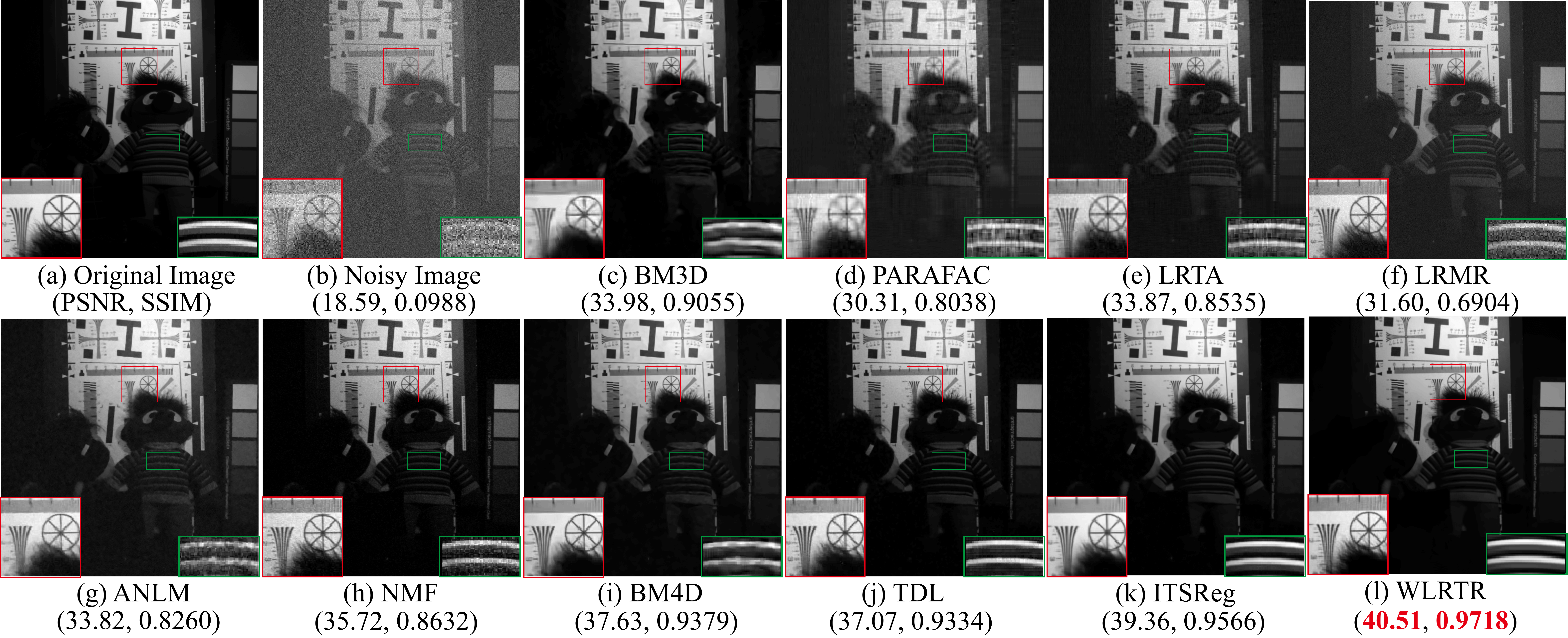}
\end{center}
   \caption{Simulated random noise removal results under noise level $\sigma$ = 30 on CAVE dataset. (a) Original image \emph{toy} at band 510nm. (b) Noisy image. Denoising results by (c) BM3D, (d) PARAFAC, (e) LRTA, (f) LRMR, (g) ANLM, (h) NMF, (i) BM4D, (j) TDL, (k) ITSReg, and (l) WLRTR.}
\label{Simulated_Toy}
\end{figure*}

\begin{figure*}
\begin{center}
    \includegraphics[width=0.905\textwidth]{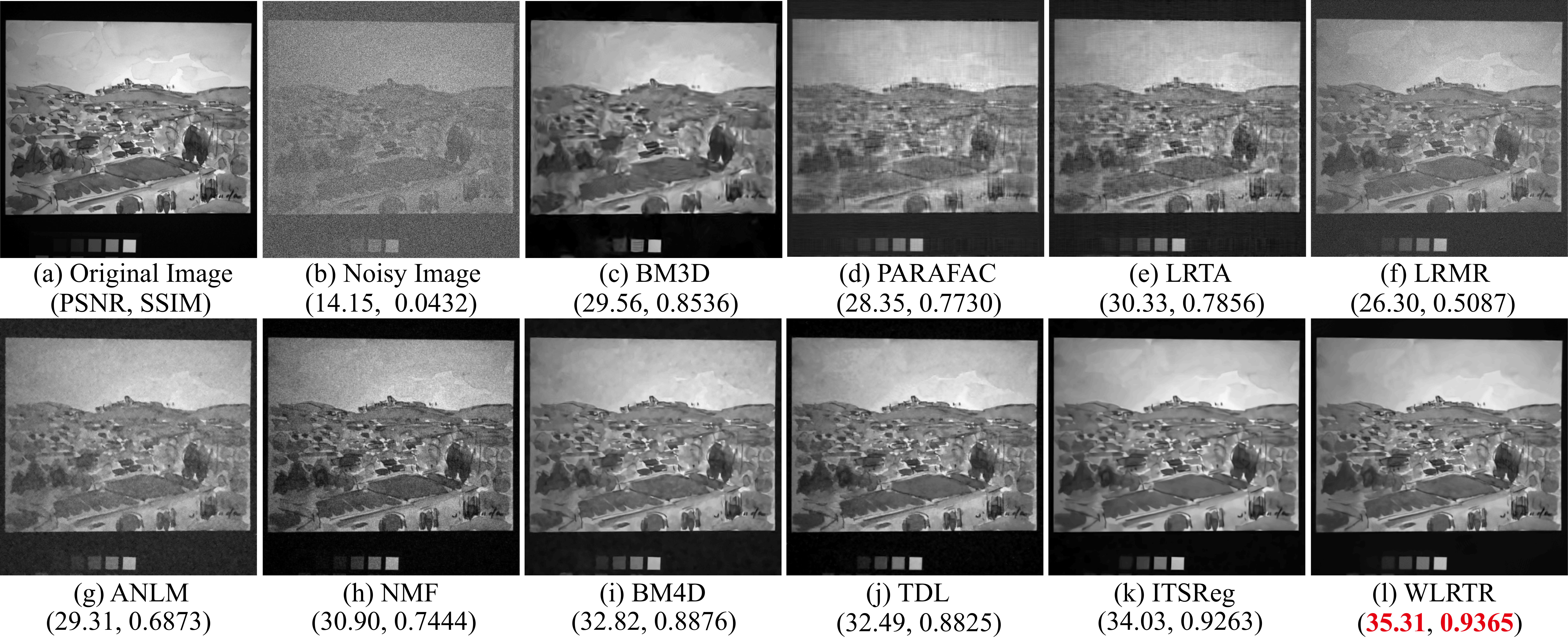}
\end{center}
   \caption{Simulated random noise removal results under noise level $\sigma$ = 50 on CAVE dataset. (a) Original image \emph{watercolors} at band 510nm. (b) Noisy image. Denoising results by (c) BM3D, (d) PARAFAC, (e) LRTA, (f) LRMR, (g) ANLM, (h) NMF, (i) BM4D, (j) TDL, (k) ITSReg, and (l) WLRTR.}
\label{Simulated_Watercolors}
\end{figure*}

\begin{figure*}
\begin{center}
    \includegraphics[width=0.9\textwidth]{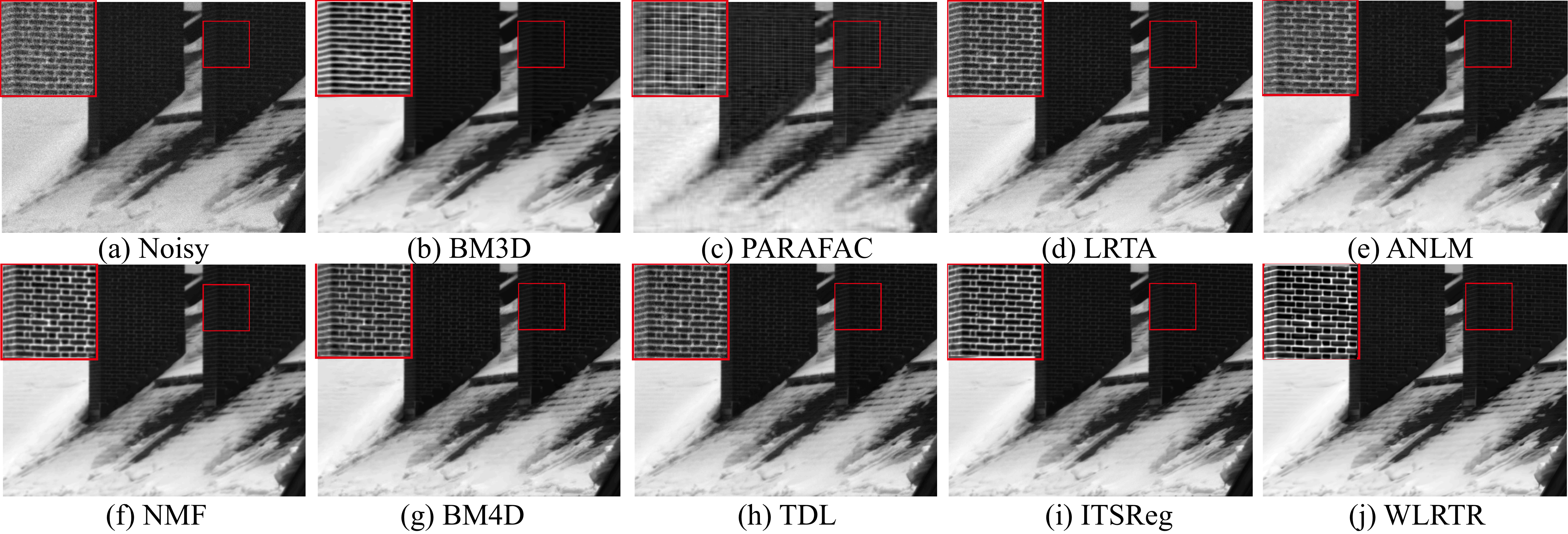}
\end{center}
   \caption{Real random noise removal results on HHD dataset. (a) Noisy image. Denoising results by (b) BM3D, (c) PARAFAC, (d) LRTA, (e) ANLM, (f) NMF, (g) BM4D, (h) TDL, (i) ITSReg, (j) WLRTR.}
\label{Real HHD}
\end{figure*}

\begin{figure*}
\begin{center}
    \includegraphics[width=0.99\textwidth]{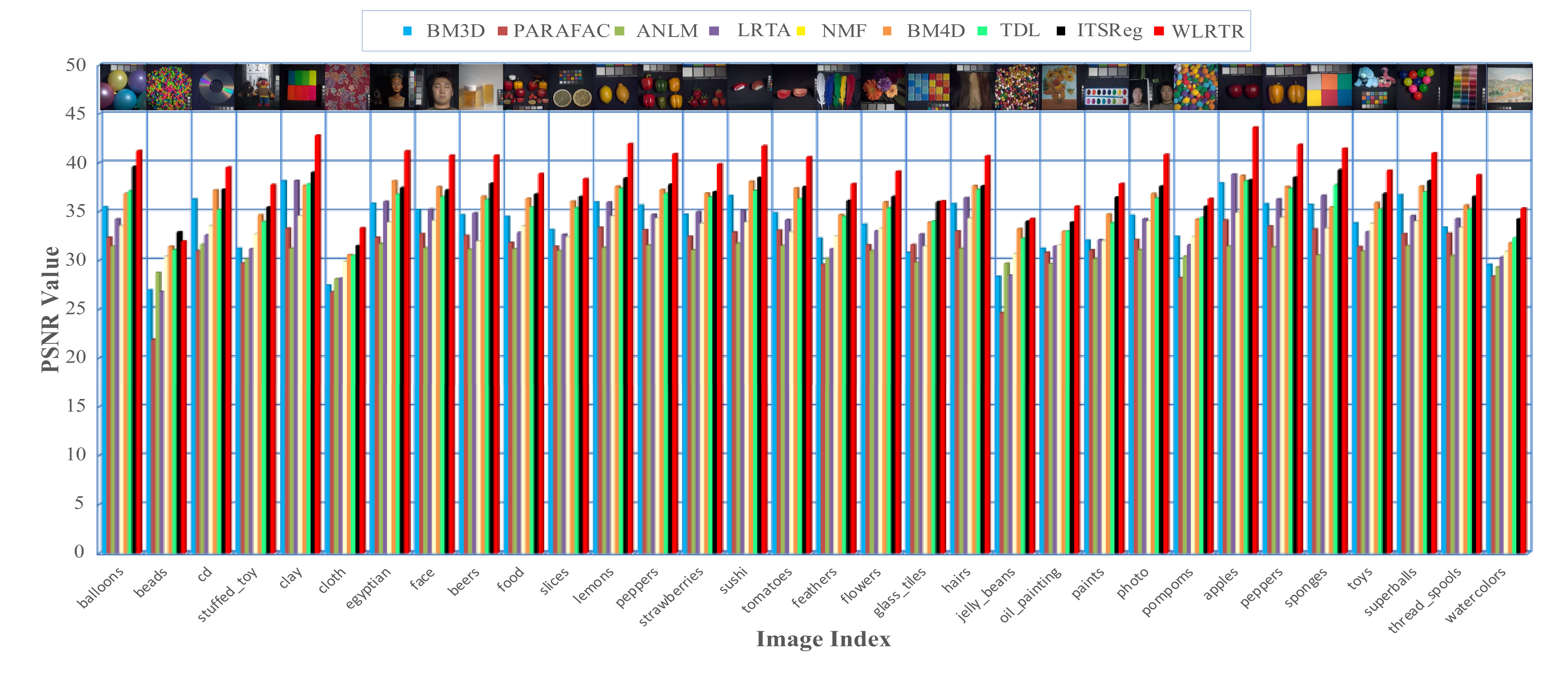}
\end{center}
   \caption{Quantitative index PSNR value comparison under noise level $\sigma$ = 50 on the dataset CAVE of all scenes.}
\label{Quantitative on CAVE}
\end{figure*}

  %%%%%%%%%%%%%%%%%%%%%%%%%%%%%%%%%%%%%%%%%%%%%%%%%%%
  \subsection{HSI Denoising}
  %%%%%%%%%%%%%%%%%%%%%%%%%%%%%%%%%%%%%%%%%%%%%%%%%%%
    To visually illustrate the denoising performance of WLRTR, we choose two images \emph{toy} and \emph{watercolor} of band 510nm under different noise level, as shown in Figs. \ref{Simulated_Toy} and \ref{Simulated_Watercolors}. In Fig. \ref{Simulated_Toy} (the green demarcated window), we can clearly see from the enlarged region that the proposed method has obtained more clear result, compared with other competing methods. Moreover, looking at the red demarcated window in Fig. \ref{Simulated_Toy}, the proposed WLRTR is capable of well reconstructing the tiny hair texture. When the noise level is high, other methods generate much more artifacts, as shown in Fig. \ref{Simulated_Watercolors}. And the result of WLRTR shows more art taste for this painting in Fig. \ref{Simulated_Watercolors}. We also test the proposed WLRTR method on real noisy HSI. Since the noise level is unknown for real noisy images, we adopted an estimation method from \cite{pyatykh2013image} to estimate the noise level beforehand. In Fig. \ref{Real HHD}, from the demarcated window, we can observe that WLRTR method obtains smoother image with clearer texture and line pattern. In summary, WLRTR has obtained better performance in terms of noise suppression, detail preserving, artifacts-free, visual pleasure and PSNR value under differnent noise level.

    We also present the overall quantitative assessments of all competing methods on CAVE in Table \ref{CAVE_Quantitative}. The highest PSNR and SSIM values and lowest ERGAS and SAM values are highlighted in bold. We have the following observations. First, WLRTR consistently achieves the best performance in four assessments, which highly demonstrate the effectiveness of WLRTR for HSIs. Second, for random noise in CAVE, with the increasing of noise level, the advantage of our method over other methods becomes bigger, almost exceed 4.3dB than BM4D at $\sigma$ = 100.

\begin{figure}
\begin{center}
    \includegraphics[width=0.4\textwidth]{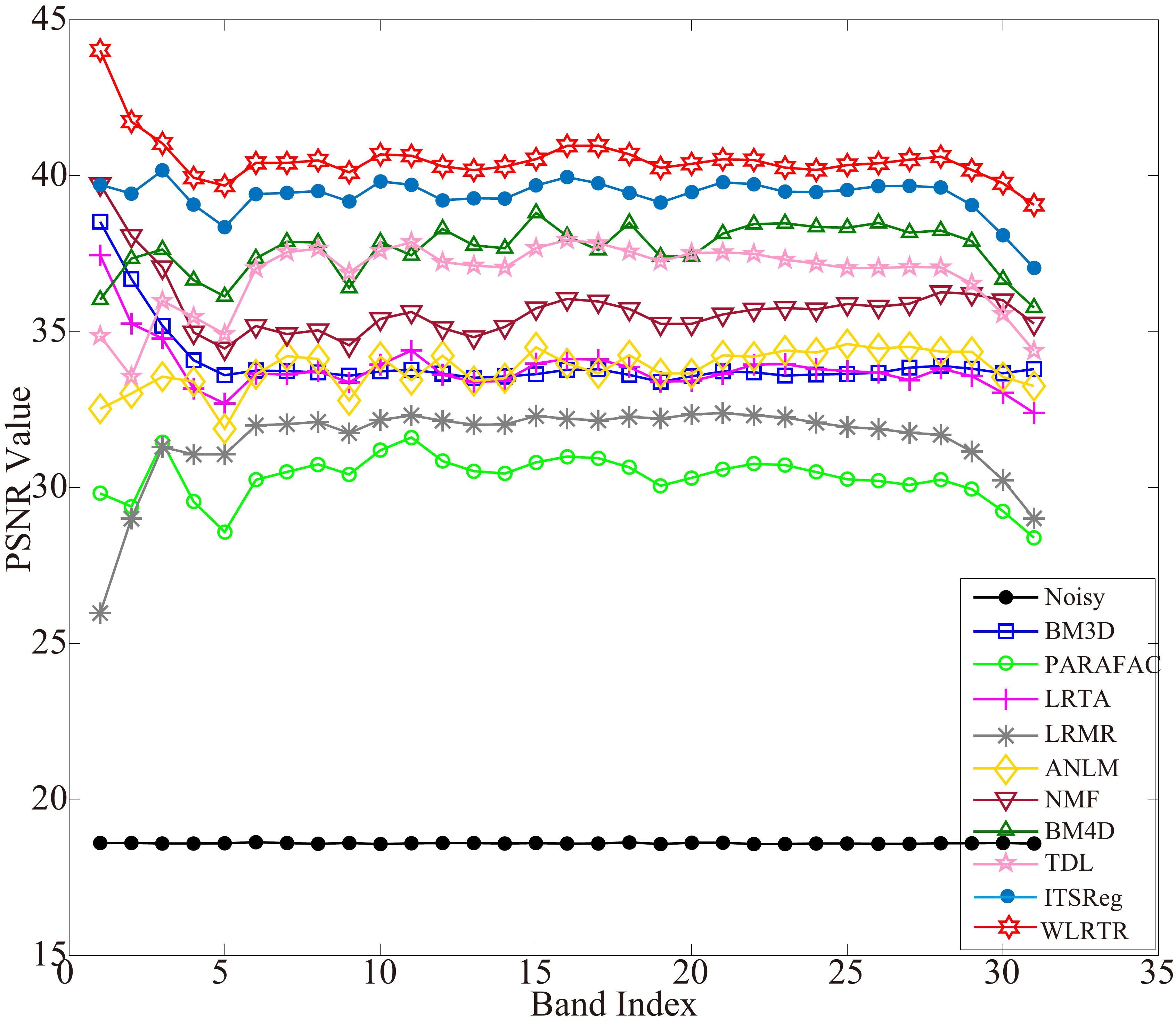}
\end{center}
   \caption{PSNR values of each band of image \emph{toy} under noise level $\sigma$ = 30 on CAVE dataset.}
\label{PSNR of each band}
\end{figure}

    For each scene in CAVE, we compute the average PSNR value of all the competing methods, as shown in Fig. \ref{Quantitative on CAVE}. The WLRTR method obtains the highest average PSNR values among each scene in term of different image contents. Furthermore, we plot the PSNR values of each band of one single image toy in CAVE as an example, as shown in Fig. \ref{PSNR of each band}. It can be observed that the PSNR values of all the bands obtained by WLRTR are significantly higher than those of the other methods.

\begin{table}[tbp]
\scriptsize
\renewcommand{\arraystretch}{1.2}
\centering
\caption{Quantitative results of differnent methods under several noise levels on BSD.}
\label{BSD_Quantitative}
\begin{tabular}{|c|c|c|c|c|c|}
\hline
\multirow{2}{*}{Sigma} & \multirow{2}{*}{Index} & \multicolumn{4}{c|}{Methods}      \\ \cline{3-6}
                       &                        & Noisy  & LSCD   & CBM3D  & WLRTR   \\ \hline
\multirow{2}{*}{10}    & PSNR                   & 28.13  & 33.85  & 35.90  & \textbf{35.91}  \\ \cline{2-6}
                       & SSIM                   & 0.7020 & 0.9188 & 0.9501 & \textbf{0.9511} \\ \hline
\multirow{2}{*}{20}    & PSNR                   & 22.17  & 30.26  & 31.85  & \textbf{31.94}  \\ \cline{2-6}
                       & SSIM                   & 0.4580 & 0.8469 & 0.8923 & \textbf{0.8953} \\ \hline
\multirow{2}{*}{30}    & PSNR                   & 18.58  & 28.22  & 29.69  & \textbf{29.87}  \\ \cline{2-6}
                       & SSIM                   & 0.3223 & 0.7854 & 0.8402 & \textbf{0.8444} \\ \hline
\multirow{2}{*}{40}    & PSNR                   & 16.08  & 27.00  & 28.10  & \textbf{28.47}  \\ \cline{2-6}
                       & SSIM                   & 0.2388 & 0.7417 & 0.7872 & \textbf{0.7973} \\ \hline
\end{tabular}
\end{table}

\begin{figure*}
\begin{center}
    \includegraphics[width=0.9\textwidth]{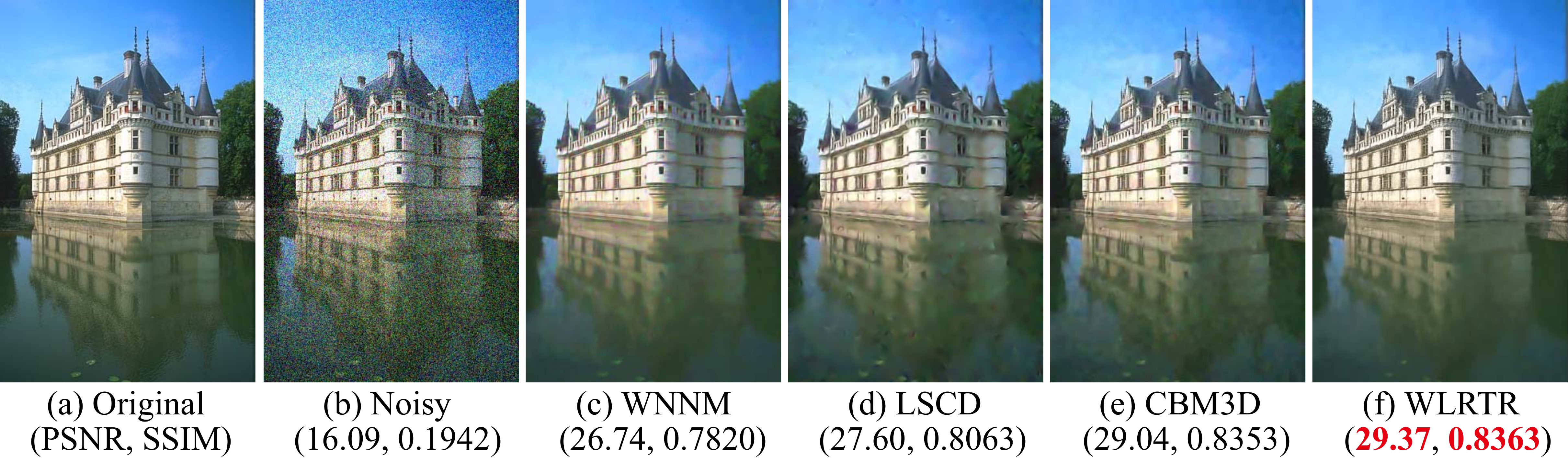}
\end{center}
   \caption{Simulated color image results under noise level $\sigma$ = 40 on BSD dataset. (a) Original image \emph{castle}. (b) Noisy image. Denoising results by (c) WNNM, (d) LSCD, (e) CBM3D, and (f) WLRTR.}
\label{Simulated castle}
\end{figure*}

  %%%%%%%%%%%%%%%%%%%%%%%%%%%%%%%%%%%%%%%%%%%%%%%%%%%
  \subsection{Color Image Denoising}
  %%%%%%%%%%%%%%%%%%%%%%%%%%%%%%%%%%%%%%%%%%%%%%%%%%%
    Although WLRTR is proposed for HSIs which possess dozens or hundreds of continuous bands, it can be also well extended to various multiple images with less bands, such as RGB color image. Most of the previous color image processing methods usually handle the RGB images in luminance space or restore each channel separately, while ingoring the spectral correlation between channel in RGB images. On the contrary, the WLRTR jointly processes the R, G and B channel. In this section, we compared WLRTR method with WNNM \cite{Gu2014Weighted}, which handles the color image in each channel, and state-of-the-arts color image denoising methods, such as LSCD \cite{rizkinia2016local}, color BM3D \cite{dabov2007color} (CBM3D). Figure \ref{Simulated castle} shows representative color images denoising result on BSD under noise level $\sigma$ = 40. Compared with WNNM, WLRTR exhibits much more details in texture regions or edges. Also, when compared with other competing color image denoising methods, WLRTR could better preserve the image details while with less chrominance color artifacts, with better human perception and higher PSNR values. The PSNR and SSIM values on BSD are reported in Table \ref{BSD_Quantitative}. From Table \ref{BSD_Quantitative}, we can conclude that the joint utilization of RGB multichannel manner in color image really improves the denoising performance.

\begin{figure}
\begin{center}
    \includegraphics[width=0.5\textwidth]{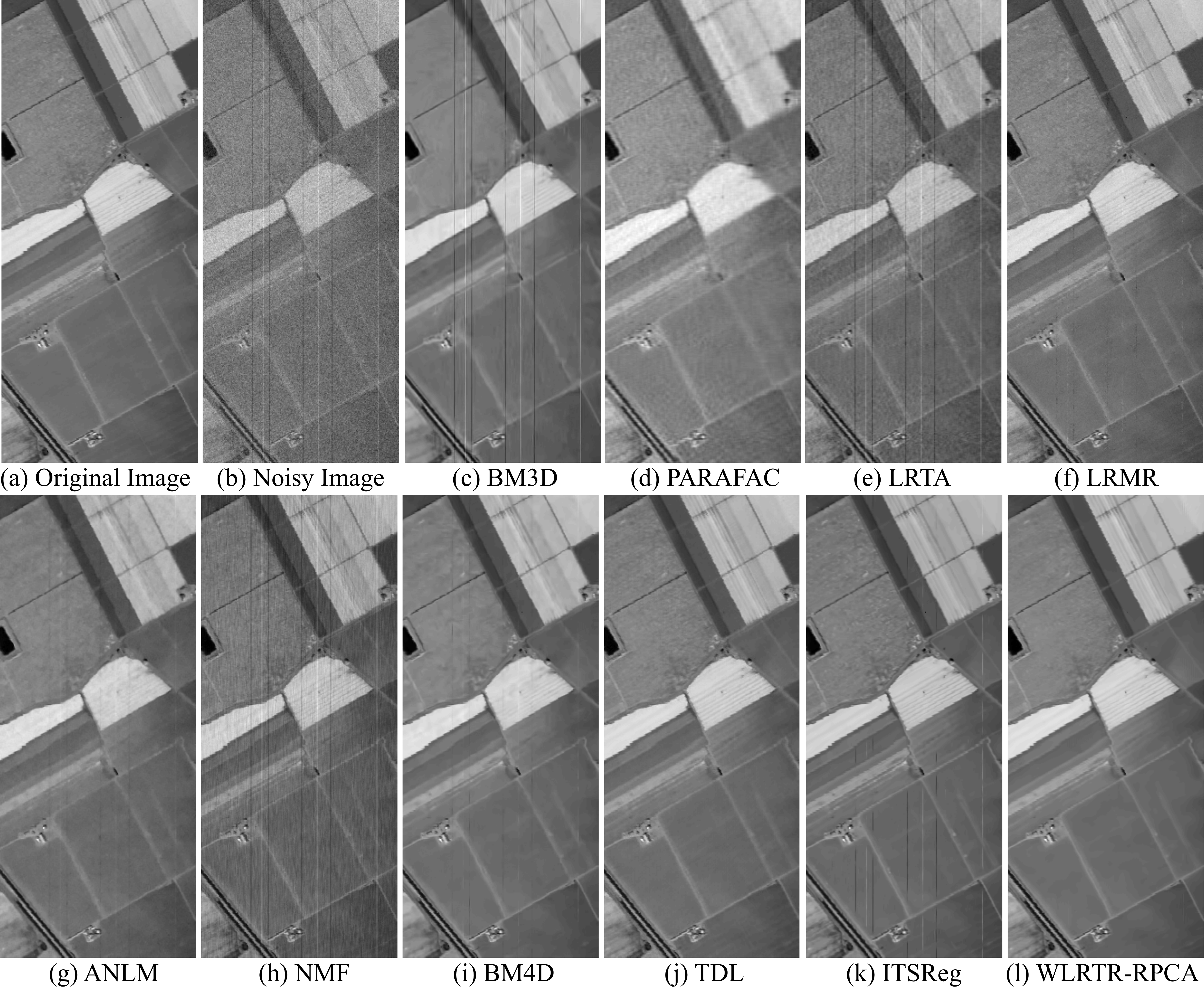}
\end{center}
   \caption{Simulated mixed noise removal results under heavy noise level on Salinas data. (a) Original image at band 80. (b) Noisy image. Denoising results by (c) BM3D, (d) PARAFAC, (e) LRTA, (f) LRMR, (g) ANLM, (h) NMF, (i) BM4D, (j) TDL, (k) ITSReg, and (l) WLRTR-RPCA .}
\label{Simulated Salinas}
\end{figure}

\begin{figure}
\begin{center}
    \includegraphics[width=0.48\textwidth]{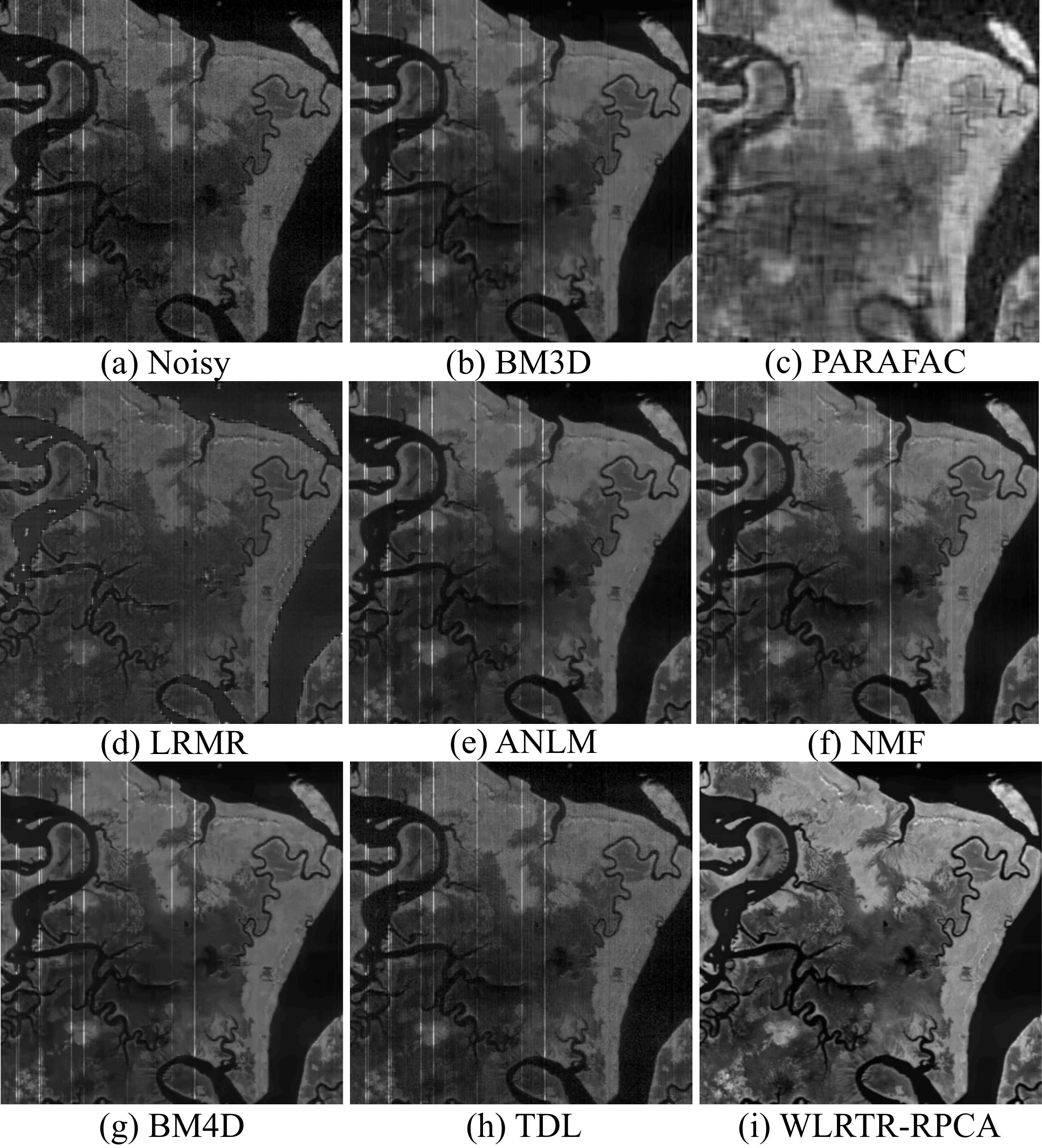}
\end{center}
   \caption{Real stripe noise removal results on Gulf Wetlands dataset. (a) Noisy image. Denoising results by (b) BM3D, (c) PARAFAC, (d) LRMR, (e) ANLM, (f) NMF, (g) BM4D, (h) TDL, (i) WLRTR-RPCA.}
\label{Real Gulf Wetlands}
\end{figure}

  %%%%%%%%%%%%%%%%%%%%%%%%%%%%%%%%%%%%%%%%%%%%%%%%%%%
  \subsection{HSI Destriping}
  %%%%%%%%%%%%%%%%%%%%%%%%%%%%%%%%%%%%%%%%%%%%%%%%%%%
    In this section, we evaluate the WLRTR-RPCA model on the very common mixed noise in HSIs: random noise and stripe noise. We randomly added the stripe on HSI Salinas, and the locations of the stripes between the neighbor bands were different. It is worth noting that we are blind to the location of the stripes. Figure \ref{Simulated Salinas} displays the noise removal results of the competing methods. It is obvious that there still exist some residual stripes in Fig. \ref{Simulated Salinas}(c), (e), (h), (h), (i), and (k), which means these methods only work well for the random noise. In Fig. \ref{Simulated Salinas}(l), the stripes are perfectly removed, and the detailed structure information in each images are well preserved without the introduction of any noticeable artifacts.

    We also test the real stripe noise including the multiplicative stripe, as shown in Fig. \ref{Real Gulf Wetlands}. The multiplicative stripes are signal-dependent, while the additive stripes are signal-independent. We can observe that only PARAFAC, LRMR and WLRTR-RPCA work well in presence of stripe noise, while other methods fail to remove the stripe noise. Unfortunately, the PARAFAC has damaged the image details, and LRMR have introduced some false artifacts in Fig. \ref{Real Gulf Wetlands}(f). The proposed WLRTR-RPCA could fully decoupled the stripe noise (sparse error component) and image components (low-rank component), which is very applicable to this mixed noise removal task.

\begin{table}[tbp]
\scriptsize
\centering
\renewcommand{\arraystretch}{1.2}
\caption{Quantitative results of the competing methods under mixed noise on Salinas dataset.}
\begin{tabular}{c c c c c}
\toprule[2pt]
\emph{Method}  & \emph{PSNR }&  \emph{SSIM} &  \emph{ERGAS} &  \emph{SAM} \\
\hline
\emph{Noisy}   &  21.57      &   0.2706     &   185.21    &  0.1787     \\
\hline
\emph{BM3D}    &  29.16      &   0.7468     &   50.93     &  0.0288     \\
\hline
\emph{PARAFAC} &  31.25      &   0.8220     &   66.52     &  0.0291     \\
\hline
\emph{LRTA}    &  28.42      &   0.6911     &   41.85     &  0.0235     \\
\hline
\emph{LRMR}    &  35.28      &   0.8755     &   41.79     &  0.0276     \\
\hline
\emph{ANLM}    &  33.99      &   0.8704     &   42.40     &  0.0278     \\
\hline
\emph{NMF}     &  29.11      &   0.6781     &   35.30     &  0.0204     \\
\hline
\emph{BM4D}    &  36.18      &   0.9155     &   32.44     &  0.0198     \\
\hline
\emph{TDL}     &  38.80      &   0.9526     &   26.81     &  0.0117     \\
\hline
\emph{ITSReg}  &  33.32      &   0.8551     &   32.14     &  0.0234     \\
\hline
\emph{WLRTR-RPCA}     &    \textbf{39.38 }    &  \textbf{ 0.9594 }   &   \textbf{23.30}    &  \textbf{0.0104 }    \\
\bottomrule[2pt]
\end{tabular}
\label{Salinas Quantitative}
\end{table}

\begin{figure*}
\begin{center}
    \includegraphics[width=0.98\textwidth]{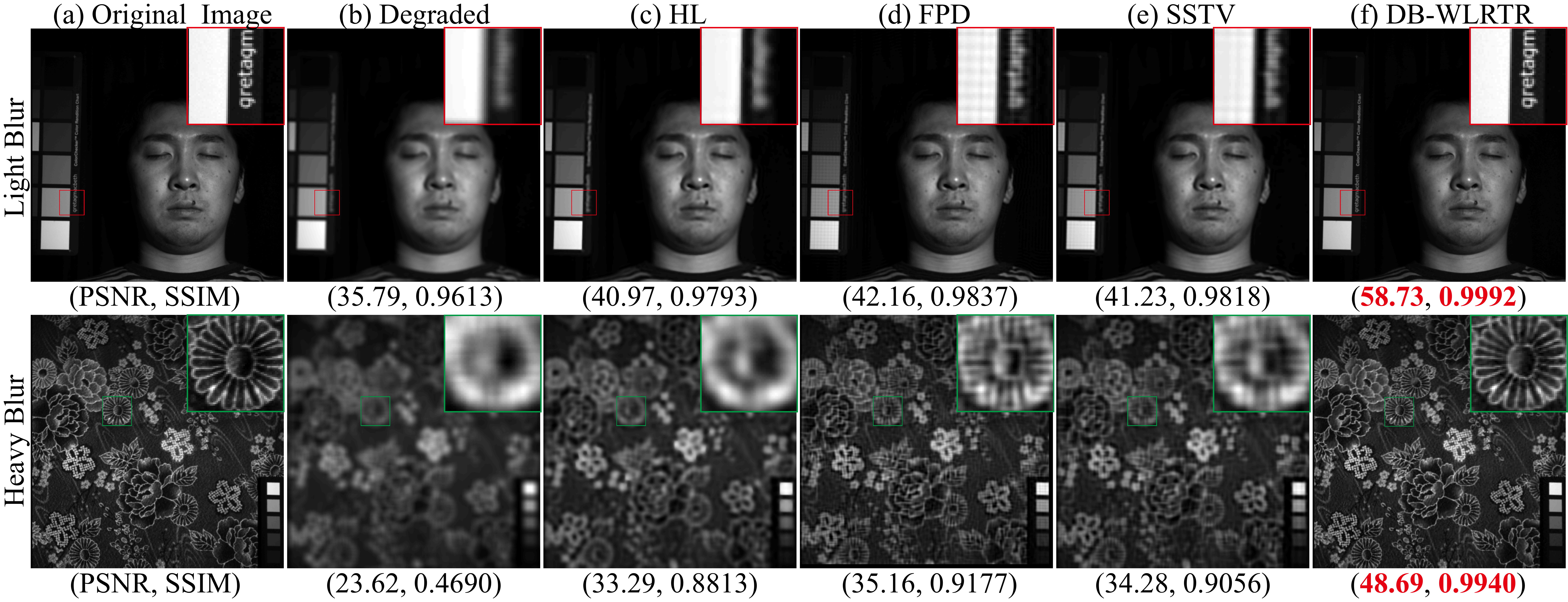}
\end{center}
   \caption{Simulated deblurring results under different blur level on CAVE dataset. The first row shows the light blur case (8*8, Sigma = 3), and second row displays the heavy blur case (17*17, Sigma = 7). (a) Original image at band 510nm. (b) Degraded image, Deblurring results by (c) HL, (d) FPD, (e) SSTV, (f) DB-WLRTR.}
\label{Deblurring}
\end{figure*}

\begin{table}[tbp]
\centering
\renewcommand{\arraystretch}{1.2}
\caption{Quantitative results of the competing methods under different blur levels on CAVE dataset.}
\label{CAVE deblurring quantitative}
\begin{tabular}{|c|c|c|c|c|}
\hline
Methods  & HL      & FPD    & SSTV   & DB-WLRTR \\ \hline
\multicolumn{5}{|c|}{Gaussian (8*8, Sigma = 3)}   \\ \hline
PSNR     & 37.28   & 38.84  & 37.61  & \textbf{55.68}    \\ \hline
SSIM     & 0.9460  & 0.9617 & 0.9527 & \textbf{0.9979}   \\ \hline
ERGAS    & 83.88   & 68.48  & 80.91  & \textbf{9.9635}   \\ \hline
SAM      & 0.0676  & 0.0734 & 0.0658 & \textbf{0.0250 }  \\ \hline
\multicolumn{5}{|c|}{Gaussian (17*17, Sigma = 7)} \\ \hline
PSNR     & 32.59   & 33.16  & 33.08  & \textbf{49.42}    \\ \hline
SSIM     & 0.8819  & 0.9114 & 0.8944 & \textbf{0.9926 }  \\ \hline
ERGAS    & 137.14  & 125.11 & 129.84 & \textbf{20.87}    \\ \hline
SAM      & 0.1075  & 0.1163 & 0.0989 & \textbf{0.0439}   \\ \hline
\end{tabular}
\end{table}

  %%%%%%%%%%%%%%%%%%%%%%%%%%%%%%%%%%%%%%%%%%%%%%%%%%%
  \subsection{HSI Deblurring}
  %%%%%%%%%%%%%%%%%%%%%%%%%%%%%%%%%%%%%%%%%%%%%%%%%%%
    There is relative fewer research paying attention on HSI deblurring. We compare the proposed DB-WLRTR method with single image based deblurring method hyper-Laplacian (HL) \cite{krishnan2009fast}, and two HSI deblurring methods FPD \cite{henrot2013fast} and SSTV \cite{fang2017hyperspectral}. The CAVE dataset is used for the comparison study. The Gaussian blur with different blur levels are tested. We assume the point spread function is known (nonblind deconvolution). From Table \ref{CAVE deblurring quantitative}, we can see that the proposed DB-WLRTR has overwhelming advantage over the other methods under different Gaussian blur levels. The visual comparisons of the deblurring methods are shown in Figs. \ref{Deblurring}, from which we can see that the DB-WLRTR method produces much cleaner and sharper image edges and textures than other methods. It is noteworthy that the lost details, such as the text and the artificial flower, can be well recovered by our method.

\begin{table}[tbp]
\centering
\renewcommand{\arraystretch}{1.2}
\caption{Quantitative results of the competing methods under different downsampling cases on CAVE dataset.}
\label{CAVE super-resolution quantitative}
\begin{tabular}{|c|c|c|c|c|}
\hline
Methods   & CNMF     & NLSTF    & NSSR     & SR-WLRTR  \\ \hline
\multicolumn{5}{|c|}{Gaussian (s = 8, 8*8, Sigma = 3)} \\ \hline
PSNR      & 46.15    & 44.56    & 46.99    & \textbf{47.39}     \\ \hline
SSIM      & 0.9901   & 0.9816   & 0.9921   & \textbf{0.9931}    \\ \hline
ERGAS     & 35.15    & 41.89    & 30.21    & \textbf{29.26}     \\ \hline
SAM       & 0.0591   & 0.0961   & 0.0528   & \textbf{0.0500}    \\ \hline
\multicolumn{5}{|c|}{Uniform (s = 8)}                  \\ \hline
PSNR      & 46.49    & 45.00    & 47.51    & \textbf{47.57}     \\ \hline
SSIM      & 0.9909   & 0.9847   & 0.9931   & \textbf{0.9934}    \\ \hline
ERGAS     & 34.56    & 39.06    & 28.84    & \textbf{28.82}     \\ \hline
SAM       & 0.0568   & 0.0869   & 0.0512   & \textbf{0.0493 }   \\ \hline
\end{tabular}
\end{table}

\begin{figure*}
\begin{center}
    \includegraphics[width=0.95\textwidth]{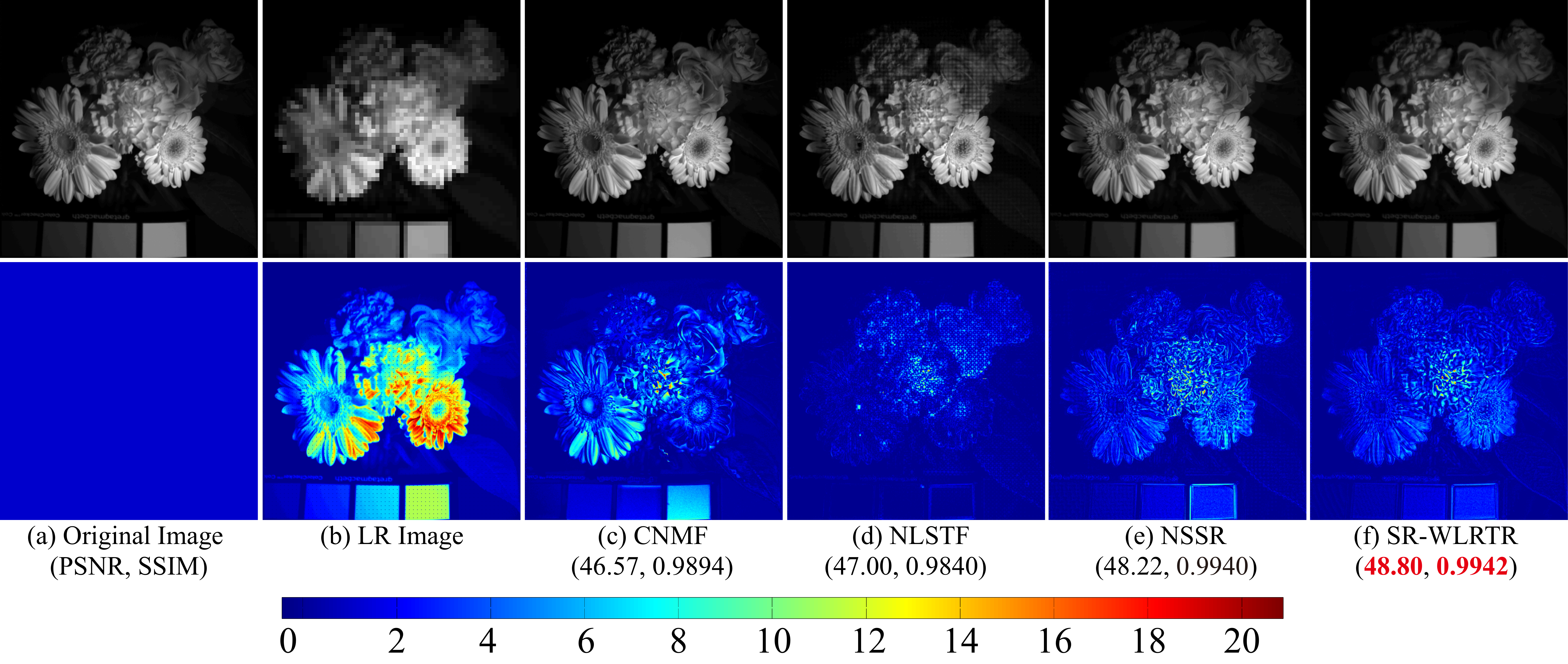}
\end{center}
   \caption{Simulated SR results on CAVE dataset. The first row shows the SR results. The second row is the corresponding error map. From the first column to the last one is (a) Original image at band 700nm, (b) Low-resolution image (s = 8, 8*8, Sigma = 3), SR results by (c) CNMF, (d) NLSTF, (e)NSSR, (f) SR-WLRTR.}
\label{super-resolution}
\end{figure*}

  %%%%%%%%%%%%%%%%%%%%%%%%%%%%%%%%%%%%%%%%%%%%%%%%%%%
  \subsection{HSI Super-resolution}
  %%%%%%%%%%%%%%%%%%%%%%%%%%%%%%%%%%%%%%%%%%%%%%%%%%%
    We also test the proposed SR-WLRTR model on HSI super-resolution. The CAVE dataset is used for the comparison study. Both the Gaussian and uniform blur are tested with scaling factors s = 8. We compare the SR-WLRTR with the representative state-of-the-arts methods, including both the matrix based CNMF \cite{yokoya2012coupled}, NSSR \cite{dong2016hyperspectral} and tensor based NLSTF \cite{dian2017hyperspectral}. The quantitative results are shown in Table \ref{CAVE super-resolution quantitative}. It can be seen that the results of proposed method are superior to the competing methods both the spatial and spectral aspects, especially the Gaussian blur case. One visual comparison results at 700nm of the \emph{flower} by all competing methods are shown in Fig. \ref{super-resolution}. All the competing methods can well recover the HR spatial structures of the scene, but the proposed method achieves the smallest reconstruction errors, especially for the sharp edges. In conclusion, compared with the matrix based methods, the SR-WLRTR could better preserve the spatial-spectral structures with better recovering the spatial details and less spectral distortion; compared with the tensor based methods NLSTF \cite{dian2017hyperspectral}, the SR-WLRTR utilizes the low-rank tensor prior in HSI, thus resulting in
    better visual pleasing result, while there is obvious gridding artifacts in the result of NLSTF.

\begin{figure}
\begin{center}
    \includegraphics[width=0.35\textwidth]{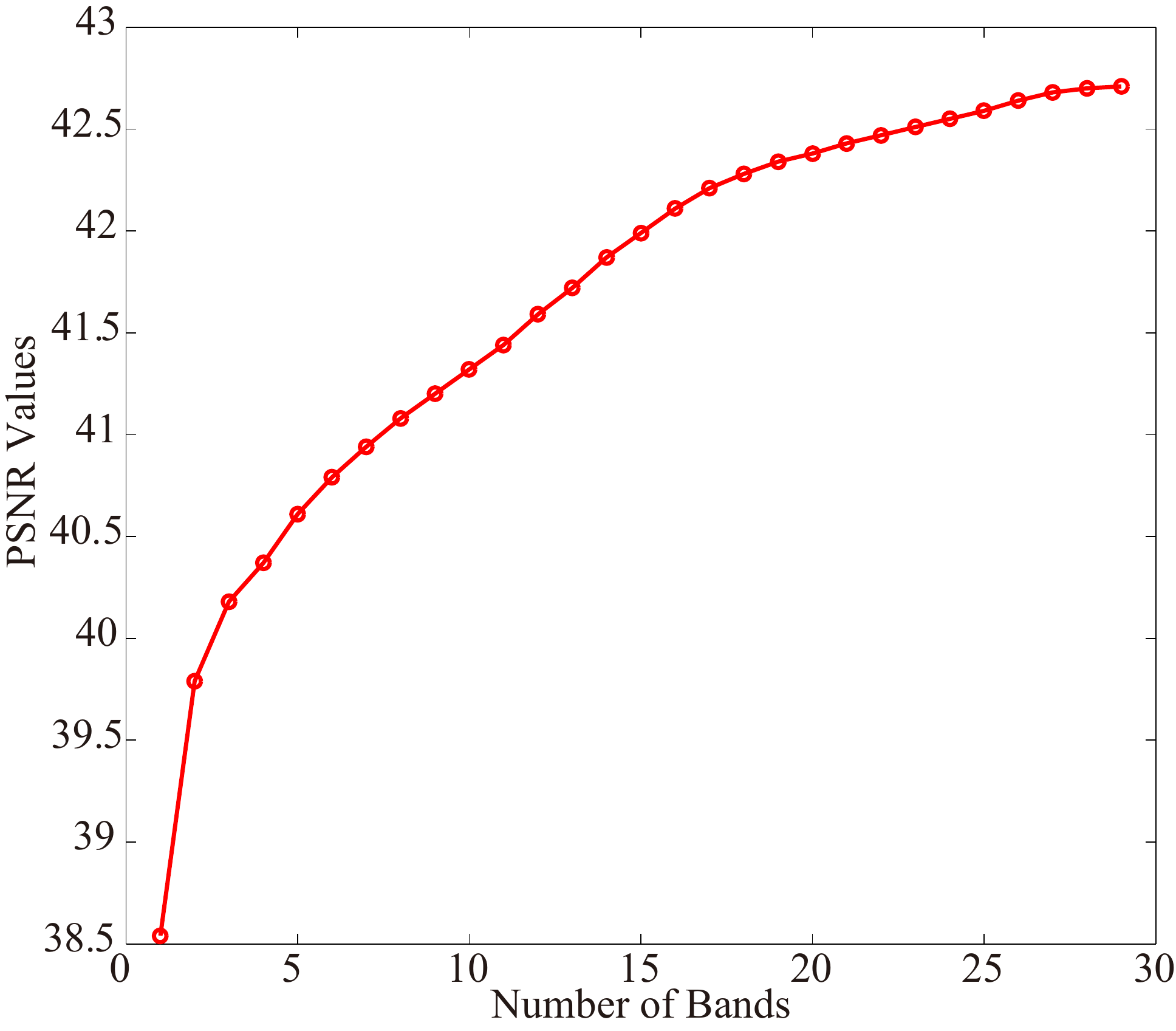}
\end{center}
   \caption{Effects of the band numbers on denoising results.}
\label{Number of Bands}
\end{figure}

\subsection{Discussion}
  %%%%%%%%%%%%%%%%%%%%%%%%%%%%%%%%%%%%%%%%%%%%%%%%%%%
  \subsubsection{\textbf{Number of Bands}}
  %%%%%%%%%%%%%%%%%%%%%%%%%%%%%%%%%%%%%%%%%%%%%%%%%%%
    Since the number of the band of input data is different, in this section, we present an analysis about the effect of the band number\footnote{We take the HSI denoising as an example, as are the following analysis.}. In Fig. \ref{Number of Bands}, we show the changes of the PSNR values with the different numbers of bands. From Fig. \ref{Number of Bands}, we can observe that the denoising results become gradually better with larger number of bands. More specifically, when the number of band is smaller than 20, the PSNR values increase rapidly. After the number of band is bigger than 20, the growing speed of the curve becomes relatively slow. It is worth noting that the curve still shows its tendency to grow up slowly. However, with the increasing size of the image bands, the memory and computational consumption also grow rapidly. Therefore, in our experiments, we empirically set the number of the bands smaller than 40. Normally, for an 512*512*31 images, it would cost about 23 minutes of running our algorithm on the personal computer with MATLAB 2014a, an Intel i7 CPU at 3.6 GHz, and 32-GB memory.

\begin{figure}
\begin{center}
    \includegraphics[width=0.5\textwidth]{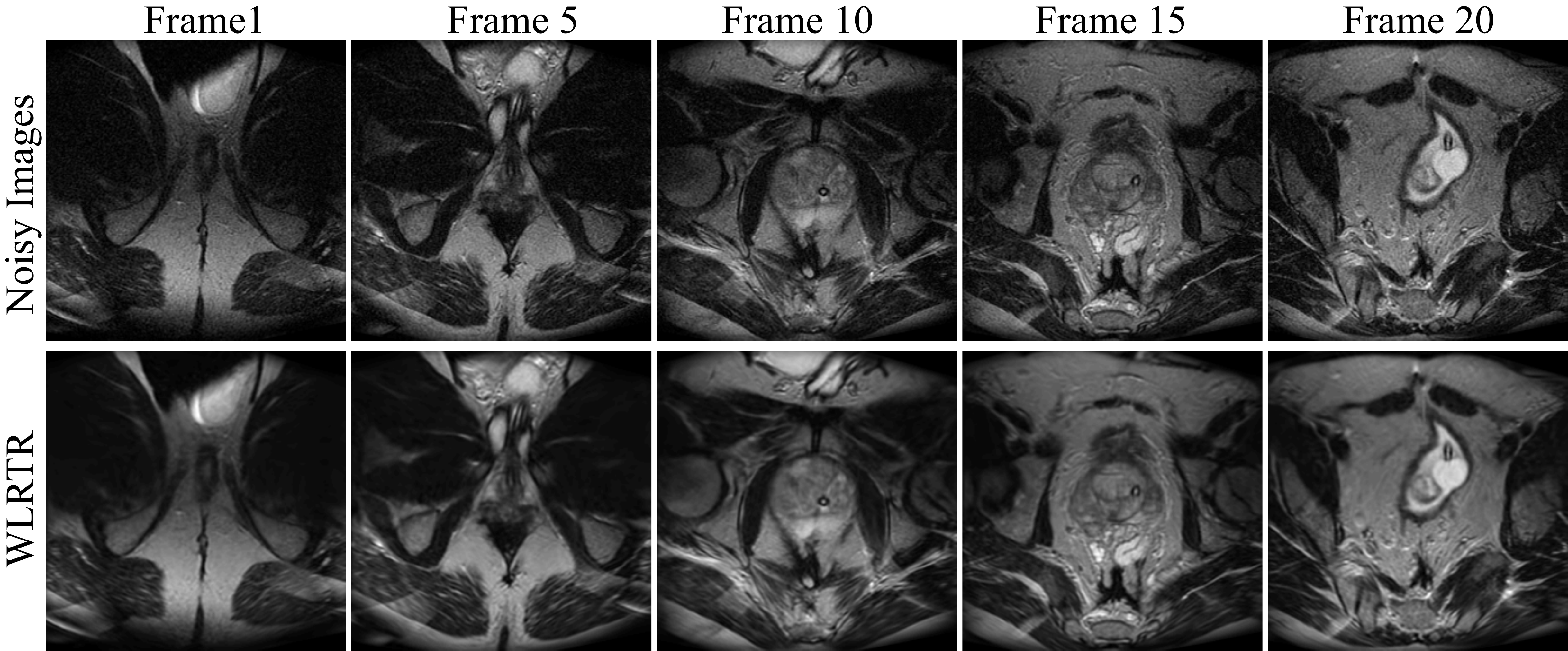}
\end{center}
   \caption{The robustness of WLRTR method under the band inconsistency situation. The first row shows the original PMRI images from frame 1 to frame 20. The second row shows the corresponding results of WLRTR.}
\label{Robustness of the band consistency}
\end{figure}

  %%%%%%%%%%%%%%%%%%%%%%%%%%%%%%%%%%%%%%%%%%%%%%%%%%%
  \subsubsection{\textbf{Band Consistency}}
  %%%%%%%%%%%%%%%%%%%%%%%%%%%%%%%%%%%%%%%%%%%%%%%%%%%
    Most HSIs restoration methods performs well when neighbor band consistency is guaranteed. What if this condition is not provided? We performed another experiment to demonstrate how WLRTR still works well. Figure \ref{Robustness of the band consistency} shows the denoising results of PMRIs image from frame 1 to 20 with far spectral difference at an interval of 5 frame. It can be seen that the original image of each frame in the first row varies sharply. The second row shows the corresponding denoising results of WLRTR. We can observe that not only the random noise is removed satisfactorily, but also the different structural edges of each frame has been preserved well. The main reason why WLRTR works well in this situation is that WLRTR utilizes the low-rank properties of the constructed 3-order tensor which contains both the band consistency and non-local cubic redundancy. Even when the band consistency cannot be guaranteed, the constructed 3-order tensor still has low-rank property induced by the non-local cubic redundancy, facilitating to obtain satisfactory denoising result.

\begin{figure}
\begin{center}
    \includegraphics[width=0.35\textwidth]{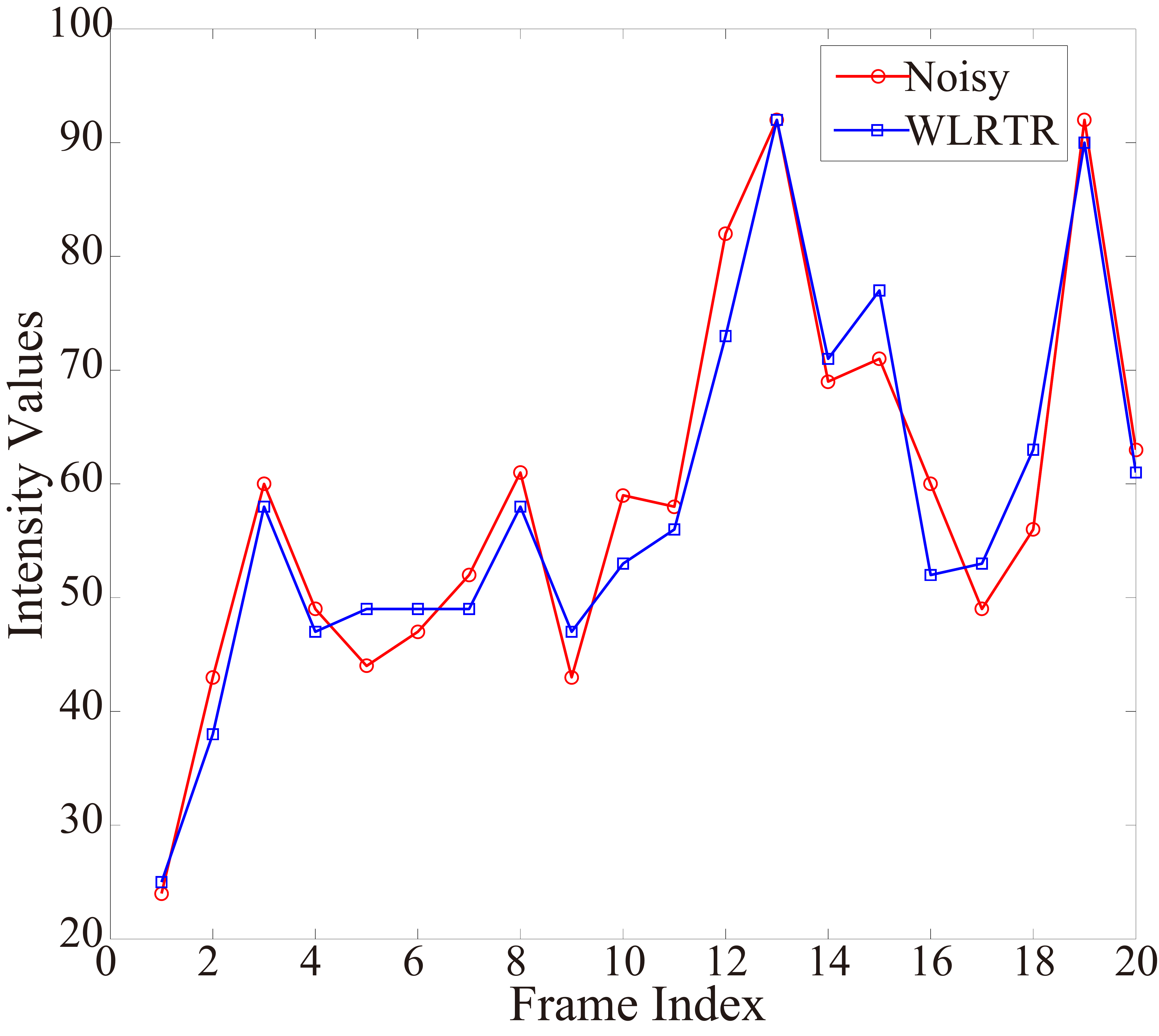}
\end{center}
   \caption{Spectral reflectance at one location from PMRI images before and after denoising.}
\label{Spectral reflectance}
\end{figure}

    To further verify that WLRTR can preserve the useful spectral/frame information while removing the noise in presence of low band consistency, in Fig. \ref{Spectral reflectance}, we show the reflectance spectra of one pixel (corresponding to Fig. \ref{Robustness of the band consistency}) before and after denoising as an example. It can be seen that the interframe information has been satisfactorily preserved with slightly difference due to the noise reduction.
    
\begin{figure}
\begin{center}
    \includegraphics[width=0.35\textwidth]{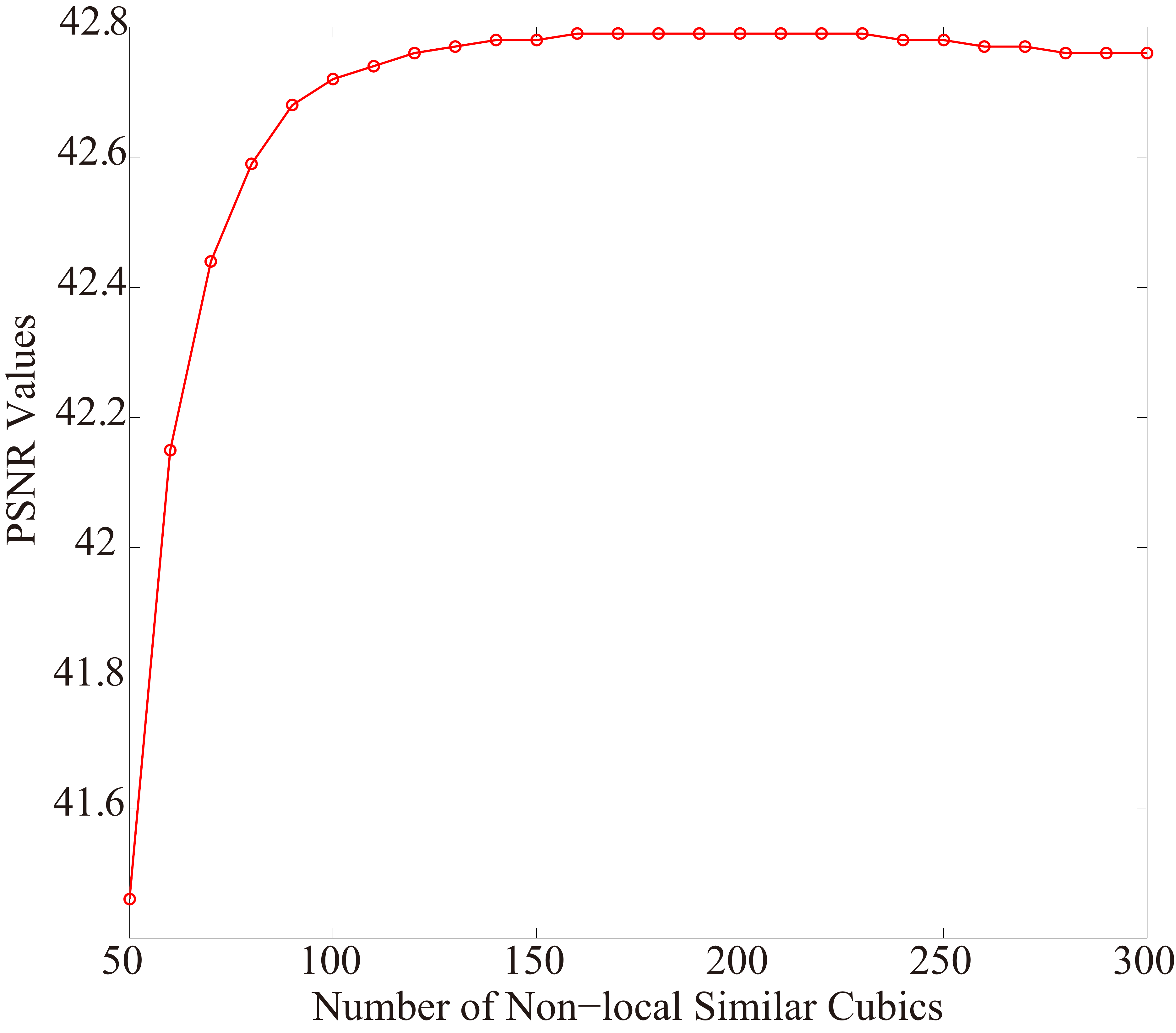}
\end{center}
   \caption{Effects of the numbers of the non-local similarity cubics on denoising results.}
\label{Number of non-local cubics}
\end{figure}

\begin{figure}
\begin{center}
    \includegraphics[width=0.35\textwidth]{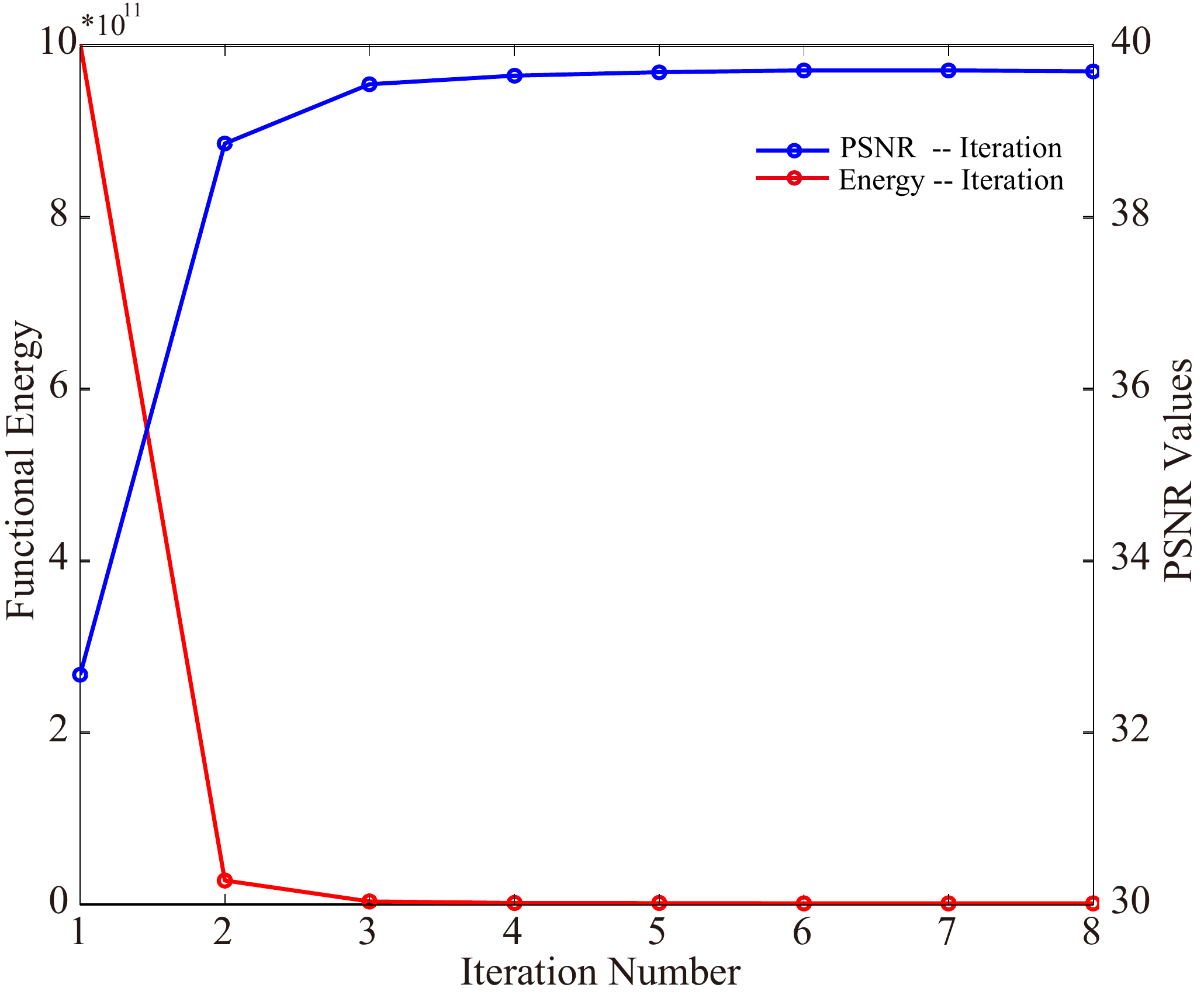}
\end{center}
   \caption{The empirical analysis of algorithm convergence.}
\label{Convergence}
\end{figure}

\begin{table*}[tbp]
\centering
\renewcommand{\arraystretch}{1.2}
\caption{A comparison of state-of-the-art HSI restoration methods and their properties.}
\label{Summary}
\begin{tabular}{c|c|c|c|c|c|c|c}
\toprule[2pt]
\emph{Method}  & \emph{Input}          & \emph{Paradigm}   & \emph{Task}              & \emph{Scalability} & \emph{Sparsity}                    & \emph{Speed} & \emph{Performance}         \\ \hline
\emph{BM3D}\cite{dabov2007image}    & single image   & 3-D tensor & denoising         & no          & spatial+nonlocal          & 2             & $\score{3}{3}$                       \\ \hline
\emph{PARAFAC}\cite{liu2012denoising} & multiple image & 3-D tensor & denoising+destriping         & no          & spatial+spectral   & 157     & $\score{2}{2}$                   \\ \hline
\emph{LRTA}\cite{renard2008denoising}    & multiple image & 3-D tensor & denoising         & no          & spatial+spectral          &   9      & $\score{2}{2}$                       \\ \hline
\emph{LRMR}\cite{zhang2014hyperspectral}    & multiple image & 2-D matrix & denoising+destriping         & yes         & spatial+spectral & 561 & $\score{2.5}{3}$             \\ \hline
\emph{ANLM}\cite{manjon2010adaptive}    & multiple image & 1-D vector & denoising         & no          & spatial+spectral+nonlocal & 163       & $\score{2}{2}$                          \\ \hline
\emph{NMF}\cite{ye2015multitask}    & multiple image & 2-D matrix & denoising         & yes         & spatial+spectral          & 258   & $\score{3}{3}$                       \\ \hline
\emph{BM4D}\cite{Maggioni2012Nonlocal}    & multiple image & 4-D tensor & denoising         & no          & spatial+spectral+nonlocal & 263   & $\score{3.5}{4}$                             \\ \hline
\emph{TDL}\cite{peng2014decomposable}     & multiple image & 3-D tensor & denoising         & no          & spatial+spectral+nonlocal & 58    & $\score{3.5}{4}$                           \\ \hline
\emph{ITSReg}\cite{xie2016multispectral} & multiple image & 3-D tensor & denoising         & yes         & spatial+spectral+nonlocal &  2454     & $\score{4}{4}$         \\ \hline
\emph{HL}\cite{krishnan2009fast}      & single image   & 1-D vector & deblurring        & yes         & spatial                   &    7   & $\score{2}{2}$                               \\ \hline
\emph{FPD}\cite{henrot2013fast}     & multiple image & 3-D tensor & deblurring        & yes         & spatial+spectral          &    207   & $\score{2.5}{3}$                               \\ \hline
\emph{SSTV}\cite{fang2017hyperspectral}    & multiple image & 3-D tensor & deblurring        & yes         & spatial+spectral          &    157   & $\score{2.5}{3}$                          \\ \hline
\emph{CNMF}\cite{yokoya2012coupled}    & multiple image & 2-D matrix & super-resolution & yes         & spatial+spectral          &   41    & $\score{3}{3}$                       \\ \hline
\emph{NLSTF}\cite{dian2017hyperspectral}   & multiple image & 3-D tensor & super-resolution & no          & spatial+spectral+nonlocal &   400   & $\score{3.5}{4}$                    \\ \hline
\emph{NSSR}\cite{dong2016hyperspectral}    & multiple image & 2-D matrix & super-resolution & yes         & spatial+nonlocal          &   132    & $\score{4.5}{5}$            \\ \hline
\emph{WLRTR}   & multiple image & 3-D tensor & comprehensive     & yes         & spatial+spectral+nonlocal &   1421    & $\score{5}{5}$                       \\ \bottomrule[2pt]
\end{tabular}
\end{table*}

  %%%%%%%%%%%%%%%%%%%%%%%%%%%%%%%%%%%%%%%%%%%%%%%%%%%
  \subsubsection{\textbf{Number of Non-local Similarity Cubics}}
  %%%%%%%%%%%%%%%%%%%%%%%%%%%%%%%%%%%%%%%%%%%%%%%%%%%
   We discuss another important parameter for the restoration performance: number of non-local similarity cubics. In Fig. \ref{Number of non-local cubics}, we show the changes of the PSNR values with the different numbers of non-local similarity cubics. From Fig. \ref{Number of non-local cubics}, we can observe that the denoising results become gradually better with larger number of bands. More specifically, when the number of band is smaller than 100, the PSNR values increase extremely fast. After the number of band is bigger than 100, the growing speed of the curve becomes relatively slow, and the PSNR value achieves its highest between 150 to 200. When the number of band is bigger than 230, the performance of WLRTR even deteriorates a little. We suppose it is due to the insufficient similarity between the target cubic and searching cubics. Therefore, in our work, we set the number of the non-local similarity cubics between 100 to 200.

  %%%%%%%%%%%%%%%%%%%%%%%%%%%%%%%%%%%%%%%%%%%%%%%%%%%
  \subsubsection{\textbf{Empirical Convergence}}
  %%%%%%%%%%%%%%%%%%%%%%%%%%%%%%%%%%%%%%%%%%%%%%%%%%%
    We provide an empirical analysis for the convergence of the proposed algorithm. Figure \ref{Convergence} illustrates the evolutional curve of functional energy and PSNR values versus the iterations, whose result is shown in Fig. \ref{Simulated_Watercolors}. We can observe that the functional energy curve rapidly and monotonically decreases to zero and the PSNR values curve rapidly increases to the stable value in a few iteration numbers. The algorithm often converges in just few iterations (empirically 3 or 4 step) in our implementation.

  %%%%%%%%%%%%%%%%%%%%%%%%%%%%%%%%%%%%%%%%%%%%%%%%%%%
  \subsubsection{\textbf{Overall Comparison}}
  %%%%%%%%%%%%%%%%%%%%%%%%%%%%%%%%%%%%%%%%%%%%%%%%%%%

    In Table \ref{Summary}, we give a detailed comparison between WLRTR and the state-of-the-art HSI restoration methods. Overall, the WLRTR and its extensions are consistent for all kinds of HSIs tasks, also other multispectral images (color image and MRIs), and obtain better visual pleasure result with fewer artifacts than the results obtained by the compared methods. Although the various tested images are much related to the particular imaging platform and degradation mechanism, our method captures the intrinsic low-rank property of the constructed  3-order tensor in three aspects: the spatial sparsity (mode-1), non-local cubic redundancy (mode-2), and band consistency (mode-3), and the weighted sparsity of the coefficients in the core tensor. Every MSI maintains these sparsity properties, no matter where it comes from. The WLRTR explicitly utilize this sparsity with low-rank tensor prior, which makes it applicable for different tasks.

\section{Conclusion}

    In this paper, we have proposed a unified weighted low-rank tensor recovery method for HSIs restoration. The proposed WLRTR explicitly utilizes the spatial sparsity, non-local spatial-spectral cubic redundancy, and spectral consistency via high order low-rank property of each constructed 3-order sub-tensor. We overcome the barriers of classical HSIs restoration methods that they are not able to preserve the spatial-spectral structures correlation and can only be applied to one specific task. On one hand, we clearly reveal the fact that tensor based sparsity model indeed fits for the HSIs processing; on the other hand, thanks to the variable splitting methods, we show that various HSIs restoration problem can be unified in a framework, and transformed into several easier subproblem with closed-form solution. Further, for the low-rank tensor prior related subproblem, we introduce the weighted strategy to improve the performance, in which its closed-form solutions has been analyzed. In addition, we consider the very common stripe noise in HSIs, ultilize its structural and directional property, and extend WLRTR to the WLRTR-RPCA model.

    Extensive simulated and real experiment results have been carried out against a number of competing state-of-the-art methods on various HSIs restoration tasks. The proposed methods have consistently outperform state-of-the-art methods in both quantitative assessments and visual appearance, especially in HSI destriping, deblurring, and super-resolution domain, where few tensor based methods have been proposed. Due to the efficiency of the low-rank tensor prior, the proposed WLRTR can be applied to other 3-D data applications, such as color image and MRIs.

    In the future, we will try to speed up the proposed method via paralleled implementation and reducing the computational complexity. It is also possible to apply our WLRTR model for HSI compress sensing, unmixing and also the video applications.

\bibliographystyle{spmpsci}      % mathematics and physical sciences

{\footnotesize
\bibliography{shortstrings,vggroup,cvww_template,ref}
}

\end{document}